\documentclass[twoside,11pt]{article}

\usepackage{jmlr2e}
\usepackage{natbib}
\usepackage{amsfonts,mathtools}
\usepackage{algorithmic}
\usepackage[]{algorithm}
\usepackage{array}
\usepackage{graphicx}
\usepackage{mathrsfs}
\usepackage{amsmath,bm}

\usepackage{cite}
\usepackage{color}
\usepackage{float}
\usepackage[font={small}]{caption}
\usepackage{epstopdf}
\usepackage{amssymb}

\usepackage{graphicx}
\usepackage{subcaption}

\usepackage{amsthm}

\DeclareMathOperator*{\argmin}{arg\,min}
\renewcommand{\b}[1]{\ensuremath{\mathbf{#1}}}		 							% bold
\newcommand{\norm}[1]{\ensuremath{\left\|#1\right\|}}						% norm operator
\providecommand{\tr}[1]{\text{tr}\left(#1\right)}

\newcommand{\arrowAS}[1]{\ensuremath{\stackrel{\text{a.s.}}{\longrightarrow}}}

\providecommand{\norm}[1]{\left \| #1 \right \|}

\providecommand{\inPro}[2]{\left \langle #1, \; #2 \right \rangle}

\providecommand{\norm}[1]{\left \| #1 \right \|}

\theoremstyle{plain} %% Theroem style plain
\newtheorem{thm}{Theorem}
\newtheorem{Lem1}{Lemma}
\theoremstyle{definition} %% Theroem style definition
\newtheorem{def1}{Definition}

\theoremstyle{remark} %% Theroem style Remark

\newtheorem{rem}{\bf Remark}%​

\def \Rn {{\mathbb{R}}}
\def \x {{\b{x}}}
\def \z {{\b{z}}}

\def \y {{\b{y}}}

\def \L {{\b{L}}}

\def \I {{{I}}}

\def \H {{{H}}}

\def \bb {{\b{b}}}
\def \cb {{\b{c}}}

\def \K {{\b{K}}}

\def \U {{{U}}}
\def \J {{{J}}}

\def \K {{{K}}}

\def \Y {{{Y}}}
\def \V {{{V}}}

\def \E {{\mathcal{E}}}
\def \G {{\mathcal{G}}}

\def \N {{\mathcal{N}}}

\def \cX {{\mathcal{X}}}

\def \bpsi {\boldsymbol{{\psi}}}

\def \S {{\mathcal{S}}}

\def \Di {{{\text{Diag}}}}

\def \bTheta {{{\Theta}}}

\def \blambda {{\boldsymbol{\lambda}}}

\def \Ao {{\mathcal A}}
\def \L {{\mathcal L}}

\def \w {\b{{w}}}

\def \Vc {{\mathcal{V}}}

\def \bzero {{\b{0}}}
\def \bone {{\b{1}}}

\begin{document}

	\title{ A Unified Framework for Structured Graph Learning\\ via Spectral Constraints}
	
	\author{\name Sandeep Kumar\email eesandeep@ust.hk \\
		\addr Department of Electronic and Computer Engineering\\
	Hong Kong University of Science and Technology\\
	Hong Kong 
		\AND
	\name Jiaxi Ying\email jx.ying@connect.ust.hk \\
		\addr Department of Electronic and Computer Engineering\\
	Hong Kong University of Science and Technology\\
	Hong Kong 
			\AND
	\name Jos\'{e} Vin\'{i}cius de M. Cardoso \email jvmirca@gmail.com \\
	\addr Department of Electrical Engineering\\
	Universidade Federal de Campina Grande\\
	Brazil
			\AND
	\name Daniel P. Palomar \email palomar@ust.hk\\
	\addr Department of Electronic and Computer Engineering\\
Hong Kong University of Science and Technology\\
Hong Kong
}

	\editor{}
	
	\maketitle

\begin{abstract}	
 Graph learning from data represents a canonical problem that has received substantial attention in the literature.
However, insufficient work has been done in incorporating prior structural knowledge onto the learning of underlying graphical models from data. Learning a graph with a specific structure is essential for interpretability and identification of the relationships among data. Useful structured graphs include the multi-component graph, bipartite graph, connected graph, sparse graph, and regular graph. In general structured graph learning is an NP-hard combinatorial problem therefore designing a general tractable optimization method is extremely challenging. In this paper, we introduce a unified graph learning framework lying at the integration of Gaussian graphical models and spectral graph theory. To impose a particular structure on a graph, we first show how to formulate the combinatorial constraints as an analytical property of the graph matrix. Then we develop an optimization framework that leverages graph learning with specific structures via { spectral constraints} on graph matrices. The proposed algorithms are provably convergent, computationally efficient, and practically amenable for numerous graph-based tasks. Extensive numerical experiments with both synthetic and real data sets illustrate the effectiveness of the proposed algorithms. The code for all the simulations is made available as open source repository.\\

\end{abstract}

\section{Introduction}
Graphs are fundamental mathematical structures consisting of sets of nodes and weighted edges among them. The weight associated with each edge represents the similarity between the two vertices it connects. Graphical models provide an effective abstraction for expressing dependence relationships among data variables available across numerous applications \citep[see][]{barabasi2016network,wang2018network,friedman2008sparse,guo2011joint,segarra2017network,banerjee2008model}. The aim of any graphical model is to encode the dependencies among the data in the form of a graph matrix, where non-zero entries of the matrix imply the dependencies among any two variables. Gaussian graphical modeling (GGM) encodes the conditional dependence relationships among a set of variables \citep{dempster1972covariance,lauritzen1996graphical}. GGM is a tool of increasing importance in a number of fields, including finance, biology, statistical learning, and computer vision \citep{friedman2008sparse}. In this framework, an undirected graph is matched to the variables, where each vertex corresponds to one variable, and an edge is present between two vertices if the corresponding random variables are conditionally dependent \citep{lauritzen1996graphical}. Putting it more formally, consider an $p-$dimensional vector $\x=[x_1,x_2,\cdots,x_p]^T $, the GGM method aims to learn a graph through the following optimization problem
\begin{align}\label{cost}
\underset{\Theta \in \S_{++}^p }{\text{maximize}} &
\begin{array}{c}
\hspace{.75cm} \log \det (\Theta)-\text{tr}\big({\Theta S }\big)-\alpha h({\Theta}), 
\end{array}
\end{align}
where $\Theta\in \Rn^{p\times p}$ denotes the desired graph matrix with $p$ the number of nodes in the graph, $\S_{++}^p $ denotes the set of positive definite matrices of size $p$, $S\in \Rn^{p\times p}$ is a similarity matrix, $h(\cdot)$ is the regularization term, and $\alpha>0$ is the regularization parameter. When the observed data is distributed according to a zero-mean $p-$variate Gaussian distribution and the similarity matrix is the sample covariance matrix (SCM), the optimization in \eqref{cost} corresponds to the maximum likelihood estimation (MLE) of the inverse covariance (precision) matrix of the Gaussian random variable also known as Gaussian Markov Random Field (GMRF).
With the graph inferred from $\bTheta$, the random vector $\x$ follows the Markov property: $\bTheta_{ij}\neq 0$ implies $x_i$ and $x_j$ are conditionally dependent given the rest \citep[see][]{lauritzen1996graphical,dempster1972covariance}. 

In many real-world applications, prior knowledge about the underlying graph structure is usually available. For example, in gene network analysis, genes can be grouped into pathways, and connections within a pathway might be more likely than connections between pathways, forming a cluster \citep{marlin2009sparse}. For better interpretability and precise identification of the structure in the data, it is desirable to enforce structures on the learned graph matrix $\bTheta$. Furthermore, the structured graph also enables performing more sophisticated tasks such as prediction, community detection, clustering, and causal inference. 

%Structured graph learning from data should undertake the tasks of structure learning (also known as connectivity inference) and the graph weight estimation \citep{ambroise2009inferring}. 

It is known that if the ultimate goal is structured graph learning, structure inference and graph weight estimation should be done in a single-step \citep{ambroise2009inferring,JMLR:v18:17-019}. Performing the structure inference (also known as model selection) prior to the weight estimation (also known as parameter estimation) in the selected model will, in fact, result in a non-robust procedure \citep{ambroise2009inferring}.
Although GGM has been extended to incorporate structures on the learned graph, most of the existing methods perform graph structure learning and graph weight estimation separately. Essentially, the methods are either able to infer connectivity information \citep{ambroise2009inferring} or with known connectivity information could perform the graph weights estimation \citep[see][]{lee2015joint,wang2015joint,cai2016joint,danaher2014joint,pavez2018learning,egilmez2017graph}. Furthermore there are few recent works considering the two tasks jointly, but those methods are limited to some specific structures \citep[e.g., multi-component in][]{JMLR:v18:17-019} which cannot be extended to other graph structures. In addition, these methods involve computationally demanding multi-stage steps, which make it unsuitable for big data applications.

In general, structured graph learning is an {NP-}hard combinatorial problem \citep{anandkumar2012high,bogdanov2008complexity} which brings difficulty in designing a general tractable optimization method. In this paper, we propose to integrate spectral graph theory with GGM graph learning, and convert combinatorial constraints of graph structure into analytical constraints on graph matrix eigenvalues. Realizing the fact that combinatorial structures of a family of graphs (e.g., multi-component graph, bipartite graph, etc.) are encoded in the eigenvalue properties of their graph matrices, we devise a general framework of \textbf{S}tructured \textbf{G}raph (SG) learning by enforcing spectral constraints instead of combinatorial structure constraints directly. We develop computationally efficient and theoretically convergent algorithms that can learn graph structures and weights simultaneously.

\subsection{Related work}

The penalized likelihood approach with sparsity regularization has been widely studied in precision matrix estimation. An $\ell_1-$norm regularization ($ h(\Theta)= \sum_{i,j}| \Theta_{ij}|_1$) which promotes element-wise sparsity on the graph matrix $\bTheta$ is a common choice of regularization function to enforce a sparse structure \citep{yuan2007model,shojaie2010penalized,shojaie2010penalized-b,ravikumar2010high,mazumder2012graphical,fattahi2019graphical}. Authors in \citet{friedman2008sparse} came up with an efficient computational method to solve \eqref{cost} and proposed the well-known \textsf{GLasso} algorithm. In addition, non-convex penalties are proposed for sparse precision matrix estimation to reduce estimation bias \citep{shen2012likelihood,lam2009sparsistency}. However, if a specific structure is required then simply a sparse graphical modeling is not sufficient, since it only enforces a uniform sparsity structure \citep{heinavaara2016inconsistency,tarzanagh2017estimation}. Towards this, the sparse GGM model should be extended to incorporate more specific structures.

In this direction, the work in \citet{ambroise2009inferring} has considered the problem of graph connectivity inference for multi-component structure and developed a two-stage framework lying at the integration of expectation maximization (EM) and the graphical Lasso framework. The works in \citet{lee2015joint,wang2015joint,cai2016joint,danaher2014joint,guo2011joint,sun2015inferring,tan2015cluster} have considered the problem of edge-weight estimation with the known connectivity information.
However, prior knowledge of connectivity information is not always available, in particular for the massive data with complex and unknown population structures \citep{JMLR:v18:17-019, jeziorski2015makes}. Furthermore, considering simultaneous connectivity inference and graph weight estimation, two-stage methods based on Bayesian model \citep{marlin2009sparse} and expectation maximization \citep{JMLR:v18:17-019} were proposed, but these methods are computationally prohibitive and limited to only multi-componet graph structures.

Other important graph structures have also been considered for example: factor models in \citet{meng2014learning}, scale free in \citep{liu2011learning}, eigenvector centrality prior in \citet{fiori2012topology}, degree-distribution in \citet{huang2008maximum}, and overlapping structure with multiple graphical models in \citet{tarzanagh2017estimation,mohan2014node}, tree structure in \citet{chow1968approximating,anandkumar2012high}. Recently, there has been a considerable interest in enforcing the Laplacian structure \citep[see][]{lake2010discovering,slawski2015estimation,pavez2016generalized,kalofolias2016learn,egilmez2017graph,pavez2018learning} but all these methods are limited to learning a graph without specific structural constraints, or just learn Laplacian weights for a graph with the connectivity information. 

Due to the complexity posed by the graph learning problem, owing to its combinatorial nature, existing methods are tailored to specific structures which cannot be generalized to other graph structures; require connectivity information for graph weight estimation; often involve multi-stage framework and become computationally prohibitive. Furthermore, there does not exist any GGM framework to learn a graph with useful structures such as bipartite structure, regular structure and multi-component bipartite structure.

\subsection{Summary of contributions}
Enforcing a structure onto a graph is generally an {NP-}hard combinatorial problem, which is difficult to solve via existing methods. In this paper, we propose a unified framework of structured graph learning. Our contributions are threefold: 

First, we introduce new problem formulations that convert the combinatorial constraints into analytical spectral constraints on Laplacian and adjancency matrices, resulting in three main formulations:
\begin{itemize}
\item \textbf{Structured graph learning via Laplacian spectral constraints}:\\
	This formulation utilizes the Laplacian matrix spectral properties to learn {\textit{multi-component graph, regular graph, multi-component regular graph, sparse connected graph, modular graph, grid graph}} and other specific structured graphs.
\item \textbf{Structured graph learning via adjacency spectral constraints}\\ This formulation utilizes spectral properties of the adjacency matrix for {\it bipartite graph} learning.
	
\item \textbf{Structured graph learning via Laplacian and adjacency spectral constraints} \\	
Under this formulation we simultaneously utilize spectral properties of Laplacian and adjacency matrices to enforce non-trivial structures including {\it bipartite-regular graph, multi-component bipartite graph}, and {\it multi-component bipartite-regular graph} structures.
\end{itemize}

Second, we develop algorithms based on the block majorization-minimization (MM) framework also known as block successive upper-bound minimization (BSUM) to solve the proposed formulations. The algorithms are theoretically convergent and computationally efficient with worst case complexity $O(p^3)$, which is same as that of $\textsf{GLasso}$.

Third, we verify the effectiveness of the proposed algorithms via extensive synthetic and real data sets experiments. We believe that the work carried out in this paper will provide a starting point for structured graph learning based on Gaussian Markov random fields and spectral graph theory, which in turn may have a significant and long-standing impact. 
{The code for all the simulations is made available as open source repository on author's website\footnote{ https://github.com/dppalomar/spectralGraphTopology}.}

\subsection{ Outline and Notation} 
This paper is organized as follows. The generalized problem formulation and related background are provided in Section \ref{prob-for-spectral-constraints}. The detailed algorithm derivations and the associated convergence results are presented in Sections \ref{sec:lap}, \ref{sec:adj}, and \ref{sec:lap-adj}. Then the simulation results with both real and synthetic data sets for the proposed algorithms are provided in Section \ref{simulations}. Finally, Section \ref{conclusion} concludes the paper with a list of plausible extensions.

%\col{the paper and highlight several plausible extensions}.

In terms of notation, lower case (bold) letters denote scalars (vectors) and upper case letters denote matrices, whose sizes are not stated if they are clear from the context. The $(i,j)$-th entry of a matrix $X$ is denoted by $[X]_{ij}$. $X^\dagger$ and $X^T$ denote the pseudo inverse and transpose of matrix $X$, respectively. The all-zero and all-one vectors or matrices of all sizes are denoted by $\bzero$ and $\bone$, respectively. $\norm{X}_1$, $\norm{X}_F$ denote $\ell_1$-norm and Frobenius norm of $X$, respectively. $\text{gdet}(X)$ is defined as the generalized determinant of a positive semidefinite matrix $X$, i.e., the product of its non-zero eigenvalues. The inner product of two matrices is defined as $\langle X, Y\rangle=\text{trace}(X^TY)$. $\text{Diag} (X) $ is a diagonal matrix with diagonal elements of $X$ filling its principal diagonal and diag$(X)$ is a vector with diagonal elements of $X$ as the vector elements. Operators are defined using calligraphic letters.

\section{Problem Formulation}\label{prob-for-spectral-constraints}
A graph is denoted by $\mathcal{G}=\left( \mathcal{V}, \mathcal{E}\right)$, where $\mathcal{V}= \{1,2,\ldots,p \}$ is the vertex set, and $\mathcal{E}\in \Vc\times\Vc$ is the edge set. If there is an edge between vertices $i$ and $j$ we denote it by $j\in \N_i$. We consider a simple undirected graph with positive weights $w_{ij}>0$, having no self-loops or multiple edges and therefore its edge set consists of distinct pairs. Graphs are conveniently represented by some matrix (such as Laplacian and adjacency graph matrices), whose nonzero entries correspond to edges in the graph. The choice of a matrix usually depends on modeling assumptions, properties of the desired graph, applications, and theoretical requirements. 
%\begin{def1}

A matrix $\bTheta \in \Rn^{p \times p}$ is called as a graph Laplacian matrix if its elements satisfy 
	\begin{align}\label{Lap-set}
	\mathcal{S}_{\Theta} =\Big\{ \Theta| \Theta_{ij} =\Theta_{ji} \leq 0 \ {\rm for} \ i\neq j; \Theta_{ii}=-\sum_{j\neq i}\Theta_{ij} \Big\}.
	\end{align}
%\end{def1}
The properties of the elements of $\Theta$ in \eqref{Lap-set} imply that the Laplacian matrix $\Theta$ is: i) diagonally dominant (i.e., $|\Theta_{ii}| =|\sum_{j\neq i}\Theta_{ij}|$); ii) positive semidefinite, implied from the diagonally dominant property \citep[see ][Proposition 2.2.20.]{den1993linear}; iii) an $M$-matrix, i.e., a positive semidefinite matrix with non-positive off-diagonal elements \citep{slawski2015estimation}; iv) zero row sum and column sum of $\Theta$ (i.e., $\Theta_{ii}+\sum_{j\neq i}\Theta_{ij}=0$), which means that the vector $\mathbf{1}=[1,1,\dots,1]^T$ satisfies $ \Theta\mathbf{1} = \mathbf{0}$ \citep{chung1997spectral}.

We introduce the adjacency matrix $\bTheta_A$ as 
\begin{align}\label{thetaA}
\bTheta_A =
\begin{cases}
-\Theta_{ij}, & \text{if}\; i \neq j\\
0, \; & \text{if}\; \;i=j.
\end{cases}
\end{align}
The non-zero entries of the matrix encode edge weights as $\bTheta_{ij} = - w_{ij}$ and $\bTheta_{ij} = 0 $ implies no connectivity between vertices $i$ and $j$.

%	
%The Laplacian matrix and adjacnecy matrices are also related as 
%\begin{align}\label{lap:def2}
%\bTheta= \text{Diag}(\mathbf{A} \mathbf{1})-\mathbf{A},
%\end{align}

% Collecting all the pairwise weights $\{w_{ij}\}$ in the {\it Adjacency} matrix $\A$, the matrix $\A$ is ${p \times p}$ symmetric matrix and contains real and non-negative values with entries
%\[
%[\mathbf{A}]_{ij} =
%\begin{cases}
%w_{ij} & \text{if}\; j \in \mathcal{N}_i\\
%0 \; & \text{if}\; \;i=j, \; \text{or}\; j \notin \mathcal{N}_i.
%\end{cases}
%\]\label{lap:def}
%Another useful graph matrix is the {\it Laplacian} of a graph which is defined as
%\begin{align}\label{lap:def2}
%	\mathbf{L}= \text{Diag}(\mathbf{A}\cdot \mathbf{1})-\mathbf{A},
%\end{align}
%where $\text{Diag}(\cdot)$ constructs a diagonal matrix containing the vector argument along the diagonal and $\mathbf{1}$ is the all-one vector. 

\begin{def1}\label{def:igmrf}
	Let $\bTheta$ be an $p \times p$ symmetric positive semidefinite matrix with rank $p-k>0$. Then $\x=[x_1,x_2,\ldots,x_p]^T$ is an improper GMRF (IGMRF) of rank $p-k$ with parameters $\bm{\mu}, \bTheta$ (assuming $\bm{\mu}=\bzero$ without loss of generality), if its density is 
	\begin{align}
	p(\x)=(2\pi)^{\frac{-(p-k)}{2}}(\text{gdet}(\bTheta))^{\frac{1}{2}}\exp(-\frac{1}{2}(\x^\top\bTheta\x))
	\end{align}
where $ \text{gdet}(\cdot) $ denotes the generalized determinant \citep{rue2005gaussian} defined as the product of non-zero eigenvalues of $\Theta$. Furthermore, $\x$ is called IGMRF w.r.t to a graph $\G=(\Vc,\E)$, where
	\begin{align}
	&\bTheta_{ij}\neq 0 	\iff \{i,j\}\in \E \;\forall \; i\neq j \\
	 &\bTheta_{ij}=0 \iff x_i\perp x_j \vert \x\slash(x_i,x_j).
	\end{align}
	
\end{def1}
It simply states that the nonzero pattern of $\bTheta$ determines $\G$, so we can read off from $\bTheta$ whether $x_i$ and $x_j $ are conditionally independent. If the rank of $\bTheta$ is exactly $p$ then $\x$ is called GMRF and parameters ($\bm{\mu},\bTheta$) represent the mean and precision matrix corresponding a $p$-variate Gaussian distribution \citep{rue2005gaussian}. In addition, if precision $\bTheta$ has non-positive off-diagonal entries \citep{slawski2015estimation} then random vector $\x$ is called an {\it attractive} improper GMRF.

%the transformation of matrix $\Theta$

%\col{$ \mathcal{T}(\bTheta)$ maps the graph Laplacian matrix $\bTheta$ to another graph matrix (Laplacian or adjacency); we will shortly make the notion of this operator precise, $\boldsymbol{\xib}(\mathcal{T}(\Theta))\in \Rn^{p\times 1}$ is the vector containing the eigenvalues of the matrix $\mathcal{T}(\Theta)$}, and $\mathcal{S}_{{\xi}}$ denotes the spectral constraints on the eigenvalues.

\subsection{A General Framework for Graph Learning under Spectral Constraints}
A general scheme is to learn the matrix $\bTheta$ as a Laplacian matrix under some eigenvalue constraints, which are motivated from the a priori information for enforcing structure on the learned graph. Now we introduce a general optimization framework for structured graph learning via spectral constraints on the graph matrices, 
\begin{align} \label{CGL_L:0:a}
\begin{array}{ll}
\underset{\Theta}{\text{maximize}} &
\begin{array}{c}
\hspace{.75cm}\log\;\text{gdet}(\Theta)-\text{tr}\big({\Theta S}\big) -\alpha h(\Theta),
\end{array}\\
\text{subject to} & \begin{array}[t]{l}
\hspace{.75cm}\Theta \in \mathcal{S}_{\Theta}, \ \blambda_{\mathcal{T}}(\Theta)\in \mathcal{S}_{\mathcal{T}}, 
\end{array}
\end{array}
\end{align}
where $S$ denotes the observed data statistics (e.g., the sample covariance matrix), $\Theta$ is the sought graph matrix to be optimized, $\mathcal{S}_{\Theta}$ is the Laplacian matrix structural constraint set \eqref{Lap-set}, $h(\cdot)$ is a regularization term (e.g., sparsity), $\blambda_{\mathcal{T}}(\Theta)$ denotes the eigenvalues of $\mathcal{T}(\Theta)$, which is the transformation of matrix $\Theta$. More specifically, if $\mathcal{T}$ is identity, then $\mathcal{T}(\Theta)=\Theta$, implying we impose constraints on the eigenvalues of the Laplacian matrix $\Theta$; if $\mathcal{T}(\Theta)=\Theta_A$ defined in \eqref{thetaA}, then we enforce constraints on the eigenvalues of the adjacency matrix $\Theta_A$, and $\mathcal{S}_{{\mathcal{T}}}$ is the set containing spectral constraints on the eigenvalues.

Fundamentally, the formulation in $\eqref{CGL_L:0:a}$ aims to learn a structured graph Laplacian matrix $\Theta$ given data statistics $S$, where $\mathcal{S}_{\Theta}$ enforces Laplacian matrix structure and $\mathcal{S}_{\mathcal{T}}$ allows to include structural constraints of desired graph structure via spectral constraints on the eigenvalues. Observe that the formulation \eqref{CGL_L:0:a} has converted the complicated combinatorial structural constraints into the simple analytical spectral constraints, due to which, now the structured graph learning becomes a matrix optimization problem under the proper choice of spectral constraints.

\begin{rem}
	Apart from motivation of enforcing structure onto a graph, the Laplacian matrix is also desirable from numerous practical and theoretical considerations: i) Laplacian matrix is widely used in spectral graph theory, machine learning, graph regularization, graph signal processing, and graph convolution networks \citep{smola2003kernels, defferrard2016convolutional, egilmez2017graph, chung1997spectral}; ii) in the high-dimensional setting where the number of the data samples is less than the dimension of the data, learning $\Theta$ as an $M-$matrix greatly simplifies the optimization problem by avoiding the need for the explicit regularization term $h(\cdot)$ \citep{slawski2015estimation}; iii) the graph Laplacian is crucial for utilizing the GMRF framework, which requires the matrix $\Theta$ to have the positive semi-definite property \citep{rue2005gaussian}; iv) the graph Laplacian allows flexibility in incorporating useful spectral properties of graph matrices\citep{chung1997spectral,spielman2011spectral}.
\end{rem}

\begin{rem}
From the probabilistic perspective, when the similarity matrix $S$ is the sample covariance matrix of Gaussian data, \eqref{CGL_L:0:a} can be viewed as penalized maximum likelihood estimation problem of structured precision matrix of an improper attractive GMRF model, see Definition \ref{def:igmrf}. In a more general setting with arbitrarily distributed data, when the similarity matrix is positive definite matrix, then formulation \eqref{CGL_L:0:a} can be related to the log-determinant Bregman divergence regularized optimization problem \citep[see][]{dhillon2007matrix,duchi2012projected,slawski2015estimation}, where the goal is to find the parameters of multivariate Gaussian model that best approximates the data.
\end{rem}

In the coming subsections, we will specialize the optimization framework in \eqref{CGL_L:0} under Laplacian eigenvalue constraints, adjacency eigenvalue constraints, and joint Laplacian and adjacency eigenvalue constraints.

\subsection{Structured Graph Learning Via Laplacian Spectral Constraints }

To enforce spectral constraints on the Laplacian matrix $\bTheta$ (i.e., $\mathcal{T}(\bTheta)=\bTheta$ in \eqref{CGL_L:0:a}), we consider the following optimization problem:
\begin{align}\label{CGL_L:0}
\begin{array}{ll}
\underset{\Theta,{\blambda}, {U}}{\text{maximize}} &
\begin{array}{c}
\hspace{.75cm}\log\;\text{gdet}(\Theta)-\text{tr}\big({\Theta S}\big) -\alpha h(\Theta),
\end{array}\\
\text{subject to} & \begin{array}[t]{l}
\hspace{.75cm}\Theta \in \mathcal{S}_{\Theta}, \ \Theta={U}{\text{Diag}(\blambda)} {U}^T, \ {\blambda} \in \mathcal{S}_{{\lambda}}, \ {U}^T{U}={I}, 
\end{array}
\end{array}
\end{align}
where $\bTheta$ is the desired Laplacian matrix and $\bTheta$ admits the decomposition $\bTheta={U}\text{Diag}(\blambda) {U}^T$, $\text{Diag}(\blambda)\in \Rn^{p\times p}$ is a diagonal matrix containing $\blambda=\{\lambda_i\}_{i=1}^p$ on its diagonal with $\blambda\in \mathcal{S}_{{\lambda}}$, and $ U\in \Rn^{p\times p}$ is a matrix satisfying ${U}^T{U}={I}$. We enforce $\bTheta$ to be a Laplacian matrix by the constraint $\Theta \in \mathcal{S}_{\Theta}$, while we incorporate some specific spectral constraints on $\bTheta$ by forcing $\bTheta={U}\text{Diag}(\blambda) {U}^T$, with $\mathcal{S}_{{\lambda}}$ containing priori spectral information on the desired graph structure.

Next, we will introduce various choices of $\S_{\lambda}$ that will enable \eqref{CGL_L:0} to learn numerous popular graph structures.

\subsubsection{$k$-component graph}

A graph is said to be $k-$component connected if its vertex set can be partitioned into $k$ disjoint subsets $\Vc=\cup_{i=1}^k\Vc_i$ such that any two nodes belonging to different subsets are not connected by an edge. Any edge in edge set $\E_i \subset \E$ have end points in $\Vc_i$, and no edge connect two different components. The $k-$component structural property of a graph is directly encoded in the eigenvalues of its Laplacian matrix. The multiplicity of zero eigenvalue of a Laplacian matrix gives the number of connected components of a graph $\mathcal{G}$.
\begin{thm}\citep{chung1997spectral}
	The eigenvalues of any Laplacian matrix can be expressed as:
	\begin{align}\label{Eig_set_K-connected}
	\mathcal{S}_\lambda= \{\{ \lambda_j=0\}_{j=1}^k, \ c_1\leq \lambda_{k+1}\leq \ldots \leq \lambda_p \leq c_2\}
	\end{align}	
	where $k\geq 1$ denotes the number of connected components in the graph, and $c_1, c_2>0$ are some constants that depend on the number of edges and their weights \citep[see][]{spielman2011spectral}.
	
\end{thm}

\begin{figure}[H]
	\begin{centering}
		\includegraphics[height=8cm]{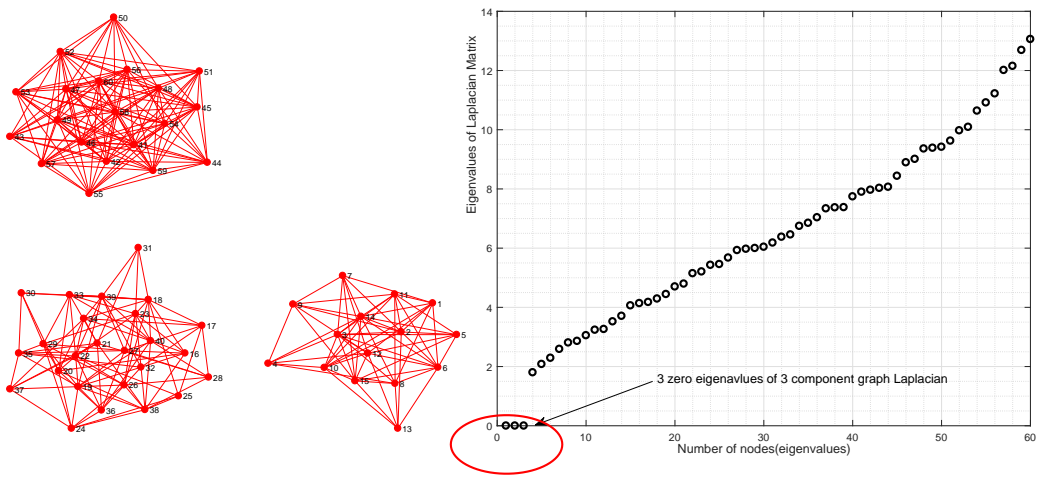}
		\par\end{centering}
	\caption{\label{fig:k_comp_plot} $3$-component graph and its Laplacian matrix eigenvalues: three zero eigenvalues correspond to three components in the graph. }
\end{figure}

 Figure \ref{fig:k_comp_plot} depicts a $k-$component graph and its Laplacian eigenvalues with $k$=3 connected components and zero eigenvalues.

\subsubsection{Connected sparse graph}A sparse graph is simply a graph with not many connections among the nodes. Often, making a graph highly sparse can split the graph into several
	disconnected components, which many times is undesirable (Sundin et al., 2017; Hassan-Moghaddam et al., 2016). The
	existing formulation cannot ensure both sparsity and connectedness, and there always exists a trade-off between the
	two properties. Within the formulation \eqref{CGL_L:0} we can achieve sparsity and connectedness by using the following spectral constraint:	
	\begin{align}\label{lap-eig-1}
	\mathcal{S}_\lambda= \{\lambda_1=0, c_1\leq \lambda_{2}\leq \ldots \leq \lambda_p \leq c_2\}
	\end{align}
	with a proper choice of $c_1>0, c_2>0$. 

%A $k$-component graph is very useful for clustering (\citep{von2007tutorial}), for constructing a new multi-resolution analysis of graph signals \citep{tremblay2016subgraph}, for spectral clustering, and for scaling inverse covariance estimation algorithms \citep{hsieh2012divide}.
%
%\col{Regular graphs are useful \citep{hoory2006expander,chow2016expander} in pure and applied mathematics, with several applications in complexity theory, design of robust computer networks, distributed optimization, error-correcting codes, for construction of well connected graphs \citep{ghosh2006growing}, and for construction of fastest mixing graphs \citep{boyd2004fastest,boyd2005mixing}. }

\subsubsection{ $d-$regular graph}
All the nodes of a $d$-regular graph have the same weighted degree ($w_i=d, \; \forall i=1,2,\ldots,p$), where weighted degree is defined as $w_i= \sum_{j\in \mathcal{N}_i}w_{ij} $, which implies:
\begin{align}
 \bTheta=d \I-\bTheta_A,\;\; \text{diag}(\bTheta)=d\b1,\;\; \bTheta_A\b1=d\b1. \nonumber
\end{align}
Within the above formulation \eqref{CGL_L_A} a $d-$regular structure on the matrix $\Theta$ can be enforced by including the following constraints
\begin{align}\label{reg1}
\mathcal{S}_\lambda= \{\lambda_1=0, c_1\leq \lambda_{2}\leq \cdots \leq \lambda_p \leq c_2\},\; \text{diag}(\bTheta)=d\b1.
\end{align}
%where $k$ denotes the number of desired component. For $k=1$ it renders a connected graph and for $k>1$ it fetches

\subsubsection{ $k-$component $d-$regular graph}

A $k-$component regular graph, also known as clustered regular graph is useful in providing improved perceptual grouping \citep{kim2009clustering} for clustering applications. Within the above formulation \eqref{CGL_L_A} we can enforce this structure by including the following constraints
\begin{align}\label{regk}
	\mathcal{S}_\lambda= \{\{ \lambda_j=0\}_{j=1}^k, \ c_1\leq \lambda_{k+1}\leq \cdots \leq \lambda_p \leq c_2\},\; \text{diag}(\bTheta)=d\b1.
\end{align}

%\subsubsection{$k-$component $d$-regular graph}

 %For example, in spectral sparsification, a sparse approximation of the given graph is done by learning graph iscorresponding 

\subsubsection{Cospectral graphs}

	In many applications, it is motivated to learn $\Theta$ with specific eigenvalues which is also known as cospectral graph learning \citep{godsil1982constructing}. One example is spectral sparsification of graphs \citep[see][]{spielman2011spectral,loukas2018spectrally} which aims to learn a graph $\Theta$ to approximate a given graph $\bar{\Theta}$, while $\Theta$ is sparse and its eigenvalues $\lambda_i$ satisfy $\lambda_i = f(\bar{\lambda}_i)$, where $\{\bar{\lambda_i}\}_{i=1}^p$ are the eigenvalues of the given graph $\bar{\Theta}$ and $f$ is some specific function. Therefore, for cospectral graph learning, we introduce the following constraint
		\begin{align}\label{lap-eig-2}
	\mathcal{S}_\lambda= \{\lambda_i= f(\bar{\lambda}_i), \quad \forall i \in [1,p] \}.
	\end{align}

\subsection{Structured Graph Learning Via Adjacency Spectral Constraints }

To enforce spectral constraints on adjacency matrix $\bTheta_A$ (i.e., $\mathcal{T}(\bTheta)=\bTheta_A$ in \eqref{CGL_L:0:a}), we introduce the following optimization problem:
\begin{align}\label{CGL_A}
\begin{array}{ll}
\underset{{\Theta},{{\bpsi}},{{V}}}{\text{maximize}} &
\begin{array}{c}
\hspace{.75cm}\log\; \text{gdet} (\Theta)-\text{tr}\big({\Theta S}\big)-\alpha h(\Theta),
\end{array}\\
\text{subject to} & \begin{array}[t]{l}
\hspace{.75cm}\Theta \in \mathcal{S}_{\Theta}, \ \Theta_A={V} {\Di(\bpsi)} {V}^T, \ 
{\bpsi} \in \mathcal{S}_{{\psi}},\ {V}^T{V}={I}, 
\end{array}
\end{array}
\end{align}
where $\bTheta$ is the desired Laplacian matrix, $\bTheta_A$ is the corresponding adjacency matrix which admits the decomposition $\bTheta_A={V}\Di({\bpsi}) {V}^T$ with ${\bpsi} \in \mathcal{S}_{{\psi}}$ and ${V}^T{V}={I}$. We enforce $\bTheta$ to be a Laplacian matrix by the constraint $\Theta \in \mathcal{S}_{\Theta}$, while we incorporate some specific spectral constraints on its adjacency matrix $\bTheta_A$ by forcing $\bTheta_A={V}\Di({\bpsi}) {V}^T$, with $\mathcal{S}_{{\psi}}$ containing priori spectral information of the desired graph structure.

Next, we will introduce various choices of $\S_{\psi}$ that will enable \eqref{CGL_A} to learn bipartite graph structures.

\subsubsection{General bipartite graph}
 
A graph is said to be bipartite if its vertex set can be partitioned into two disjoint subsets $\Vc=\cup_{i=1}^2\Vc_i$ such that no two points belonging to the same subset are connected by an edge \citep{zha2001bipartite}, i.e. for each $(l,m)\in \Vc_i\times \Vc_i$ then $(l,m)\notin \E$. Spectral graph theory states that a graph is bipartite if and only if the spectrum of the associated adjacency matrix is symmetric about the origin \citep[Ch.5]{van2010graph} \citep{mohar1997some}.
\begin{thm}\citep[see][]{mohar1997some}\label{adj:bip}
	A graph is bipartite if and only if the spectrum of the associated adjacency matrix is symmetric about the origin 
	 \begin{align}\label{bipartite-eig}
	\mathcal{S}_{{\psi}}= \{\psi_i=-\psi_{p-i+1},\;& \forall i\;=1,\ldots, p \\
	& \psi_1 \geq \psi_2\geq\cdots\geq\psi_{p}\nonumber
	\big\}.
	\end{align}
\end{thm}

\subsubsection{Connected bipartite graph}

\begin{figure}[H]
	\begin{centering}
		\includegraphics[height=8cm]{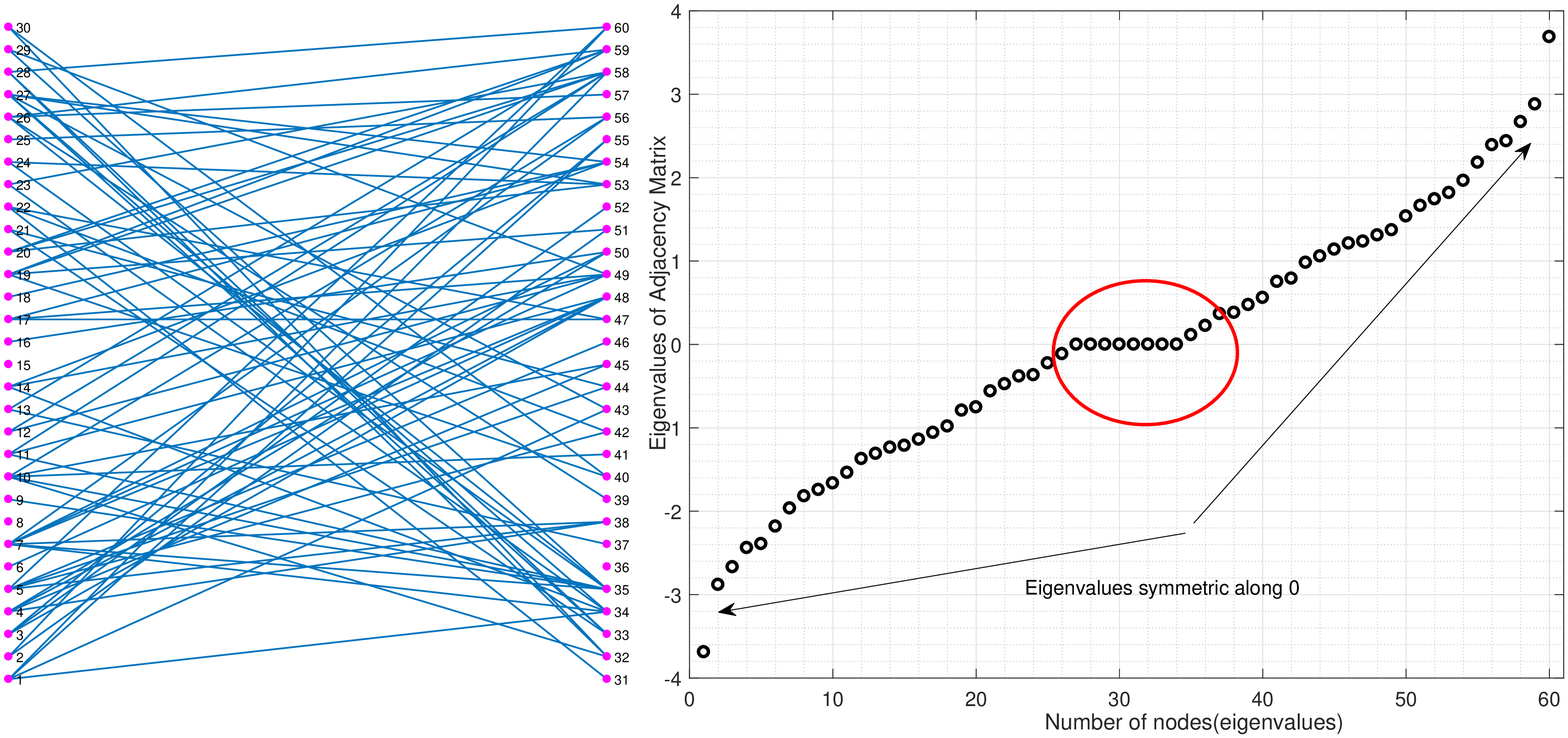}
		\par\end{centering}
	\centering{}\caption{\label{fig:bipartite_plot} Bipartite graph adjacency matrix and its eigenvalues: eigenvalues are symmetric around 0.}
\end{figure}

The Perron-Frobenius theorem states that if a graph is connected, then the largest eigenvalue $\psi_p$ of its adjacency matrix ${A}$ has multiplicity 1 \citep{mohar1997some}. Thus, a connected bipartite graph can be learned by including additional constraint on the multiplicity to be one on the largest and smallest eigenvalues, i.e. $\psi_1, \psi_p$ are not repeated. Figure \ref {fig:bipartite_plot} shows a connected bipartite graph and its adjacency symmetric eigenvalues.

\begin{thm}\citep[see][]{mohar1997some}\label{adj:bip-2}
	A graph is connected bipartite graph if and only if the spectrum of the associated adjacency matrix is symmetric about the origin with non-repeated extreme eigenvalues 
	\begin{align}\label{bipartite-eig-2}
	\mathcal{S}_{{\psi}}= \{\psi_i=-\psi_{p-i+1},\;& \forall\; i\;=1,\cdots, p \\
	& \psi_1 > \psi_2\geq\cdots\geq \psi_{p-1}>\psi_{p}\nonumber
	\big\}.
	\end{align}
\end{thm} 

\subsection{Structured Graph Learning Via Joint Laplacian and Adjacency Spectral Constraints }\vspace{-0.2cm}

To enforce spectral constraints on Laplacian matrix $\bTheta$ and adjacency matrix $\bTheta_A$, we introduce the following optimization problem:
\begin{align}\label{CGL_L_A}
\begin{array}{ll}
\underset{{\Theta},{\blambda},{\bpsi},{{U}}, {{V}}}{\text{maximize}} &
\begin{array}{c}
\hspace{.75cm}\log \; \text{gdet}(\Theta)-\text{tr}\big({\Theta S}\big)-\alpha h(\Theta),
\end{array}\\
\text{subject to} & \begin{array}[t]{l}
\hspace{.75cm}\Theta \in \mathcal{S}_{\Theta}, \ \Theta={U}{\Di(\blambda)} {U}^T, \ \Theta_A={V}\Di({\bpsi}) {V}^T,\\
\hspace{.75cm} {\blambda} \in \mathcal{S}_{{\lambda}},\ {U}^T{U}={I},\ {\bpsi} \in \mathcal{S}_{{\psi}},\ \ {V}^T{V}={I}.
\end{array}
\end{array}
\end{align}
where $\bTheta$ is the desired Laplacian matrix which admits the decomposition $\Theta={U}{\Di(\blambda)} {U}^T,$ with
${\blambda} \in \mathcal{S}_{{\lambda}}$, ${U}^T{U}={I}$, and $\bTheta_A$ is the corresponding adjacency matrix which admits the decomposition $\bTheta_A={V}{\Di(\bpsi)} {V}^T$ with ${\bpsi} \in \mathcal{S}_{{\psi}}$ and ${V}^T{V}={I}$. Observe that the above formulation learns a graph Laplacian matrix $\Theta$ with a specific structure by enforcing the spectral constraints on the adjacency and Laplacian matrices simultaneously. Next, we will introduce various choices of $\S_{\lambda}$ and $\S_{\psi}$ that will enable \eqref{CGL_L_A} to learn non-trivial complex graph popular graph structures.

 \subsubsection{$k-$component bipartite graph}
 A $k-$component bipartite graph, also known as bipartite graph clustering, has a significant relevance in many machine learning and financial applications \citep{zha2001bipartite}. Recall that the bipartite structure can be enforced by utilizing the adjacency eigenvalues property (i.e., the constraints in \eqref{bipartite-eig}) and $k-$component structure can be enforced by the Laplacian eigenvalues (i.e., the zero eigenvalues with multiplicity $k$). These two disparate requirements can be simultaneously imposed in the current formulation \eqref{CGL_L_A}, by choosing:
 \begin{align}\label{k-comp-bipartite}
 &	\mathcal{S}_\lambda= \{\{ \lambda_j=0\}_{j=1}^k, \ c_1\leq \lambda_{k+1}\leq \ldots \leq \lambda_p \leq c_2\}\\
 & \mathcal{S}_\psi= \{\psi_i: \psi_i=-\psi_{p-i+1}, \forall \; i=1,\cdots,p\}.\nonumber
 \end{align}

\subsubsection{$k-$component regular bipartite graph}

The eigenvalue property of $d-$regular graph relates the eigenvalues of its adjacency matrix and Laplacian matrix, which is summarized in the following lemma.
\begin{thm}\citep{mohar1997some}
	Collecting the Laplacian eigenvalues in increasing order ($\{\lambda_j\uparrow\}_{j=1}^p$) and the adjacency eigenvalues in decreasing order ($\{\psi_i\downarrow\}_{i=1}^p$), then the eigenvalue pairs for a $d$-regular graph are related as follows:
	\begin{align}\label{d-reg-eig}
	\lambda_i=d-\psi_i,\; \forall\; i=1,\cdots,p.
	\end{align}
\end{thm}

A $k-$component regular bipartite structure can be enforced by utilizing the adjacency eigenvalues property (for bipartite structure), Laplacian eigenvalues (for $k-$component structure) along with the joint spectral properties for the regular graph structure:
\begin{align}
&	\mathcal{S}_\lambda= \{\{ \lambda_j=0\}_{j=1}^k, \ c_1\leq \lambda_{k+1}\leq \ldots \leq \lambda_p \leq c_2\}\\
& \mathcal{S}_\psi= \{\psi_i: \psi_i=d- \lambda_i;\; \psi_i=-\psi_{p-i+1}, \forall \; i=1,\cdots,p\},\nonumber
\end{align}

\subsection{Block Successive Upper-bound Minimization algorithm}\label{bsum-section}
The resulting optimization formulations presented in \eqref{CGL_L:0}, \eqref{CGL_A}, and \eqref{CGL_L_A} are still complicated. The aim here is to develop efficient optimization methods with low computational complexity based on the BSUM and majorization-minimization framework \citep{razaviyayn2013unified,SunBabPal2017-MM}. To begin with, we present a general schematic of the BSUM optimization framework 
\begin{align}\label{bsum}
\begin{array}{ll}
\underset{\x}	{\text{minimize}} &
\begin{array}{c}
f(\x)
\end{array}\\
\text{subject to} & \begin{array}[t]{l}
\x \in \mathcal{X}, 
\end{array}
\end{array}
\end{align}
where the optimization variable $\x$ is partitioned into $m$ blocks as $\x=(\x_1,\x_2,\cdots,\x_m)$, with $\x_i \in \mathcal{X}_i$, $\mathcal{X}=\prod_{i=1}^{m}\mathcal{X}_i$ is a closed convex set, and $f: \x\rightarrow \Rn$ is a continuous function. 
At the $t-$th iteration, each block $\x_i$ is updated in a cyclic order by solving the following: 
\begin{align}\label{bsum-maj}
\begin{array}{ll}
\underset{\x_i}	{\text{minimize}} &
\begin{array}{c}
g_i(\x_i\vert \x_1^t, \cdots,\x_{i-1}^t,\x_{i+1}^{t-1},\cdots, \x_m^{t-1}),
\end{array}\\
\text{subject to} & \begin{array}[t]{l}
\x_i \in \mathcal{X}_i,
\end{array}
\end{array}
\end{align}
where $g_i(\x_i\vert\y_i^t)$ with $\y_i^t:=(\x_1^t, \cdots, \x_{i-1}^{t}, \x_{i}^{t-1}, \x_{i+1}^{t-1},\cdots, \x_m^{t-1})$ is a majorization function of $f(\x)$ at $\y_i^t $ satisfying
\begin{subequations}\label{bsum-maj-0}
	\begin{align}
&g_i(\x_i\vert \y_i^t)\; \text{is continuous in } \; (\x_i, \y_i^t), \; \forall\; i, \label{bsum-maj-4}\\
&g_i(\x^t_i\vert \y_i^t)=	f( \x_1^t, \cdots,\x_{i-1}^t,\x_{i}^t,\x_{i+1}^{t-1},\cdots, \x_m^{t-1}),\label{bsum-maj-1}\\
&g_i(\x_i\vert \y_i^t)\geq 	f( \x_1^t, \cdots,\x_{i-1}^t,\x_{i},\x_{i+1}^{t-1},\cdots, \x_m^{t-1}),\;\forall \; \x_i\in \cX_i,\forall \; \y_i\in \cX, \forall \; i,\label{bsum-maj-2} \\
&{ g'_i(\x_i;\b d_i\vert \y_i^t)\vert_{\x_i=\x_i^t} = f'( \x_1^t, \cdots,\x_{i-1}^t,\x_{i},\x_{i+1}^{t-1},\cdots, \x_m^{t-1};\b d)},\nonumber\\
& \hspace{3cm}{ \forall\;\; \b d=(\b 0, \cdots, \b d_i, \cdots, \b 0)\;\text{such that}\; \x_i^t+\b d_i \in \cX_i,\; \forall\; i,} \label{bsum-maj-3}
	\end{align} 
\end{subequations}
where $ f'(\x;\b d)$ {stands for the directional derivative at $\x$ along $ \b d$} \citep{razaviyayn2013unified}. In summary, the framework is based on a sequential inexact block coordinate approach, which updates the variable in one block keeping the other blocks fixed. If the surrogate functions $g_i$ is properly chosen, then the solution to \eqref{bsum-maj} could be easier to obtain than solving \eqref{bsum} directly.

%\vert_{\x_i=\x_{i}^{t}}  \vert_{\x_i=\x_{i}^{t}}

\section{Structured Graph Learning Via Laplacian Spectral Constraints (\textsf{SGL})}\label{sec:lap}
{In this section, we develop a BSUM-based algorithm for \textbf{S}tructured \textbf{G}raph learning via \textbf{L}aplacian spectral constraints (\textsf{SGL}). In particular}, we consider solving \eqref{CGL_L:0} under $k-$component Laplacian spectral constraints \eqref{Eig_set_K-connected}. To enforce sparsity we use the $\ell_1-$regularization function (i.e., $h(\Theta):=\norm{\bTheta}_1$). Next observing that the sign of $\bTheta$ is fixed by the constraints $\bTheta_{ij} \leq 0 \, {\rm for} \, i\neq j$ and $\bTheta_{ij} \geq 0 \, {\rm for} \, i= j$, the regularization term $\alpha\norm{\bTheta}_1$ can be written by $\tr{\bTheta \H}$, where $\H = \alpha (2\I - \bone\bone^T)$, problem \eqref{CGL_L:0} becomes
\begin{align}\label{CGL_L}
\begin{array}{ll}
\underset{{\bTheta},{ \blambda}, \U}{\text{minimize}} &
\begin{array}{c}
- \log \;\text{gdet} (\bTheta)+\tr{\bTheta \K},
\end{array}\\
\text{subject to} & \begin{array}[t]{l}
{\bTheta} \in \S_{\bTheta}, \ \bTheta=\U \Di(\blambda) \U^T, \ \blambda\in \S_{\lambda}, \ \U^T\U=\I, 
\end{array}
\end{array}
\end{align}
where $\K=S+ \H$. The resulting problem is complicated and intractable in the current form due to i) Laplacian structural constraints $\S_{\bTheta}$, ii) coupling variables $\bTheta=\U \Di(\blambda) \U^T$, and iii) generalized determinant on $\bTheta$. In order to derive a more feasible formulation, we first introduce a linear operator $\L$ which transforms the Laplacian structural constraints to simple algebraic constraints and then relax the eigen-decomposition expression into the objective function. 

\subsection{ Graph Laplacian operator $\L$}\label{L-opeartor}
The Laplacian matrix $\bTheta$ belonging to $\S_{\bTheta}$ satisfies i) $\Theta_{ij} =\Theta_{ji} \leq 0 $, ii) $\bTheta\bone=\bzero$, implying the target matrix is symmetric with degrees of freedom of ${\bTheta}$ equal to $p(p-1)/2$. Therefore, we introduce a linear operator $\L$ that transforms a non-negative vector $\w \in \mathbb{R}_{+}^ {p(p-1)/2}$ into the matrix $\L \w \in \Rn^{p \times p}$ that satisfies the Laplacian constraints ($[\L \w]_{ij} =[\L \w]_{ji}, {\rm for} \, i\neq j $ and $[\L \w] \bone =\bzero$).

\begin{def1}\label{operator-L}
The linear operator $\L:\w \in \mathbb{R}_{+}^ {\frac{p(p-1)}{2}}\rightarrow \L\w \in \Rn^{p \times p}$ is defined as
\[
[\L\w]_{ij} =
\begin{cases}
-w_{i+d_j}\; & \; \;i >j,\\
[\L\w]_{ji} \; & \; \;i<j,\\
\sum_{i\neq j}[\L\w]_{ij}\; &\;\; i=j,
\end{cases}
\]\label{lap:operator}
where $d_j=-j+\frac{j-1}{2}(2p-j).$ 
\end{def1}

We derive the adjoint operator $\L^*$ of $\L$ by making $\L^*$ satisfy $\langle \L\w,\Y \rangle=\langle \w,\L^*\Y \rangle$.
\begin{Lem1}\label{adjoint-L}
The adjoint operator $\L^*: \Y \in \Rn^{p\times p} \mapsto \L^* \Y \in \Rn^{\frac{p(p-1)}{2}}$ is defined by
	\begin{equation*}
	[\L ^* \Y]_k=y_{i,i}-y_{i,j}-y_{j,i}+y_{j,j},\; \; k=i-j+\frac{j-1}{2}(2p-j),
	\end{equation*}
	where $i,j \in \mathbb{Z}^+$ satisfy $ k=i-j+\frac{j-1}{2}(2p-j)$ and $i>j$.
\end{Lem1}
%k=(j-1)n+i-j-\sum_{s=0}^{j-1}s %%%% 	for any $i>j$ with $i \in \{2,3,\cdots,n\}$ and $j \in \{1,2,\cdots,n-1\}$. 

A toy example is given to illustrate the operators $\L$ and $\L^*$ more clearly. Consider a weight vector $\w=[w_1,w_2,w_3,w_4,w_5,w_6]^T$. The Laplacian operator $\L$ on $\w$ gives
{
	\begin{equation}
	\L \w= \left[
	\begin{array}{cccc}
	\sum_{i=1,2,3} w_i & -w_1 & -w_2 & -w_3\\
	-w_1 & \sum_{i=1,4,5} w_i & -w_4 & -w_5\\
	-w_2 & -w_4 & \sum_{i=2,4,6} w_i & -w_6\\
	-w_3 & -w_5 & -w_6 & \sum_{i=3,5,6} w_i
	
	\end{array}
	\right].
	\end{equation}
}

The operation of $\L^*$ on a $4 \times 4$ symmetric matrix $\Y$ returns a vector 
\begin{equation}
\L^* \Y= \left[
\begin{array}{c}
y_{11}-y_{21}-y_{12}+y_{22}\\
y_{11}-y_{31}-y_{13}+y_{33}\\
y_{11}-y_{41}-y_{14}+y_{44}\\
y_{22}-y_{32}-y_{23}+y_{33}\\
y_{22}-y_{42}-y_{24}+y_{44}\\
y_{33}-y_{43}-y_{34}+y_{44}
\end{array}
\right].
\end{equation}
By the definition of $\L$, we have Lemma \ref{lem-lap-norm}.

\begin{Lem1}\label{lem-lap-norm}
	The operator norm $\Vert \L \Vert_2 $ is $\sqrt{2p}$, where $\Vert \L \Vert_2 \doteq \sup_{\norm{\x}=1}\norm{\L \x}_F$ with $\x \in \Rn ^{\frac{p(p-1)}{2}}$. 
\end{Lem1}
\noindent
\begin{proof}
Follows from the definitions of $\L$ and $\L^*$: see Appendix \ref{proof:lem-lap-norm} for detailed proof.
\end{proof}

% See Appendix \ref{proof:lem-lap-norm}. Further, $\Vert \L \Vert_2 = \sqrt{2p}$ implies that $\lambda_{\max}(\L^* \L) = 2p$.

We have introduced the operator $\L$ that helps to transform the complicated structural matrix variable $\bTheta$ to a simple vector variable $\w$. The linear operator $\L$ is an important component of the \textsf{SGL} framework.

\subsection{ \textsf{SGL} algorithm}

To solve \eqref{CGL_L}, we represent the Laplacian matrix $\bTheta \in \S_{\bTheta}$ as $\L \w$ and then develop an algorithm based on quadratic methods \citep{nikolova2005analysis, ying2018vandermonde}. We introduce the term $\frac{\beta}{2}\| \L \w -{\U} \Di(\blambda) {\U}^T\|_F^2$ to keep $\L \w$ close to ${\U} \Di(\blambda) {\U}^T$ instead of exactly solving the constraint $\L \w ={\U} \Di(\blambda) {\U}^T$, where $\beta>0$. Note that this relaxation can be made tight by choosing $\beta$ sufficient large or iteratively increasing $\beta$. Now, the original problem can be formulated as 
\begin{align}\label{CGL_L_cost}
\begin{array}{ll}
\underset{{\w},{ \blambda},{\U}}{\text{minimize}} &
\begin{array}{c}
- \log \text{gdet} (\Di(\blambda))+\tr{\K \L \w}+\frac{\beta}{2}\| \L \w -\U \Di(\blambda) \U^T\|_F^2,
\end{array}\\
\text{subject to} & \begin{array}[t]{l}
\w \geq 0, \ \blambda\in \S_{\lambda},\ \U^T\U=\I,
\end{array}
\end{array}
\end{align}
where $\w \geq 0$ means each entry of $\w$ is non-negative. When solving \eqref{CGL_L_cost} to learn the $k-$component graph structure with the constraints in \eqref{Eig_set_K-connected}, the first $k$ zero eigenvalues as well as the corresponding eigenvectors can be dropped from the optimization formulation. Now the $\blambda$ only contains $q=p-k$ non-zero eigenvalues in increasing order $\{\lambda_{j}\}_{j=k+1}^p$, then we can replace generalized determinant with determinant on $\Di(\blambda)$ in \eqref{CGL_L_cost}. ${\U }\in \mathbb{R}^{p \times q}$ contains the eigenvectors corresponding to the non-zero eigenvalues in the same order, and the orthogonality constraints on $\U$ becomes $\U^T\U=\I_q$. The non-zero eigenvalues are ordered and lie in the given set, 
\begin{align}\label{Eig_set_K-connected-}
\mathcal{S}_{{\lambda}}= \{ \ c_1\leq \lambda_{k+1}\leq \ldots \leq \lambda_p\leq c_2\}.
\end{align}
Collecting the variables in three block as $\x=\left( \w\in \Rn^{p(p-1)/2}, \blambda\in \Rn^{q}, \U\in \Rn^{p\times q}\right)$, we develop a BSUM-based algorithm which updates only one variable each time with the other variables fixed. 

\subsubsection{Update of $\w$}\label{sec-w-up}

Treating $\w$ as a variable with $\U$ and $\blambda$ fixed, and ignoring the terms independent of $\w$, we have the following sub-problem:
\begin{align}\label{sub-x}
\underset{\w \geq 0}{\text{minimize}}\qquad \tr{\K \L \w} +\frac{\beta}{2}\| \L \w -\U \Di(\blambda) {\U}^T\|_F^2.
\end{align}
The problem \eqref{sub-x} can be written as a non-negative quadratic problem,
\begin{align}\label{quad-x}
\underset{\w \geq 0}{\text{minimize}}\qquad f(\w)=\frac{1}{2}\norm{\L\w}^2_F - {\cb}^T\w,
\end{align}
where $\cb=\L^*(\U \Di(\blambda) {\U}^T- \beta^{-1}\K)$.

\begin{Lem1}\label{lem1}
The sub-problem \eqref{quad-x} is a strictly convex optimization problem.
\end{Lem1}
\begin{proof}
 From the definition of operator $\L$ and the property of its adjoint $\L^*$, we have 
\begin{align}\label{Porf1}
\| \L \w\|_F^2=\langle \L \w, \L \w \rangle=\langle \w, \L^*\L \w \rangle= \w^T \L^*\L \w >0, \;\; \forall\; \w \neq \bzero.
\end{align}
The above result implies that $f(\w)$ is a strictly convex function. Together with the fact that the non-negativity set is convex, we conclude the sub-problem \eqref{quad-x} is strictly convex. But, it is not possible here to derive a closed-form solution due to the non-negativity constraint ($\w\geq0$), and thus we derive a majorijation function.
\end{proof}

\begin{Lem1}\label{lem2} 
	The function $f(\w)$ in \eqref{quad-x} is majorized at $\w^{t}$ by the function
	\begin{align}
	g( \w|\w^{t} )&=f( \w^t) + (\w - \w^{t} )^T \nabla f ( \w^{t}) +\frac{L_1}{2}\norm{\w - \w^{t}}^2, \label{sur-lap}
	\end{align}
where $\w^t$ is the update from previous iteration and $L_1=\norm{\L}_2^2=2p$ (see Lemma \ref{lem-lap-norm}). 
\end{Lem1}

It is easy to check the conditions \eqref{bsum-maj-0} for the majorization function \citep[See more details in][]{SunBabPal2017-MM,song2015sparse} and we ignore the proof here. Note that the majorization function as $g( \w|\w^{t} )$ in \eqref{sur-lap} is in accordance with the requirement of the majorization as in \eqref{bsum-maj-1}, because in the problem \eqref{quad-x}, $\w^t$ and the other coordinates ($\blambda^t, U^t$) are fixed. For notational brevity, we present the majorization function as $g( \w|\w^{t} )$ instead of $g(\w|\w^{t}, U^t, \blambda^t )$.

%\begin{Lem1}
After ignoring the constant terms in \eqref{sur-lap}, the majorized problem of \eqref{quad-x} at $\w^t$ is given by
\begin{align}
\underset{\w \geq 0}{\text{minimize}}\qquad g(\w|\w^{t})=\frac{1}{2}{\w}^T\w - {\bm a}^T\w, \label{majr}
\end{align}
where ${a}= \w^t-\frac{1}{L_1}\nabla f(\w^t)$ and $\nabla f(\w^t)=\L^*(\L\w^t)-\b c$. 
%\end{Lem1}

\begin{Lem1}\label{lem-kkt-w}
From the KKT optimality conditions we can easily obtain the optimal solution to \eqref{majr} as
\begin{align}\label{w_up:algo-1}
\w^{t+1}= \left( \w^{t}-\frac{1}{L_1}\nabla f(\w^{t})\right)^+,
\end{align}
where $(x)^+:=\max(x,0)$.
\end{Lem1}

\subsubsection{Update of $\U$}\label{sub:algo-w-up}
Treating $\U$ as a variable block, and fixing for $\w$ and $\blambda$, we obtain the following sub-problem:
\begin{align}\label{sub-U}
\underset{\U}{\text{minimize}} \qquad \frac{\beta}{2}\| \L \w -\U \Di(\blambda) \U^T \|_F^2, \quad \text{subject to} \ \U^T\U=\I_q.
\end{align}
The equivalent problem is reformulated as follows
\begin{align}\label{sub-U1}
\underset{\U}{\text{maximize}} \qquad {\rm tr}(\U^T \L \w\U \Di(\blambda) ), \quad \text{subject to} \quad \ \U^T\U=\I_q.
\end{align}
The problem \eqref{sub-U1} is an optimization on the orthogonal Stiefel manifold $\text{St}(p,q)=\{{\U }\in \mathbb{R}^{p \times q}: \U^T\U=\I_q\} $. From \citep{absil2009optimization,benidis2016orthogonal} the maximizer of \eqref{sub-U1} is the eigenvectors of $\L \w$ (suitably ordered).
\begin{Lem1}
From the KKT optimality conditions the solution	to \eqref{sub-U1} is given by
	\begin{align}\label{algo1:U-update}
	\U=\textsf{eigenvectors}(\L\w)[k+1:p]
	\end{align}
that is, the $p-k$ principal eigenvectors of the matrix $\L \w$ in the increasing order of the eigenvalue magnitude \citep{absil2009optimization,benidis2016orthogonal}. 
\end{Lem1}

\subsubsection{Update for ${\lambda}$}\label{sec-lambda-algo1}
We obtain the following sub-problem for the $\blambda$ update
\begin{align}\label{sub_lambda}
\underset{ \blambda \in \S_{\lambda}}{\text{minimize}} \quad & - \log \det \Di(\blambda)+\frac{\beta}{2}\| \L \w -\U \Di(\blambda) \U^T\|_F^2.
\end{align}
The optimization \eqref{sub_lambda} can be rewritten as
\begin{align}\label{sub_lambda1}
\underset{ \blambda \in \S_{\lambda}}{\text{minimize}} \quad & - \log \det\Di(\blambda)+\frac{\beta}{2}\| \U^T (\L \w)\U -\Di(\blambda)\|_F^2.
\end{align}

With slight abuse of notation and for ease of exposition, we denote the indices for the non-zero eigenvalues $\lambda_i$ in \eqref{Eig_set_K-connected-} from $1$ to $q=p-k$ instead of $k+1$ to $p$. The problem \eqref{sub_lambda1} can be further written as
\begin{align}\label{sub-lambda}
\underset{c_1 \leq \lambda_1 \leq \cdots \leq \lambda_q \leq c_2}{\text{minimize}} \quad & - \sum_{i=1}^{q} \log \lambda_i+\frac{\beta}{2}\| \blambda -\b d \|_2^2,
\end{align}
where $\blambda=[\lambda_1,\cdots,\lambda_q]^T$ and $ \b d=[d_1,\cdots, d_q]^T$ with $d_i$ the $i$-th diagonal element of ${\rm Diag}({\U}^T (\L \w){ \U})$. We derive a computationally efficient method to solve \eqref{sub-lambda} from KKT optimality conditions. The update rule for $\blambda$ follows an iterative procedure summarized in Algorithm \ref{algo-lambda}.
The sub-problem \eqref{sub-lambda} is a convex optimization problem. One can solve the convex problem \eqref{sub-lambda} with a solver (e.g., \textsf{CVX}) but we can do it more efficiently with our algorithm for large scale problems.

\begin{Lem1}\label{alg1}
	The iterative-update procedure summarized in Algorithm \ref{algo-lambda} converges to the KKT point of Problem \eqref{sub-lambda}.
\end{Lem1}

\begin{proof}
	Please refer to the Appendix \ref{apendix-lambda} for the detailed proof.
\end{proof}

\begin{algorithm}
	\caption{ Update rule for $\lambda_1,\cdots,\lambda_q$ \label{algo-lambda} }
	\begin{algorithmic}[1]
		\STATE \textbf{Compute:} $\lambda_i=(d_i+\sqrt{d_i^2+4/\beta})/2$ for $1\leq i\leq q$.
		\IF {$\blambda$ satisfies $c_1 \leq \lambda_1 \leq \cdots \leq \lambda_q\leq c_2$}
		\STATE RETURN $\lambda_1,\cdots,\lambda_q$.
		\ENDIF
		\WHILE {$\blambda$ violates $c_1 \leq \lambda_1 \leq \cdots \leq \lambda_q\leq c_2$}
		
		\STATE \textbf{check situation 1:} 
		\STATE \textbf{if} {$c_1\geq\lambda_1\geq\cdots\geq\lambda_r$ with at least one inequality strict and $r\geq 1$}, 
		\STATE \quad Set $\lambda_1=\cdots=\lambda_r=c_1$.
		
		\STATE \textbf{end if} 
		
		\STATE \textbf{check situation 2:} 
		\STATE \textbf{if} { $\lambda_s \geq \cdots \geq \lambda_q \geq c_2$ with at least one equality strict and $ s \leq q$},
		\STATE \quad Set $\lambda_s=\cdots=\lambda_q=c_2$.
		\STATE \textbf{end if} 
		
		\STATE \textbf{check situation 3:} 
		\STATE \textbf{if} { $\lambda_i\geq \cdots \geq \lambda_m$ with at least one equality strict and $1 \leq i \leq m \leq q $}, 
		%$\lambda_{m-1}<\lambda_m$ ($c_1 <\lambda_1$ if $j=1$) and $\lambda_m < \lambda_{m+1}$ ($\lambda_q< c_2$ if $m=q$)} 
		\begin{align}
		\text{Set}\; \lambda_i=\cdots=\lambda_m=\big(\bar{d}_{i\rightarrow m}+\sqrt{\bar{d}_{i\rightarrow m}^2+4/\beta}\big)/2,
		\text{with} \; \bar{d}_{i\rightarrow m}=\frac{1}{m-i+1}\sum\nolimits_{j=i}^md_j. \nonumber
		\end{align}
		\STATE \textbf{end if}

		\ENDWHILE
		%		\ENDFOR
		%		\STATE $\beta \leftarrow \rho\beta$
		%		\ENDWHILE
		\STATE RETURN $\lambda_1,\cdots,\lambda_q$\\
	\end{algorithmic}
\end{algorithm}

To update $\lambda_i$'s, Algorithm \ref{algo-lambda} iteratively check situations [cf. steps 6, 10 and 14] and updates the $\lambda_i$'s accordingly until $c_1 \leq \lambda_1\leq \ldots,\leq \lambda_q\leq c_2$ is satisfied. If some situation happens, then the corresponding $\lambda_i$'s need to be updated accordingly. Note that the situations are independent from each other, i.e., each $\lambda_i$ will not involve two situations simultaneously. Furthermore, $\lambda_i$'s are updated iteratively according to the above situations until all of them satisfy the KKT conditions, the maximum number of iterations is $q+1$.

\vspace*{2 ex}

\begin{rem}\label{iso-remark}
	The problem of the form \eqref{sub-lambda} is popularly known as a regularized isotonic regression problem. The isotonic regression is a well-researched problem that has found applications in numerous domains see \citep[see][]{best1990active, lee1981quadratic,barlow1972isotonic,luss2014generalized,bartholomew2004isotonic}. 
	To the best of our knowledge, however, there does not exist any computationally efficient method comparable to the Algorithm \ref{algo-lambda}.
	The proposed algorithm can obtain a globally optimal solution within a maximum of $q+1$ iterations for the $q$-dimensional regularized isotonic regression problem, and can be potentially adapted to solve other isotonic regression problems. The computationally efficient Algorithm \ref{algo-lambda} also holds an important contribution for the isotonic regression literature.
\end{rem}

\subsubsection{\textsf{SGL} algorithm summary}
\textsf{SGL} in Algorithm \ref{algo-1} summarizes the implementation of the structured graph learning via Laplacian spectral constraints. 

\begin{algorithm}
	\caption{\textsf{SGL}} \label{algo-1} 
	\begin{algorithmic}[1]
		%		\FOR{ $t=1,\cdots,$}
		\STATE	\textbf{Input:} SCM $S$, $k, c_1, c_2,\beta$.
		\STATE	\textbf{Output:} $\hat{\bTheta}$
		\STATE $t \leftarrow 0$
		\WHILE {Stopping criteria not met}
		\STATE Update $\w^{t+1} $ as in \eqref{w_up:algo-1}.
		\STATE Update $\U^{t+1}$ as in \eqref{algo1:U-update}.
		\STATE Update $\blambda^{t+1}$ as in Algorithm \ref{algo-lambda}.
		\STATE $t\leftarrow t+1$
		\ENDWHILE
		%		\ENDFOR
		%		\STATE $\beta \leftarrow \rho\beta$
		%		\ENDWHILE
		\STATE RETURN $\hat{\bTheta}^{t+1}=\L\w^{t+1}$\\
	\end{algorithmic}
\end{algorithm}

In Algorithm \ref{algo-1}, the computationally most demanding step is the eigen-decomposition step required for the update of $\U$. Implying $O(p^3)$ as the worst-case computational complexity of the algorithm. This can further be improved by utilizing the sparse structure and the properties of the symmetric Laplacian matrix for eigen-decomposition. The most widely used \textsf{GLasso} method \citep{friedman2008sparse} has similar worst-case complexity, although the \textsf{GLasso} learns a graph without structural constraints. 
While considering specific structural requirements, the \textsf{SGL} algorithm has a considerable advantage over other competing structured graph learning algorithms in \citet{marlin2009sparse,JMLR:v18:17-019,ambroise2009inferring}.

\begin{thm}\label{thm:conv}
	The sequence $(\w^{t}, \U^{t}, \blambda^{t})$ generated by Algorithm \ref{algo-1} converges to the set of KKT points of \eqref{CGL_L_cost}.
\end{thm}

\begin{proof}
The detailed proof is deferred to the Appendix \ref{apendix-thm}.
\end{proof} 
\begin{rem}
Note that the SGL is not only limited to $k-$component graph learning, but can be easily adapted to learn other graph structures under aforementioned spectral constraints in \eqref{lap-eig-1}, \eqref{reg1}, \eqref{regk}, and \eqref{lap-eig-2}. Furthermore, the SGL can also be utilized to learn popular connected graph structures (e.g., Erdos-Renyi graph, modular graph, grid graph, etc.) even without specific spectral constraints just by choosing the eigenvalue constraints corresponding to one component graph (i.e., $k=1$) and setting $c_1,c_2$ to very small and large values respectively. Detailed experiments with important graph structures are carried out in the simulation section.
\end{rem}

\section{Structured Graph Learning Via Adjacency Spectral Constraints (\textsf{SGA})}\label{sec:adj}
In this section, we develop a BSUM-based algorithm for \textbf{S}tructured \textbf{G}raph learning via \textbf{A}djacencny spectral constraints (\textsf{SGA}). In particular, we consider to solve \eqref{CGL_A} for connected bipartite graph structure by introducing the spectral constraints on the adjacency eigenvalues \eqref{bipartite-eig}. Since $\bTheta$ is a connected graph, the term $\log\text{gdet}(\bTheta)$ can be simplified according to the following lemma.
\begin{Lem1}\label{lem-J}
	If $\bTheta $ is a Laplacian matrix for a connected graph, then
	\begin{align}\label{gdet-det}
	\text{gdet}(\bTheta)=\text{det}(\bTheta+\J),
	\end{align}
	where $\J=\frac{1}{p}\bone\bone^T$. 
\end{Lem1}
\begin{proof}
	It is easy to establish \eqref{gdet-det} by the fact that $\bTheta\bone=\bzero$.
\end{proof}

\subsection{ Graph adjacency operator $\Ao$}\label{A-opeartor}

To guarantee the structure of adjacency matrix, we introduce a linear operator $\Ao$.

\begin{def1}\label{operator-A}
	We define a linear operator $\Ao:\w \in \Rn_{+}^{p(p-1)/2} \rightarrow \Ao \w \in \Rn^{p \times p}$ satisfying 
	\[
	[\Ao\w]_{ij} =
	\begin{cases}
	-w_{i+d_j}\; & \; \;i >j,\\
	[\Ao\w]_{ji} \; & \; \;i<j,\\
	0\; &\;\; i=j,
	\end{cases}
	\]\label{ad:operator}
	where $d_j=-j+\frac{j-1}{2}(2p-j)$.

\end{def1}
An example for $	\Ao \w $ on weight vector of 6 elements $\w=[w_1,w_2,\cdots,w_6]^T$ is given below
{
	\begin{equation}
	\Ao \w= \left[
	\begin{array}{cccc}
	0 & w_1 & w_2 & w_3\\
	w_1 & 0 & w_4 & w_5\\
	w_2 & w_4 & 0& w_6\\
	w_3 & w_5 & w_6 & 0
	
	\end{array}
	\right].
	\end{equation}
}

We derive the adjoint operator $\Ao^*$ of $\Ao$ by making $\Ao^*$ satisfy $\langle \Ao\w,\Y \rangle=\langle \w,\Ao^*\Y \rangle$.
\begin{Lem1}\label{adjoint-A}
	The adjoint operator $\Ao^*: \Y \in \Rn^{p\times p} \mapsto \Ao^* \Y \in \Rn_{+}^{p(p-1)/2}$ is defined by
	\begin{equation}
	[\Ao^* \Y]_k=y_{ij}+y_{ji},
	\end{equation}
	where $i$, $j \in \mathbb{Z}^+$ satisfy $i-j+ \frac{j-1}{2}(2p-j) = k$ and $i>j$.
\end{Lem1}

\begin{Lem1}\label{lem-lap-adj}
	The operator norm $\Vert \Ao\Vert_2 $ is $\sqrt{2}$, $\Vert \Ao \Vert_2 \doteq \sup_{\norm{\x}=1}\norm{\Ao \x}_F$ with $\x \in \Rn ^{\frac{p(p-1)}{2}}$. 
\end{Lem1}
\begin{proof}
	Directly from the definition of operator norm, we have
	\begin{align}
	\norm{\Ao}_2 = \sup_{\norm{\x}=1}\norm{\Ao\x}_F =\sup_{\norm{\x}=1} \sqrt{2}\norm{ \x}=\sqrt{2}, 
	\end{align}
	concluding the proof.
\end{proof}

\subsection{\textsf{SGA} algorithm }
%\colr{ We consider solving \eqref{CGL_A} for learning a bipartite graph structure which has the symmetric eigenvalue constraints its adjacency matrix in \eqref{bipartite-eig}. Note that for some of the adjacency matrix $\Ao \w $ of a bipartite graph and may have $z\geq 0$ number of zero eigenvalues. If there is a need to learn a bipartite graph with $z$ zero eigenvalues. In the formulation, the zero eigenvalues and the corresponding eigenvectors can be dropped: from the symmetry property of eigenvalues, the positioning of the zero eigenvalues and the corresponding eigenvectors are known. After removing $z$ zero eigenvalues and corresponding eigenvectors, we have a diagonal matrix of size $b=n-z$ containing the non-zero eigenvalues ${\bPsi}\in\mathbb{R}^{b\times b}$ at its diagonal in descending order, and matrix of size $b\times b$ containing the eigenvectors corresponding to the non-zero eigenvalues in the same order ${\V }\in \mathbb{R}^{p \times b}$.}

% Without loss of generality, the number of nodes $p$ can be assumed even and then the spectral constraint $\S_\psi$ \eqref{CGL_A_cost} for connected bipartite graph learning becomes
%\begin{align}\label{Eig_bipartite-1}
%\S_{ \psi}:=\big\{ &\psi_i=-\psi_{p+1-i}\; \; \text{for}\;\; i=1, \cdots, p \\
%& c_1\geq\psi_1 \geq \psi_2\geq\cdots\geq\psi_{p}\geq c_2,\nonumber
%\big\}\;.
%%&\; \psi_i=-\psi_{q+1-i}, \; \\
%\end{align}
%where $c_1,c_2>0$ are some constants. 

By introducing the operators $\L$ and $\Ao$ and relaxing the constraint $\Ao\w=\V \Di(\bpsi)\V^T$ in \eqref{CGL_A} to the objective function, we obtain the following formulation: 
\begin{align}\label{CGL_A_cost}
\begin{array}{ll}
\underset{{\w},{\bpsi},{ \V}}{\text{minimize}} &
\begin{array}{c}
- \log \det (\L \w+\J)+\tr{\K\L \w}+\frac{\gamma}{2}\Vert \mathcal{A} \w-\V \Di(\bpsi) \V^T \Vert_F^2,
\end{array}\\
\text{subject to} & \begin{array}[t]{l}
\w \geq 0, \ \bpsi \in \S_{\psi}, \ \V^T\V=\I, 
\end{array}
\end{array}
\end{align}
where $\gamma>0$ is the penalty parameter. Suppose there are $ z$ zero eigenvalues in the set $\S_{\psi}$ with $ z\geq 0$. From the symmetry property of the eigenvalues, the zero eigenvalues are positioned in the middle, i.e., in \eqref{bipartite-eig} the eigenvalues $\psi_{\frac{p-z}{2}+1}$ to $\psi_{\frac{p+z}{2}}$ will be zero. Both $(p+z)$ and $(p-z)$ must be even by the symmetry property. As a consequence, the zero eigenvalues and the corresponding eigenvectors can be dropped from the formulation. Now $\bpsi \in \Rn^{b}$ contains $b$ number of the non-zero eigenvalue and $V \in \Rn^{p\times b}$ contains the corresponding eigenvectors, and satisfy $V^TV=I_b$. The non-zero eigenvalues are required to lie in the following set:
\begin{align}\label{Eig_bipartite}
 \S_{ \psi}:=\big\{ &\psi_i=-\psi_{b+1-i}\; \; \text{for}\;\; i=1, \cdots, b/2 \\
 & c_1\geq\psi_1 \geq \psi_2\geq\cdots\geq\psi_{b/2}\geq c_2,\nonumber
 \big\}\;.
 %&\; \psi_i=-\psi_{q+1-i}, \; \\
\end{align}
where $c_1,c_2>0$ are some constants which depend on the graph properties. Collecting the variables in three block as $\x=\left(\w\in \Rn^{p(p-1)/2}, \bpsi\in \Rn^{b}, \V\in \Rn^{p \times b}\right)$, we develop a BSUM-based method which updates one variable each time with the other blocks fixed.

\subsubsection{Update of $\w$}

Optimization of \eqref{CGL_A_cost} with respect to $\w$: ignoring the terms independent of $\w$, we have the following sub-problem:
\begin{align}
\underset{{\w \geq 0}}{\text{minimize}} \quad &
- \log \det (\L \w + \J)+\tr{\K\L \w}+\frac{\gamma}{2}\Vert \mathcal{A} \w - \V \Di(\bpsi) \V^T \Vert_F^2. \label{w1}
\end{align}
The problem \eqref{w1} can be equivalent written as 
\begin{align}
\underset{\w \geq 0}{\text{minimize}}\qquad - \frac{1}{\gamma}\log \det (\L \w + \J)+ \frac{1}{2}\norm{\Ao \w}^2_F - {\tilde{\cb}}^T\w, \label{w2}
\end{align}
where $\tilde{\cb}=\mathcal{A}^*(\V \Di(\bpsi) \V^T)- \gamma^{-1}\L^*\K$.

\vspace*{2 ex}
\begin{Lem1}\label{lem3}
	The sub-problem \eqref{w2} is strictly convex.
\end{Lem1}
\begin{proof}
	First, we can see $- \log \det (\L \w + \J)$ is a convex function. From the definition of the operator $\Ao$ and the property of its adjoint $\Ao^*$, we have 
	\begin{align}\label{Porf2}
	\| \Ao \w\|_F^2=\langle \Ao \w, \Ao \w \rangle=\langle \w, \Ao^*\Ao \w \rangle=\w^T \Ao^*\Ao \w >0,\; \text{for any}\; \w\neq\bzero,
	\end{align}
	implying the cost function in \eqref{w2} is a convex function. Furthermore, the non-negativity set is convex. Therefore, the sub-problem \eqref{w2} is strictly convex. However, due to the non-negativity constraint ($\w\geq0$), there is no closed-form solution to \eqref{w2} and thus we derive a majorized function.
\end{proof}

\begin{Lem1}\label{lem4} 
	The function $h(\w)= - \frac{1}{\gamma}\log \det (\L \w + \J)+\frac{1}{2}\norm{\Ao \w}^2_F - {\tilde{\cb}}^T\w$ in \eqref{w2} is majorized at $\w^{t}$ by the function
	\begin{align}\label{sur1-adj}
	&q( \w|\w^t )= h( \w^t) +(\w - \w^t )^T \left(\nabla h ( \w^t)\right) + \frac{L}{2}(\w - \w^t )^T (\w - \w^t)
	\end{align}
	where {$L=(\norm{\Ao}^2_2+L_2/\gamma)$}, in which $\norm{\Ao}^2_2=2$, and $-\log \det (\L \w+\J)$ is assumed to be $L_2$-Lipschitz continuous gradient.
\end{Lem1}

\begin{proof}
Easy to see, when $h_1(\w) = -\log \det (\L \w+\J)$ is $L_2$-Lipschitz continuous gradient, we have 
\begin{align} \label{L11p1}
h_1(\w) & \leq h_1(\w^t) + (\w - \w^t )^T \nabla h_1 ( \w^t) + \frac{L_2}{2}\norm{\w - \w^t}^2.
\end{align}
By Taylor expansion for $h_2(\w) = \frac{1}{2}\norm{\Ao \w}^2_F - {\tilde{\cb}}^T\w $, we can get
\begin{align}\label{L11p2}
h_2(\w) &= h_2(\w^t) + (\w - \w^t )^T \nabla h ( \w^t) + \frac{1}{2} \inPro{\w - \w^t}{\Ao^*\Ao (\w - \w^t)} \nonumber \\
& = h_2(\w^t) + (\w - \w^t )^T \nabla h ( \w^t) + \frac{1}{2}\norm{\Ao(\w - \w^t) }_F^2 \\
& \leq h_2(\w^t) + (\w - \w^t )^T \nabla h ( \w^t) + \frac{1}{2}\norm{\Ao}^2_2 \norm{\w - \w^t}^2, \nonumber
\end{align}
where the inequality is established due to Lemma \ref{lem-lap-adj}. We can finish the proof by combining \eqref{L11p1} and \eqref{L11p2}.
\end{proof}

\vspace*{2 ex}

After ignoring the constant terms in \eqref{sur1-adj}, the majorized problem of \eqref{w2} at $\w^t$ is given by
\begin{align}
\underset{\w \geq 0}{\text{minimize}}\qquad \frac{1}{2}{\w}^T\w - {\bb}^T\w, \label{majr:adj}
\end{align}
where $\bb= \w^t-\frac{1}{L}\left(\frac{1}{\gamma}\L^*(\L \w^t+\J)^{-1}+\Ao^*\Ao(\w^t)- \tilde{\cb}\right)$.
\begin{Lem1}
	From the KKT optimality conditions we can
	obtain the optimal solution as
	\begin{align}\label{w_up:algo-2}
	\w^{t+1}= \left(\w^t-\frac{1}{L}\left(\frac{1}{\gamma}\L^*(\L \w^t+\J)^{-1}+\Ao^*\Ao(\w^t)- \tilde{\cb}\right)\right)^+,
	\end{align}
	where $(a)^+:=\max(a,0)$.
\end{Lem1}
\begin{rem}
The Lipschtiz constant $L_2$ of the function $-\log \det (\L \w + \J)$ is related to the smallest non-zero eigenvalue of $\L\w$. The smallest non-zero eigenvalue $\lambda_{\min}(\L\w)$ is also known as the algebraic connectivity of the graph, which is bounded as follows $\lambda_{\min}(\L\w) \geq \frac{\epsilon_w}{(p-1)^2}$ \citep[see Lemma 1,][]{rajawat2017stochastic}, where $\epsilon_w>0$ is the minimum non-zero graph weight.
Note that, the graph weight in our formulation is not imposed to be lower bounded away from zero and thus the Lipschitz constant $L_2$ may be unbounded. However, for practical purposes the edges with very small weights can be ignored and set to be zero, and we can assume that the non-zero weights are bounded by some constant $\epsilon_w>0$. To strictly force the minimum weight property, one can modify the non-negativity constraint $\w\geq 0$ in problem \eqref{CGL_A_cost} as $\w\geq \epsilon_w$. On the other hand, we do not need a tight Lipschtiz constant $L_2$. Actually, any $L'_2 \geq L_2$ can still make the function $q( \w|\w^t )$ satisfy the majorization function conditions \eqref{bsum-maj-0}. 
\end{rem}

\subsubsection{Update of $V$}
To update $\V$, we get the following sub-problem:
\begin{align}\label{V_A:1}
\begin{array}{ll}
\underset{{ \V}}{\text{maximize}} &
\begin{array}{c}
{\rm \tr{\V^T \mathcal{A} \w \V \Di(\bpsi)}},
\end{array}\\
\text{subject to} & \begin{array}[t]{l}
\V^T\V=\I_b.
\end{array}
\end{array}
\end{align}
The problem \eqref{V_A:1} is an optimization on the orthogonal Stiefel manifold $\text{St}(p,b)=\{\V\in \mathbb{R}^{p \times b}: \V^T\V=\I_b\} $.
\begin{Lem1}
	From the KKT optimality conditions the solution to \eqref{V_A:1} is given by
	\begin{align}\label{u_up:algo-2}
	\V=\textsf{eigenvectors}(\Ao\w)\left[1:\frac{(p-z)}{2}, \ \frac{(p+z)}{2}+1:p\right]
	\end{align}
	that is, the eigenvectors of the matrix $\Ao \w$ in the same order of the eigenvalues, where $z$ is the number of zero eigenvalues. 
\end{Lem1}
The solution \eqref{u_up:algo-2} satisfies the optimality condition of \eqref{V_A:1} on the orthogonal Stiefel manifold and more details about the derivation can be found in \citet{absil2009optimization}.

\subsubsection{Update for $\mathbf{\psi}$}\label{psi-adj-bip}
Solving for \eqref{CGL_A_cost} with respect to $\bpsi$, ignoring the terms independent of $\bpsi$, we have the following sub-problem:
\begin{align}\label{Sigma_A}
\begin{array}{ll}
\underset{{\bpsi}}{\text{minimize}} &
\begin{array}{c}
\Vert \mathcal{A} \w-\V \Di(\bpsi) \V^T \Vert_F^2.
\end{array}\\
\text{subject to} & \begin{array}[t]{l}
\ \bpsi \in \S_{\psi}, 
\end{array}
\end{array}
\end{align}
The optimization \eqref{Sigma_A} can also be written as
\begin{align}\label{sub_psi}
\underset{
	\boldsymbol{\psi}	\in \S_{ \psi}
}{\text{minimize}} \quad & \| \boldsymbol{\psi}-{\b e} \|_2^2,
\end{align}
where $\boldsymbol{\psi}=[\psi_1,\psi_2,\cdots,\psi_b]^T$ and ${\b e}=[e_1, e_2,\cdots,e_b]^T$, in which $\psi_i$ and $e_i$ correspond to the $i$-th diagonal element of $\Di(\bpsi)$ and $\text{Diag}(\V^T\Ao\w\V)$, respectively. The problem in \eqref{sub_psi} is a convex optimization problem with simple linear constraints. 
Furthermore, upon utilizing the symmetry property of the constraint set $ \S_{ \psi}$ (i.e., $\psi_i=-\psi_{b+1-i}$, $i=1, \cdots, b/2$), we can reformulate an equivalent problem which only requires to solve a problem for the first half of the variables $[\psi_1,\psi_2,\cdots, \psi_{b/2}]^T$. 

\begin{Lem1}\label{iso-adj}
	Consider the following isotonic regression problem for ${\tilde{\boldsymbol{\psi}}=[\psi_1,\psi_2,\cdots, \psi_{b/2}]^T}$
	\begin{align}\label{sub_psi_iso}
	\begin{array}{ll}
	\underset{{\tilde{\boldsymbol{\psi}}}}{\text{minimize}} &
	\begin{array}{c}
	\| \tilde{\boldsymbol{\psi}}-\tilde{\b e} \|_2^2,
	\end{array}\\
	\text{subject to} & \begin{array}[t]{l}
	c_1 \geq\psi_1\geq \psi_2 \geq \cdots \geq \psi_{b/2}\geq c_2, 
	\end{array}
	\end{array}
	\end{align}
	where $c_1,c_2>0$ and $\tilde{\b e}=[\tilde{e}_1,\tilde{e}_2,\cdots, \tilde{e}_{b/2}]^T$, $\tilde{e}_i=\frac{e_i+e_{b+1-i}}{2}, \; i=1,2, \cdots, b/2 $. The first half of the solution to the problem \eqref{sub_psi} is same as the solution to \eqref{sub_psi_iso}.
\end{Lem1}
\begin{proof}This holds from the symmetry property of the eigenvalue constraints, $\psi_i=-\psi_{b+1-i}$, $i=1, \cdots, b/2$. See the Appendix \ref{appendix-isotonic} for the derivation. The formulation in \eqref{sub_psi_iso} is also similar to the problem in \eqref{sub-lambda} without the log determinant term. The solution to the problem \eqref{sub_psi_iso} can be obtained by following the iterative procedure discussed in Algorithm \ref{algo-lambda}, by setting $ {\psi}_i$=$\tilde{e}_i$ for $i=1, 2, \cdots, b/2$, and iteratively updating and checking all the situations until all the $\psi_i$'s satisfy $c_1 \geq \psi_1\geq \psi_2,\cdots, \psi_{b/2}\geq c_2$. Next, the solution for the other half of the variables $\{\psi_i\}_{i=b/2+1}^b$ in \eqref{sub_psi} are obtained by setting $\psi_{b/2+1+i}=-\psi_{i}$, for $i=1,\cdots,b/2. $
\end{proof}

%Furthermore, the problem of solving a Laplacian system has been extensively studied. There exist a number of practical, out-of-the-box implementations that incurs a complexity of $O(|\E|\log n)$, where $|\E|$ is the nonzero number of elements in $ \w$; see e.g., \citep{livne2012lean,koutis2011combinatorial}.
%\begin{algorithm}

\subsubsection{\textsf{SGA} algorithm summary}

Algorithm \ref{algo-2} summarizes the implementation of structured graph learning (\textsf{SGA}) via adjacency spectral constraints. The computationally most demanding steps of \textsf{SGA} are the eigenvalue decomposition and matrix inversion of $p \times p$ matrices, implying $O(p^3)$ as the worst-case computational complexity of the algorithm. But it can be improved by utilizing the sparse structure and the properties of the symmetric adjacency matrix and Laplacian matrix for the eigenvalue decomposition and matrix inversion \citep{livne2012lean,koutis2011combinatorial}, respectively.
\begin{algorithm} 
	\caption{\textsf{SGA} }
	\label{algo-2}
	\begin{algorithmic}[1]
		\STATE	\textbf{Input:} SCM $S$, $ c_1, c_2,\gamma$
		\STATE	\textbf{Output:} $\hat{\bTheta}$
		\STATE $t \leftarrow 0$
		\WHILE {Stopping criteria not met}
		\STATE Update $\w^{t}$ as in \eqref{w_up:algo-2};
		\STATE Update $ \V^{t}$ as in \eqref{u_up:algo-2};
		\STATE Update $ \bpsi^{t}$ by solving \eqref{sub_psi_iso};
		\STATE $t\leftarrow t+1$
		\ENDWHILE
		%		\UNTIL Inner loop convergence
		%		\STATE $\gamma \gets \rho \gamma$ with $\gamma >1$;
		%		\UNTIL Outer loop convergence
		\STATE RETURN $\hat{\bTheta}^{t+1}=\L\w^{t+1}$\\
	\end{algorithmic}
\end{algorithm}

The subsequence convergence result for \textsf{SGA} algorithm is established.
\begin{thm}\label{thm:conv:adj}
	The sequence $(\w^{t}, V^{t}, \bpsi^{t})$ generated by Algorithm \ref{algo-2} converges to the set of KKT points of Problem \eqref{CGL_A_cost}.
\end{thm}
\begin{proof}
	The proof of Theorem \ref{thm:conv:adj}is similar to that of Theorem \ref{thm:conv} and thus is omitted.
\end{proof}

\begin{rem}
	Combinatorially, finding a bipartite structure is equivalent to a max-cut problem between $\Vc_1$ and $\Vc_2$, which is an NP-hard problem. A recent work in \citet{pavez2018learning} considered an approximate bipartite graph estimation under structured GGM setting. The algorithm consists of a two-stage procedure: first, they learn a bipartite structure by Goemans-Williamson (GM) algorithm \citep{goemans1995improved}; then, they learn the Laplacian weights by the generalized graph Laplacian (GGL) learning method \citep{egilmez2017graph}. The GM algorithm, dominated by a semi-definite programming and Cholesky decomposition of $p \times p $ matrix, has a worst-case complexity of order $O(p^3)$ and $O(p^3)$, respectively, while the GGL method has the computational complexity of order $O(p^3)$. Therefore, \textsf{SGA} algorithm enjoys a smaller worst-case computational complexity than that in \citet{pavez2018learning}. In addition, to the best of our knowledge, SGA is the first single stage algorithm for learning bipartite graph structure directly from the data sample covariance matrix. 
\end{rem}

\section{Structured Graph Learning Via Joint Laplacian and Adjacency Spectral Constraints (\textsf{SGLA})}\label{sec:lap-adj}

Now we consider the problem \eqref{CGL_L_A} for \textbf{S}tructured \textbf{G}raph learning via joint \textbf{L}aplacian and \textbf{A}djacency spectral constraints (\textsf{SGLA}). Following from the discussions in previous sections: upon utilizing the Laplacian operator $\L$, the adjacency operator $\Ao$ and moving the constraints $\Ao\w=\V \Di(\bpsi)\V^T$ and $ \L\w=\U\Di(\blambda)\U^T$ into the objective function, a tractable approximation of \eqref{CGL_L_A} can be written as 
\begin{align}\label{CGL_A-Rlx}
\underset{{\w},{\blambda},{\U}, \bpsi, \V}{\text{minimize}} &\;\;\; - \log \det \Di(\blambda)+\tr{\K \L \w}+\frac{\beta}{2}\| \L \w -\U \Di(\blambda)\U^T\|_F^2 \\ \nonumber 
&\hspace{6.5cm}+\frac{\gamma}{2}\| \Ao \w -\V \Di(\bpsi) \V^T\|_F^2,\\
\text{subject to} & \;\;\;\w \geq 0,\ \blambda \in \S_{\lambda}, \ \U^T\U=\I,\ \bpsi \in \S_{\psi}, \ \V^T\V=\I,
\end{align}

where $\beta, \gamma>0$ are the tradeoff between each term in \eqref{CGL_A-Rlx}. Following the discussion from the derivation of Algorithms in \ref{algo-1} and \ref{algo-2} and collecting the variables $\x=\big(\w,\blambda, \U, \bpsi, \V \big)$ : a BSUM-type method for solving the problem \eqref{CGL_A-Rlx} is summarized below.

Ignoring the terms independent of $\w$, we have the following sub-problem:
\begin{align}\label{sub-x-lap-adj}
\underset{\w \geq 0}{\text{minimize}}\qquad \tr{\K \L \w} +\frac{\beta}{2}\| \L \w -\U \Di(\blambda) \U^T\|_F^2+\frac{\gamma}{2}\| \Ao \w -\V \Di(\bpsi) \V^T\|_F^2.
\end{align}

Following some simple manipulation, an equivalent problem is 
\begin{align}\label{quad-x-lap-adj}
\underset{\w \geq 0}{\text{minimize}}\qquad f(\w):=f_1(\w)+f_2(\w)
\end{align}
where $f_1(\w)=\frac{\beta}{2}\norm{\L\w}^2_F-\cb_1^T\w$, $f_2(\w)=\frac{\gamma}{2}\norm{\Ao\w}^2_F -\cb_2^T\w$, and $\cb_1=\beta\L^*(\U \Di(\blambda) \U^T- \beta^{-1}\K)$, $\cb_2=\gamma\Ao^*(\V \Di(\bpsi) \V^T)$. A closed form update for problem can be obtained by utilizing the MM framework.

\begin{Lem1}\label{lemadj-lap} 
	The function $f(\w)$ in \eqref{quad-x-lap-adj} is majorized at $\w^{t}$ by the function
	\begin{align}
	g( \w|\w^{t} )&=f_1( \w^t) + (\w - \w^{t} )^T \nabla f_1 ( \w^{t}) +\frac{L_1}{2}\norm{\w - \w^{t}}^2, \label{sur} \\
	&+f_2( \w^t) + (\w - \w^{t} )^T \nabla f_2 ( \w^{t}) +\frac{L_2}{2}\norm{\w - \w^{t}}^2
	\end{align}
	where $\w^t$ is the update from previous iteration, $L_1=\beta\norm{\L}^2_2=2p\beta$ and $L_2=\gamma\norm{\Ao}^2_2=2\gamma$. The condition for the majorization function can be easily checked \citep{SunBabPal2017-MM,song2015sparse}.
\end{Lem1}

\begin{Lem1}
	From the KKT optimality conditions the optimal solution for $\w^{t+1}$ takes the following form
	\begin{align}\label{w-up-lap-adj_1}
	\w^{t+1}=\left[\w^t-\frac{1}{L_1+L_2}\big(\nabla f_1(\w^t)+\nabla f_2(\w^t)\big)\right]^+,
	\end{align}
	where $(a)^+:=\max(a,0)$, and $\nabla f_1(\w^t)=\beta \L^*(\L\w^t)-\cb_1 $, $\nabla f_2(\w^t)=\gamma\Ao^*(\Ao\w^t)-\cb_2 $, $L_1+L_2=2(p\beta+\gamma)$.
\end{Lem1}

The update for $\U$ and $\V$ take the same forms as discussed in \eqref{algo1:U-update} and \eqref{u_up:algo-2}. The update for $\blambda$ can be solved by Algorithm \ref{algo-lambda}, since the Laplacian spectral set $\S_{\lambda}$ for any graph structure can always be presented in the form of \eqref{Eig_set_K-connected}. Next, the sub-problem for $\bpsi$ takes a similar form as presented in \eqref{sub_psi} with a specific $\S_{ \psi}$ depending on the graph structure. For the structures belonging to bipartite graph families (e.g., connected bipartite, multi-component bipartite and regular bipartite graph), the sub-problem for $\bpsi$ can be solved efficiently by the algorithm in \ref{psi-adj-bip}. When we consider a more general convex set $\S_{\psi}$, the problem \eqref{sub_psi} is still a convex optimization problem and thus can be solved by using some standard solvers like \textsf{CVX}. Note that the sub-problems involving $(U, \blambda)$ and $(V, \bpsi)$ are decoupled. As a consequence of it, the update steps for $U,\blambda$ and $V,\bpsi$ can be done in parallel.

\subsection{\textsf{SGLA} algorithm summary}

Algorithm \ref{algo-3} summarizes the implementation of the structured graph learning (\textsf{SGLA}) via joint Laplacian and adjacency spectral constraints. In Algorithm \ref{algo-3}, the computationally most demanding step is the eigenvalue decomposition, implying $O(p^3)$ as the worst-case computational complexity of the algorithm. Note that we need to conduct eigenvalue decomposition twice in each iteration and how to cut down the computational complexity will be considered in future work.

\begin{algorithm}
	\caption{\textsf{SGLA}}
	\label{algo-3}
	\begin{algorithmic}[1]
		\STATE	\textbf{Input:} SCM $S$, $ \S_\lambda, \S_{\psi}, \beta, \gamma$.
		\STATE	\textbf{Output:} $\hat{\bTheta}$
		\STATE $t \leftarrow 0$
		\WHILE {Stopping criteria not met}
		\STATE Update $\w^{t}$ as in \eqref{w-up-lap-adj_1};
		\STATE Update $ \U^{t}$ using \eqref{algo1:U-update};
		\STATE Update $ \V^{t}$ using \eqref{u_up:algo-2};
		\STATE Update $ \blambda^{t}$ using Algorithm \ref{algo-lambda} under $\S_\lambda$;
		\STATE Update $ \bpsi^{t}$ by solving \eqref{sub_psi} under $\S_\psi;$
		\STATE $t\leftarrow t+1$
		\ENDWHILE	
		%		\UNTIL Inner loop convergence
		%		\STATE $\beta \gets \rho \beta$ with $\beta >1$;
		%		\UNTIL Outer loop convergence
		\STATE RETURN $\hat{\bTheta}^{t+1}=\L\w^{t+1}$
	\end{algorithmic}
\end{algorithm}

In the \textsf{SGLA} algorithm, consideration of the spectral constraints of the two graph matrices jointly is the enabling factor for enforcing some of the intricate structure. Consider case of learning a $k$-component bipartite graph structure popularly also known as bipartite graph clustering. This structure seeks partition of the nodes into $k$ disjoint partitions along with $k$ bi-partitions, which seems to be an incompatible goal. For the $k-$disjoint components, it requires that there are edges only connecting the nodes within the same partition (i.e., no intra-partition edges), while for the $k$ bi-partition we require the edges connecting two nodes not belonging to the same partition (i.e., no inter-partition edges). The intricate structure requirement makes such a problem extremely challenging and until, there is no one-stage method capable of learning this structure directly from the data. Whereas within the \textsf{SGLA} algorithm we can easily satisfy the structural requirements by plugin the Laplacian spectral properties for the $k$-component structure along with the adjacency spectral constraints for the bipartite structure (i.e., $\S_\lambda$ and $\S_{ \psi}$ as in \eqref{k-comp-bipartite}) in Algorithm \ref{algo-3} [cf. step 8, 9].

The subsequence convergence result for \textsf{SGLA} algorithm is established.

\begin{thm}\label{thm:conv:lap-adj}
	The sequence $(\w^{t}, U^{t}, \blambda^t, V^{t}, \bpsi^{t})$ generated by Algorithm \ref{algo-3} converges to the set of KKT points of Problem \eqref{CGL_A-Rlx}.
\end{thm}
\begin{proof}
The detailed proof is deferred to the Appendix \ref{appendix-conv:lap-adj}.
\end{proof}

\section{Experiments}\label{simulations}
In this section, we present experimental results for comprehensive evaluation of the proposed algorithms SGL, SGA, and SGLA\footnote{An \textsf{R} package for \textsf{SGL, SGA,} and \textsf{SGLA} is available at https://github.com/dppalomar/spectralGraphTopology. }. The advantage of incorporating spectral information in the proposed framework is clearly illustrated. First, we introduce the experimental settings in Subsection \ref{exp-set} and the benchmarks for comparison in \ref{benchmarks}. Then the experiments are organized in the following three parts: Subsection \ref{simulation-algo1} evaluates \textsf{SGL} for learning the following graph structures: Grid, Modular, and multi-component graphs; Subsection \ref{simulation-algo2} shows bipartite graph learning by \textsf{SGA}, and Subsection \ref{simulation-algo2} shows multi-component bipartite graph learning via \textsf{SGLA}.

\subsection{Experimental settings}\label{exp-set}
For synthetic experiments we create several synthetic data sets based on different graph structures $\G$. First, we generate an improper GMRF model parameterized by the true precision matrix $\bTheta_\G$, which follows the Laplacian constraints in \eqref{Lap-set} as well as the specific graph structure. Then, a total of $n$ samples $\{\x_i\in \Rn^p\}_{i=1}^n$ are drawn from the IGMRF model with $\bTheta_\G$: $\x_i \sim \mathcal{N}(\bzero, \bTheta^\dagger_\G)$, $\forall i$. The sample covariance matrix $S$ is computed as,
\begin{align}\label{scm}
S=\frac{1}{n}\sum_{i=1}^n(\x_i-\bar{\x}_i)(\x_i-\bar{\x}_i)^T,\; \; \text{with}\;\; \bar{\x}_i=\frac{1}{n}\sum_{i=1}^n{\x_i}.
\end{align}
% $S=\frac{1}{n}\sum_{i=1}^n(\x_i-\bar{\x}_i)(\x_i-\bar{\x}_i)^T$ with $\bar{\x}_i=\frac{1}{n}\sum_{i=1}^n{\x_i} $.
 Algorithms use the SCM $S$ and prior information regarding target graph families, if available (e.g., number of connected components $k$, bipartite, and, etc.). We set $c_1$ and $c_2 $ to very small and large value, respectively, and the choice of $\beta$, $\gamma$ and $\alpha$ are discussed for each case separately. For each scenario, 20 Monte Carlo simulations are performed. For performance evaluation, we use following metrics, namely, relative error (\textsf{RE}) and F-score (\textsf{FS}):
\begin{align}
& \textsf{Relative Error}=\frac{\norm{\hat{\bTheta}-\bTheta_{\text{true}}}_F}{\norm{\bTheta_{\text{true}}}_F},\;\; \textsf{F-Score}=\frac{2\textsf{tp}}{2\textsf{tp}+\textsf{fp}+\textsf{fn}},
\end{align}
where $\hat{\bTheta}=\L\hat{\w}$ is the final estimation result of the algorithm and $\bTheta_{\text{true}} $ is the true reference graph Laplacian matrix, true positive (\textsf{tp}) stands for the case when there is an actual edge and the algorithm detects it; false positive (\textsf{fp}) stands for the case when algorithm detects an edge but no actual edge present; and false negative (\textsf{fn}) stands for the case when algorithm misses an actual edge present. Further, we disregard an edge if its weight value is less than $0.1$. The F-score metric takes values in $[0, 1]$ where 1 indicates perfect structure recovery \citep{egilmez2017graph}. To check the performance evolution for each iteration $t$ we evaluate the \textsf{RE} and \textsf{ FS} with $\hat{\bTheta}^t$. Algorithms are terminated when the relative change in ${\w}^t$ is relatively small.

\subsection{Benchmarks}\label{benchmarks}
The \textsf{CGL} algorithm proposed in \citet{egilmez2017graph} is the state-of-the-art method for estimating a connected combinatorial graph Laplacian matrix from the sample covariance matrix. For synthetic data experiments with connected graph structure (e.g., modular, grid, and connected bipartite), we compare the performance of the \textsf{SGL} algorithm against \textsf{CGL}. Additionally, for more insight, we also compare against some heuristic based approaches. These are i) the pseudo-inverse of the sample covariance matrix $S^\dagger$, denoted as \textsf{Naive} and {ii}) the solution of following quadratic program $\w_{\text{qp}}=\argmin_{\w \geq 0} \norm{S^\dagger-\L\w}^2_{F},$ denoted as $\textsf{QP}$. 

For the comparison on multi-component graph learning, as per our knowledge, there is no existing method to learn graph Laplacian matrix with multiple components (e.g., $k-$component and $k-$component bipartite). Thereby, for the sake of completeness, we compare against \textsf{Naive} and \textsf{QP}, which are expected to give meaningful comparisons for high sample scenarios.

For experiments with real data, we compare the algorithm performance with \textsf{GLasso} \citep{friedman2008sparse}, \textsf{GGL}\footnote{Code for the methods \textsf{CGL}, \textsf{GGL} is available at https://github.com/STAC-USC/Graph\_Learning}, constrained Laplacian rank algorithm \textsf{CLR} \citep{nie2016constrained}, \textsf{Spectral clustering} \citep{ng2002spectral}, and $k-$\textsf{means clustering}. Unlike \textsf{CGL}, the \textsf{GGL} algorithm aims to estimate a generalized graph Laplacian matrix. As observed in \citet{egilmez2017graph}, \textsf{GGL} performance is always superior than \textsf{CGL}, therefore, for real data we omit the comparison with \textsf{CGL}. Note that the \textsf{GGL} and \textsf{GLasso} cannot estimate the standard Laplacian matrix in \eqref{Lap-set}, thereby, we cannot compare against those for the synthetic experiments. For \textsf {CGL, GGL, and GLasso} the sparsity parameter $\alpha$ is chosen according to the suggested procedures \citep{egilmez2017graph,Zhao:2012:HPH:2503308.2343681}. 

\subsection{Performance evaluation for SGL Algorithm}\label{simulation-algo1}
In this Subsection, we evaluate the performance of the \textsf{SGL} algorithm, i.e., Algorithm \ref{algo-1} on grid graph, modular graph, multi-component graph, popular synthetic structures for clustering, and real data (\textsf{animal:cancer}).

\subsubsection{Grid graph}
\begin{figure}[!htb]
	\centering
	\begin{subfigure}[b]{0.3\textwidth}
		\includegraphics[width=\textwidth]{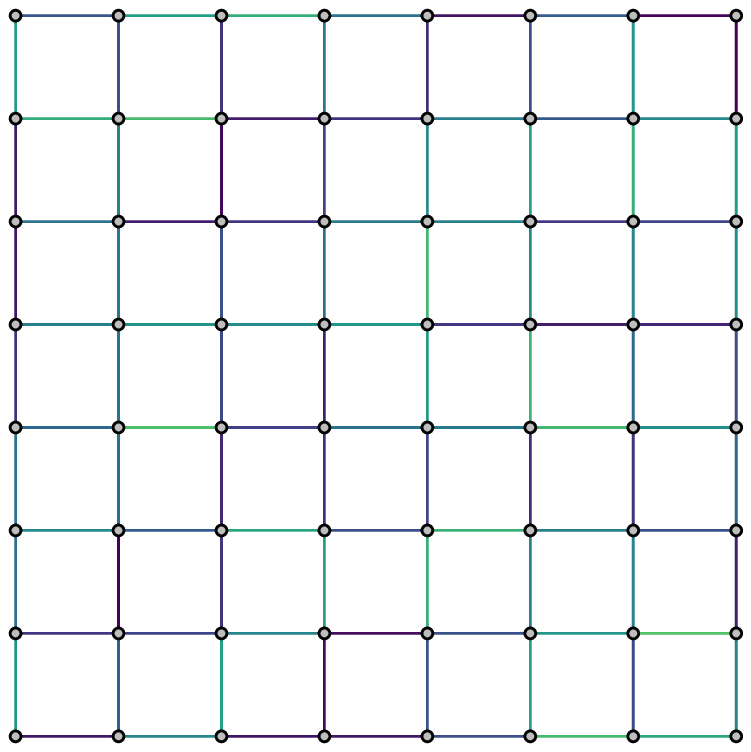}
		\caption{True grid graph}
	\end{subfigure}
	~ %add desired spacing between images, e. g. ~, \quad, \qquad, \hfill etc.
	%(or a blank line to force the subfigure onto a new line)
	\begin{subfigure}[b]{0.3\textwidth}
			\includegraphics[width=\textwidth]{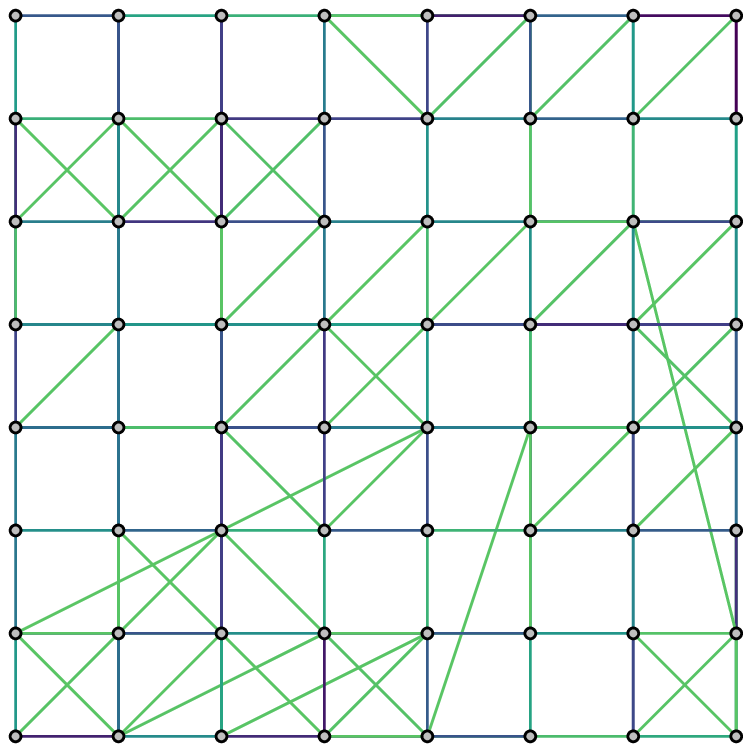}
		\caption{\textsf{CGL}}
	\end{subfigure}
	~
	\begin{subfigure}[b]{0.3\textwidth}
		\includegraphics[width=\textwidth]{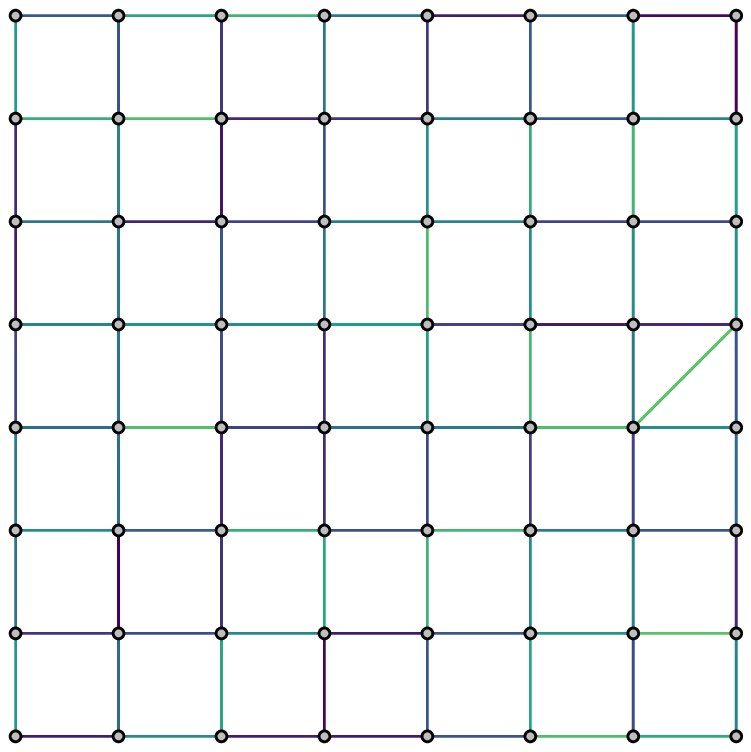}
		\caption{\textsf{SGL}(Proposed)}
	\end{subfigure}
	\caption{Sample results of learning $\mathcal{G}_{\mathsf{grid}}{(64)}$ (a) True grid graph, (b) \textsf{CGL} $(\mathsf{RE}=0.09163, \mathsf{FS}=0.8057) $, and (c) \textsf{SGL} $(\mathsf{RE}= 0.0490, \mathsf{FS}= 0.9955). $}
	\label{fig:sample-grid}
\end{figure}

We consider a grid graph structure denoted as $\G_{\mathsf{grid}}(p)$, where $p=64$ are the number of nodes, each node is attached to their $4$ nearest neighbors (except the vertices at the boundaries), edge weights are selected randomly uniformly from $[0.1,3]$.
Figure~\ref{fig:sample-grid} depicts the graph structures learned by \textsf{SGL} and \textsf{CGL} for $n/p = 100$, edges smaller than $0.05$ were discarded. For \textsf{CGL} we use $\alpha = 0.005$ whereas, for \textsf{SGL}, we fix $\beta = 20$ and $\alpha=0.005$. 

\begin{figure}[!htb]
	\centering
\hspace{-1.2cm}	\begin{subfigure}[b]{0.45\textwidth}
		\includegraphics[width=1\textwidth]{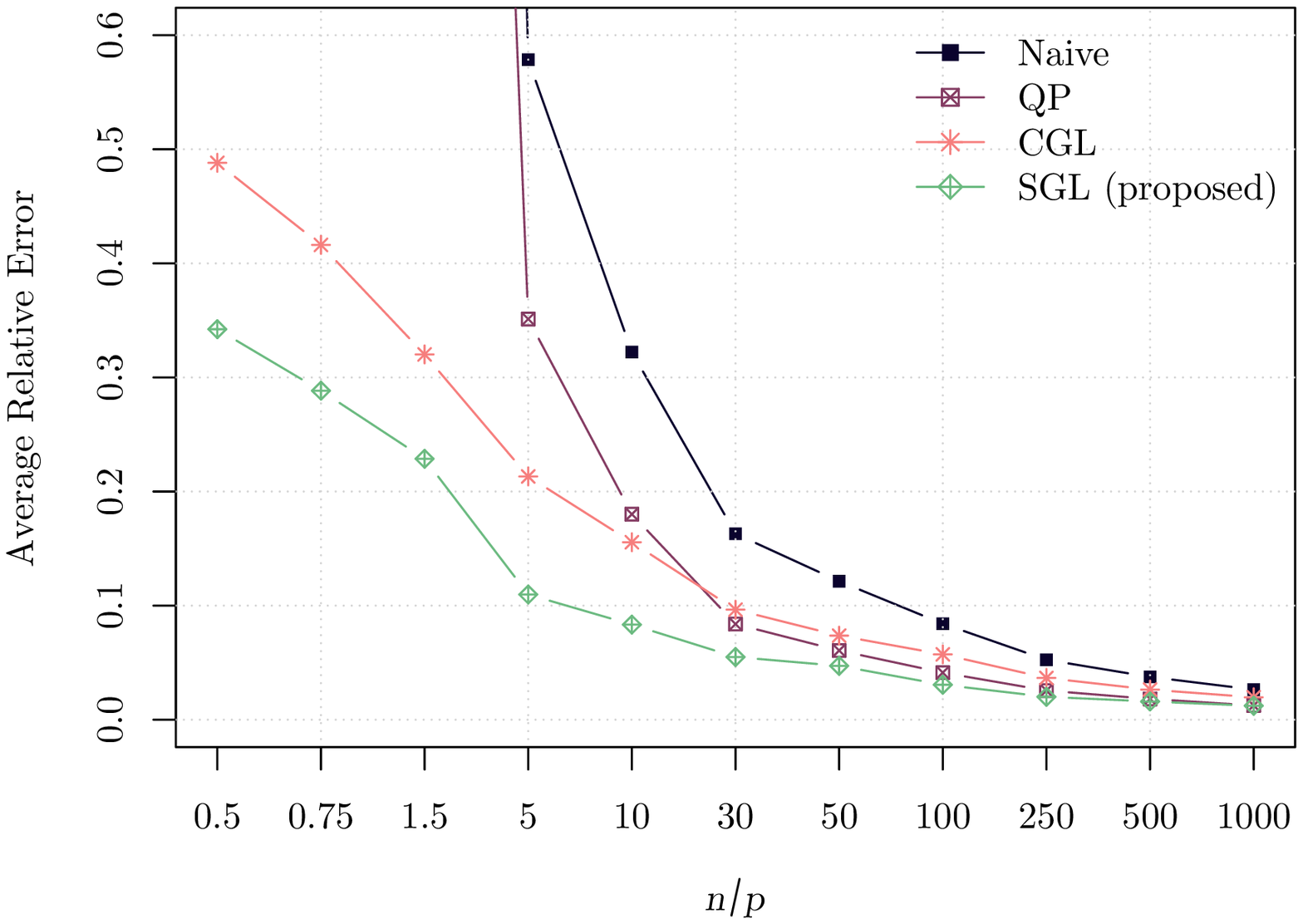}
	\end{subfigure}\qquad 
	~ 
	\begin{subfigure}[b]{0.45\textwidth}
		\includegraphics[width=1\textwidth]{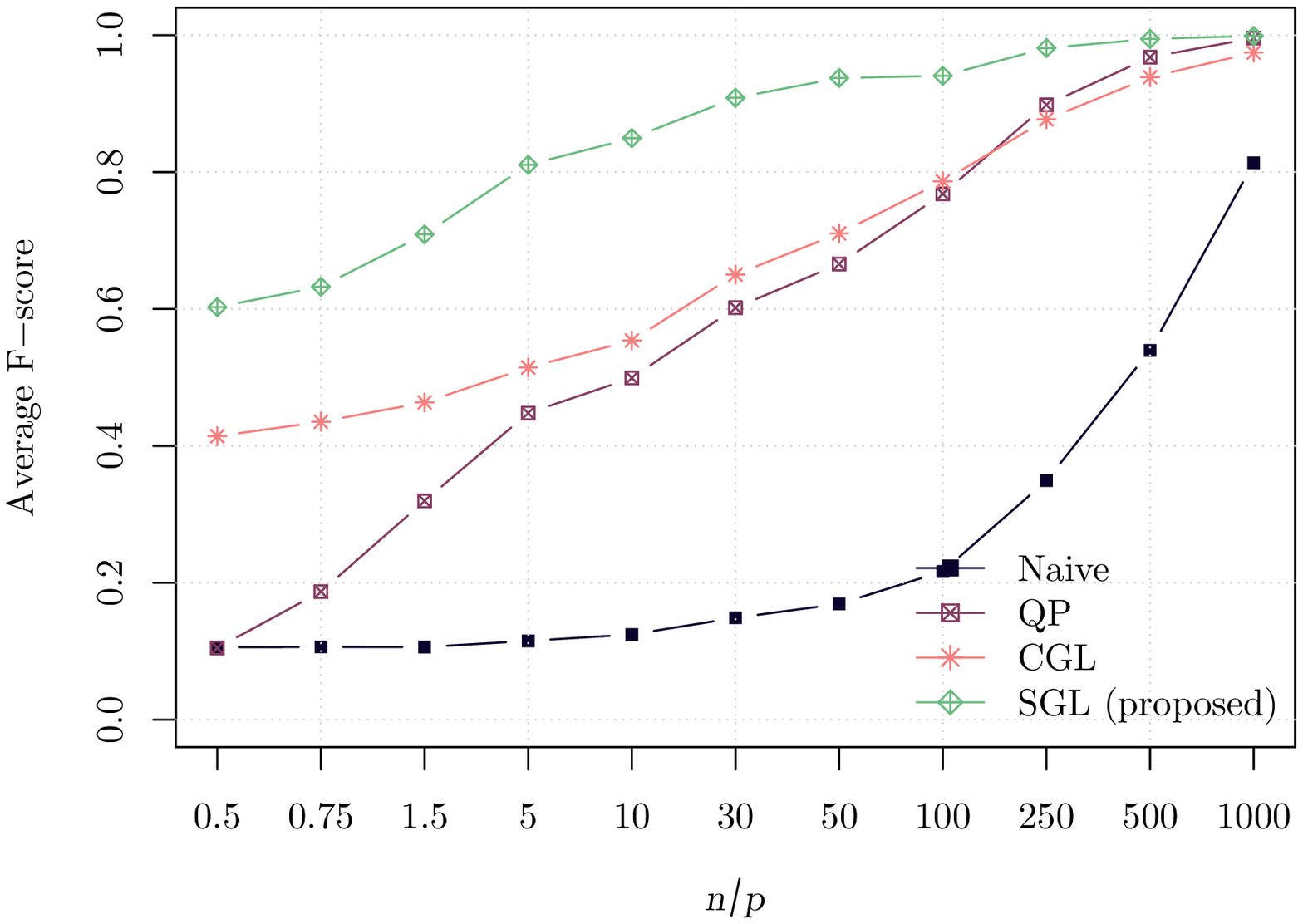}
	\end{subfigure}
	\caption{Average performance results for learning Laplacian matrix of a $\mathcal{G}_{\mathsf{grid}}$ graph. The \textsf{SGL} algorithm outperforms \textsf{Naive, QP}, and \textsf{CGL} for all the sample ratios. }
	\label{fig:performance-grid}
\end{figure}

Figure~\ref{fig:performance-grid} compares the performance of the algorithms for different sample size regimes on the grid graph model.
This is with respect to the number of data samples, used to calculate sample covariance $S$, per number of vertices (i.e., $n/p$), see \eqref{scm}. For $n/p <= 100$, we fix $\beta = 10$, otherwise we fix $\beta = 100$. Additionally, we fix $\alpha = 0$.
For \textsf{QP} and \textsf{Naive} we do not need to set any parameters. It is observed in Figure~\ref{fig:performance-grid}, the \textsf{SGL} algorithm significantly outperforms the baseline approaches: for all the sample ratios \textsf{SGL} can achieve a lower average \textsf{RE} and higher average \textsf{FS}. For instance, to achieve a low \textsf{RE} (e.g., 0.1), \textsf{SGL} requires a lower sample ratio ($n/p = 5$) than \textsf{Naive} ($n/p = 80$), \textsf{QP} ($n/p =29$) and \textsf{CGL} ($n/p = 30$).

\subsubsection{Modular graph}
\begin{figure}[!htb]
	\centering
	\begin{subfigure}[b]{0.45\textwidth}
		\includegraphics[width=1\textwidth]{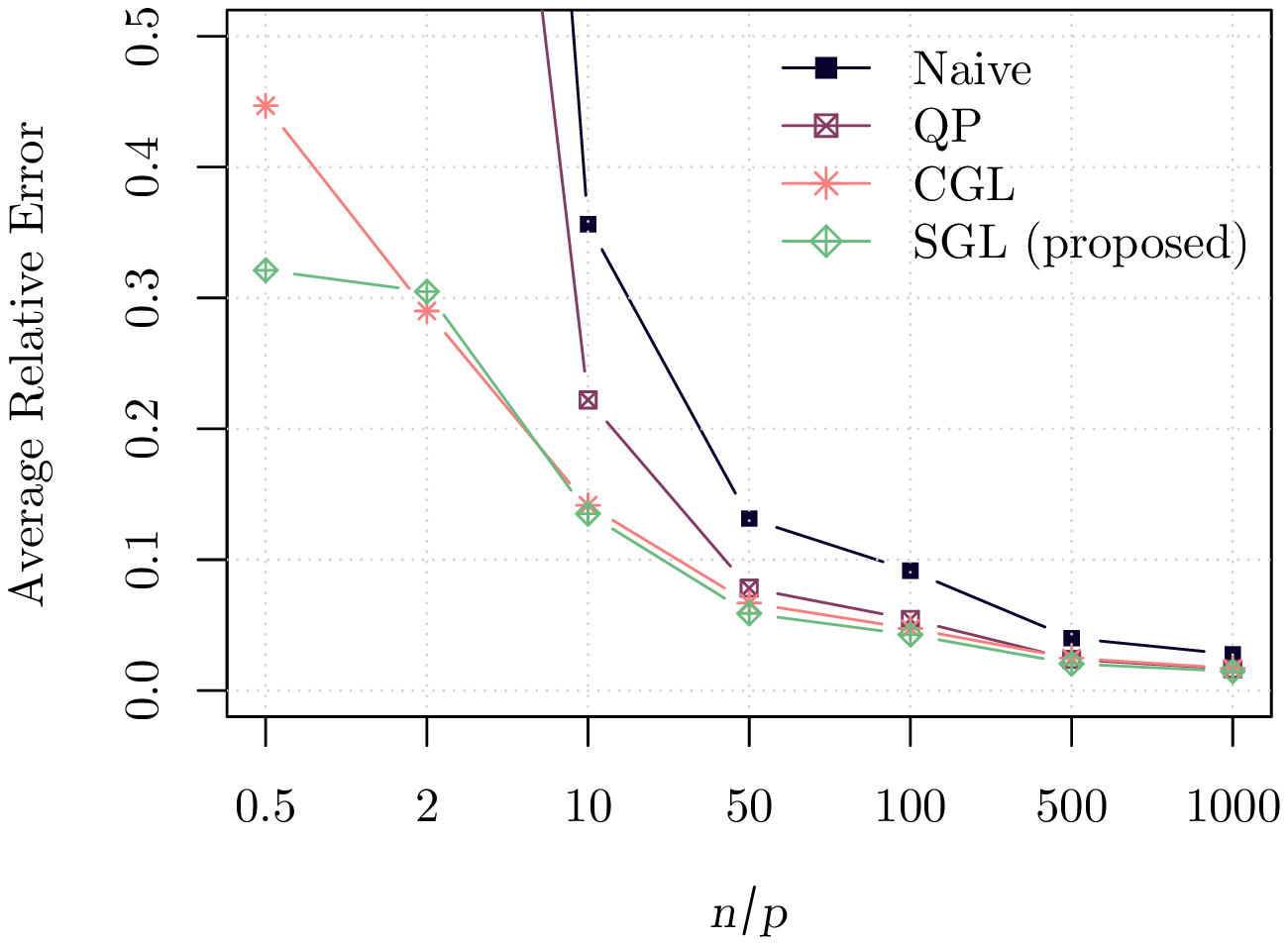}
	\end{subfigure}\qquad \quad
	~ 
	\begin{subfigure}[b]{0.45\textwidth}
		\includegraphics[width=1\textwidth]{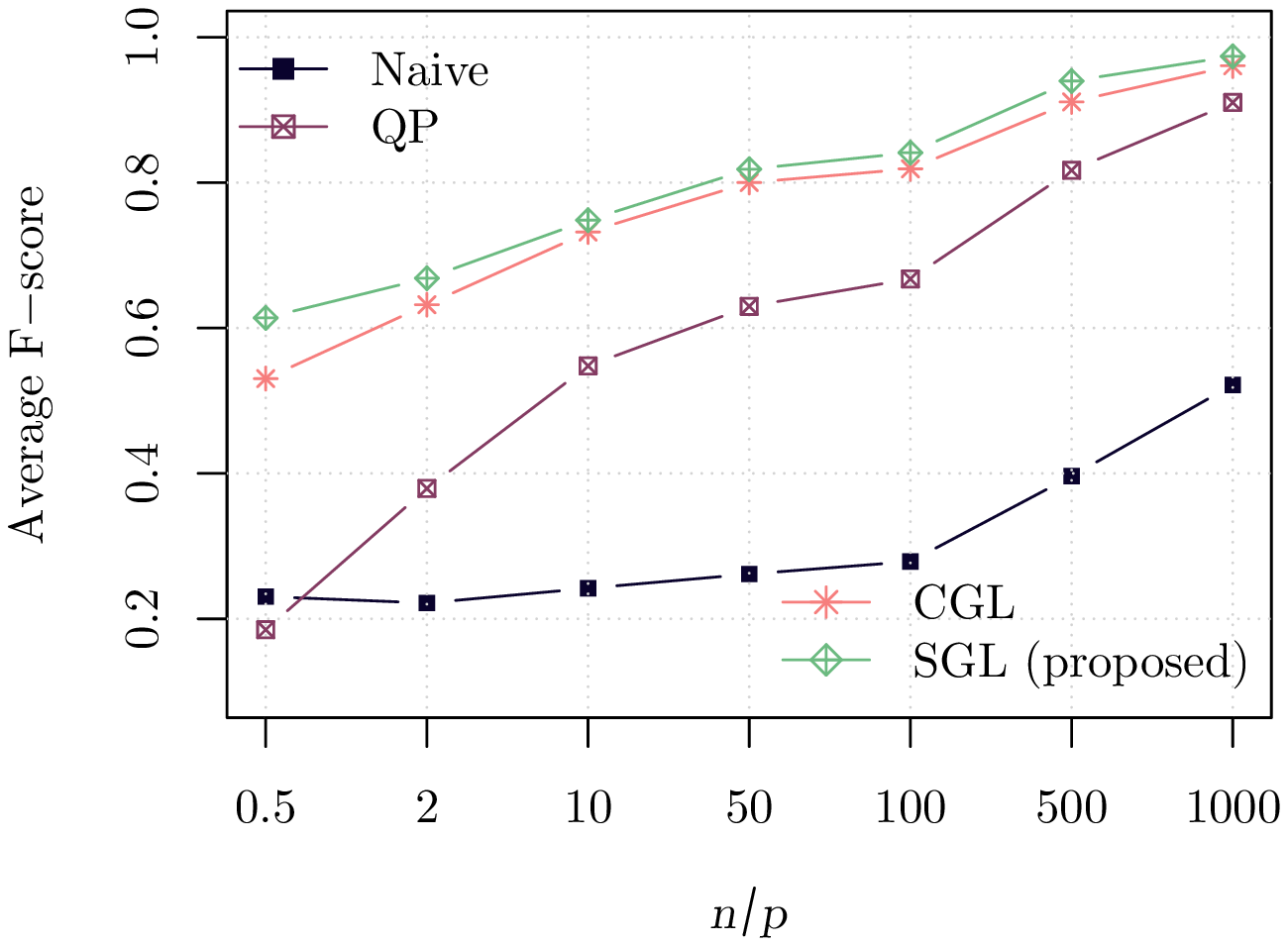}
	\end{subfigure}
	%	~
	%	\begin{subfigure}[b]{0.3\textwidth}
	%		\includegraphics[width=\textwidth]{examples/modular/27march/accuracy_modular.eps}
	%	\end{subfigure}
	\caption{Average performance results for learning Laplacian matrix of a modular graph $\mathcal{G}_{\mathsf{mo}}$ with four modules. The proposed $ \textsf{SGL}$ for $\beta=100, \alpha=0 $ method outperforms the base line approaches.}
	\label{fig:performance-modular}
\end{figure}

We consider a random modular graph, also known as stochastic block model, $ \G_{\mathsf{mo}}{(p,k, \wp_1,\wp_2)}$
with $p=64$ vertices and $k=4$ modules (subgraphs), where $\wp_1=0.01$ and $\wp_2=0.3$ are the probabilities of having an edge across modules and within modules, respectively. Edge weights are selected randomly uniformly from $[0.1,3]$. Figure~\ref{fig:performance-modular} illustrates the graph learning performances under different nodes to sample ratio ($n/p$). It is observed in Figure~\ref{fig:performance-grid}, the \textsf{SGL} and the \textsf{CGL} algorithm significantly outperforms the \textsf{Naive} and \textsf{QP}. Furthermore, for low sample ratio (i.e., $n/p<2$) \textsf{SGL} achieves better performance than \textsf{CGL}, while they perform similarly for a higher sample ratio (i.e., $n/p>2$).

%\subsubsection{Erdos-Renyi graph}
%
%
%\colr{We consider Erdos-Renyi graph structure denoted as $\mathcal{G}_{\mathsf{ER}}\ {(p, \wp)}$ with $p=64$ and $\wp = 0.1$, where $p$ is the number of nodes and $\wp$ is the probability of having an edge between any two pair of nodes. Figure~\ref{fig:performance-erdos-renyi} depicts the performance of algorithms for ER graph structure. We fix $\beta = 1.29$ and $\alpha = 0.0013$ for the values of $p$ such that $n/p> 1$, otherwise we fix $\alpha = 0$ and start with $\beta = 0.01$ and we exponentially increase it up to $\beta = 1$.}
%\begin{figure}[!htb]
%	\centering
%	\begin{subfigure}[b]{0.47\textwidth}
%		\includegraphics[width=\textwidth]{examples/erdos-renyi/relative_error_erdos_renyi.eps}
%	\end{subfigure}
%	~ %add desired spacing between images, e. g. ~, \quad, \qquad, \hfill etc.
%	%(or a blank line to force the subfigure onto a new line)
%	\begin{subfigure}[b]{0.47\textwidth}
%		\includegraphics[width=\textwidth]{examples/erdos-renyi/fscore_erdos_renyi.eps}
%	\end{subfigure}
%	\caption{
%		\colr{We will be updating this section with new plot. Ignore the mistake in legends.}Average performance results for learning Laplacian matrix of a $\mathcal{G}_{\mathsf{ER}}$. The \textsf{SGL} algorithm outperforms \textsf{Naive, QP}, and \textsf{CGL} for all the sample ratios.}
%	\label{fig:performance-erdos-renyi}
%\end{figure}

\subsubsection{Multi-component graph}
\begin{figure}
	\begin{subfigure}[b]{0.48\textwidth}
		\includegraphics[width=1\textwidth]{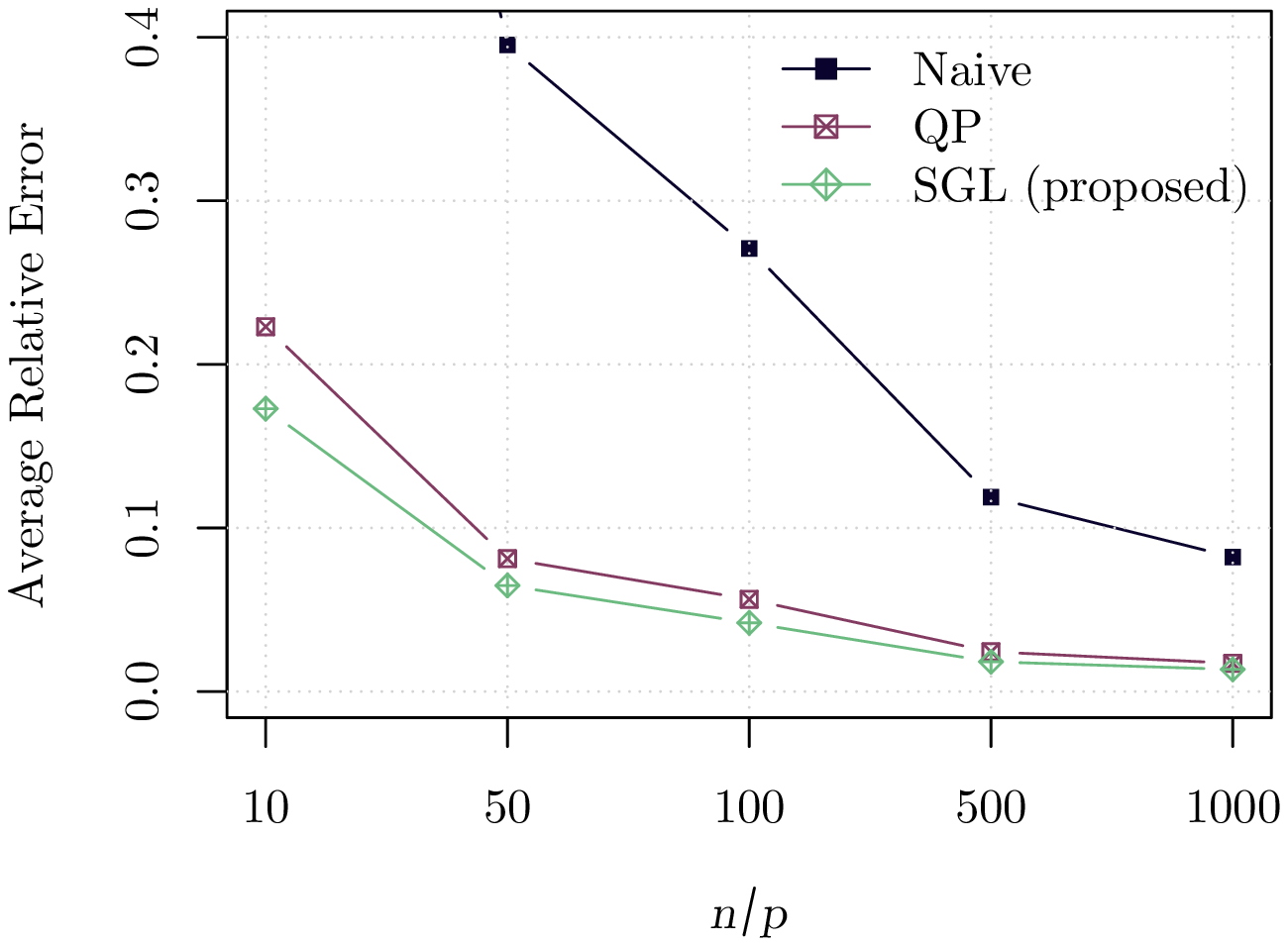}
%		\caption{Average Relative Error vs $n/p$}
	\end{subfigure}\quad 
	\begin{subfigure}[b]{0.48\textwidth}
		\includegraphics[width=1\textwidth]{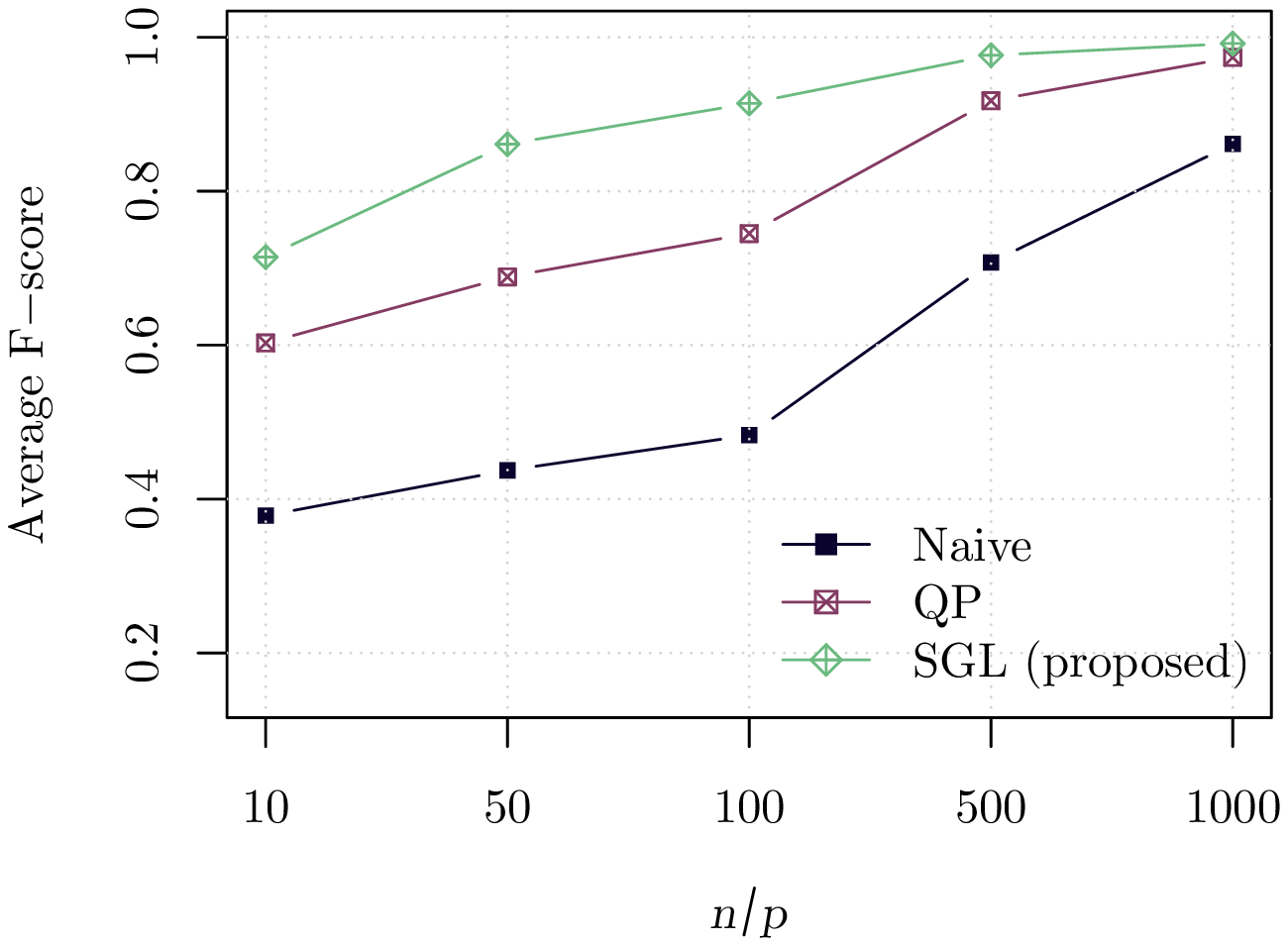}
%		\caption{Average F-score vs $n/p$}
	\end{subfigure}
%%	\begin{subfigure}[b]{0.45\textwidth}
%	\includegraphics[width=\textwidth]{examples/block-diagonal/conv1april/accuracy_blockdiag.eps}
%	\caption{Average F-score vs $n/p$}
%\end{subfigure}
	\caption{Average performance results as, a function of the number of samples, for learning
		Laplacian matrix of a 4-component graph. The SGL method demonstrates good performance for
		multi-component graph learning, and significantly outperforms the
		baseline approaches \textsf{Naive} and \textsf{QP}.}
	\label{fig:performance:n/p}
\end{figure}

 We consider to learn a multi-component graph also known as block-structured graph denoted as $\G_{\mathsf{mc}}(p,k,\wp)$, with $p=64$, $k=4$ and $\wp=0.5$, where $p$ is the number of nodes, $k$ is the number of components, and $\wp$ is the probability of having an edge between any two nodes inside a component while the probability of having an edge between any two nodes from different components is zero. Edge weights are selected randomly uniformly from $[0.1,3]$. Figure~\ref{fig:performance:n/p} illustrates the graph learning performances of different methods in terms of average \textsf{RE} and \textsf{FS}.

\subsubsection{Multi-component graph: noisy setting}
\begin{figure}[!htb]
	\centering
	\begin{subfigure}[b]{0.3\textwidth}
		\includegraphics[width=\textwidth]{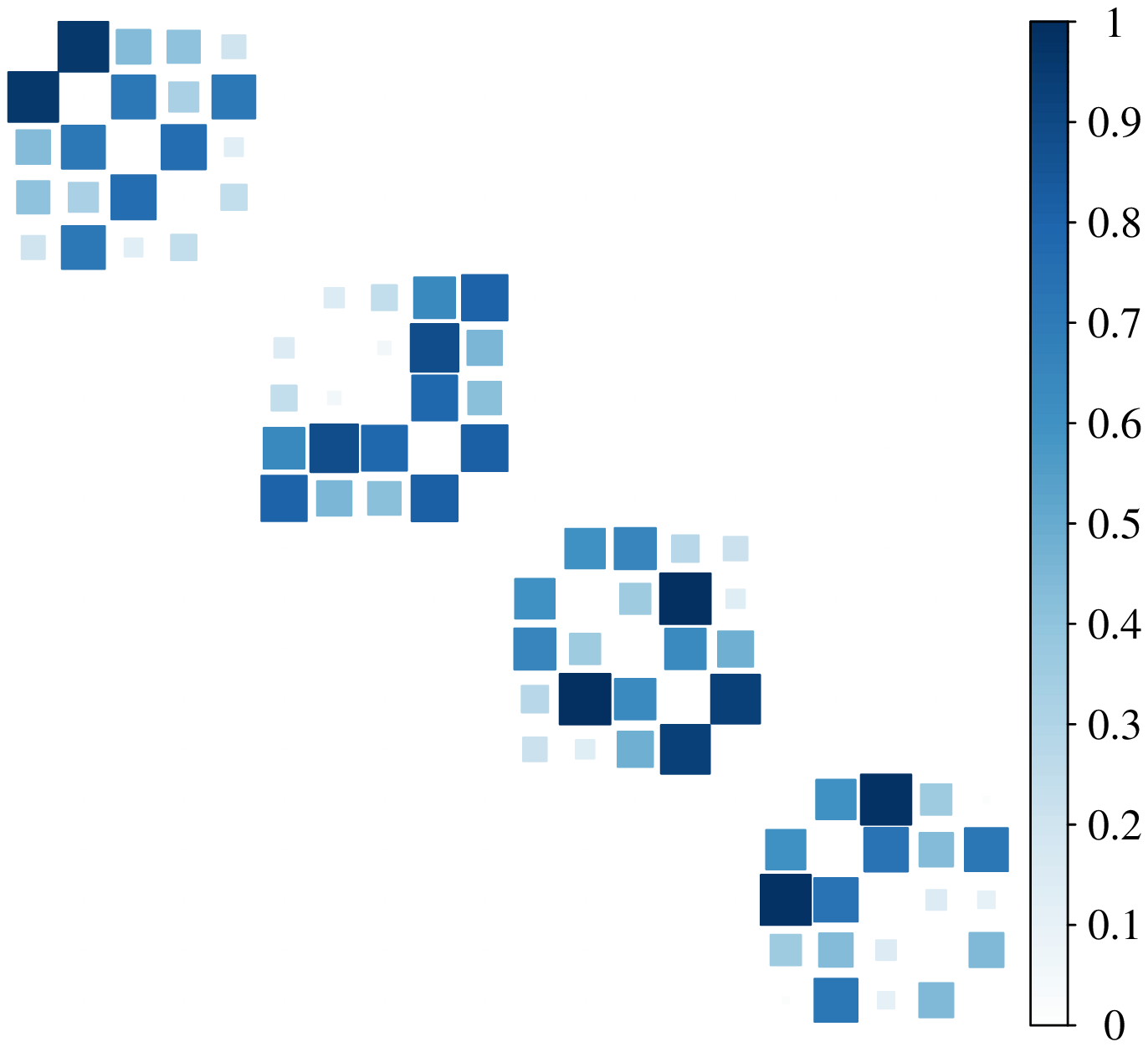}
		\caption{ $\Theta_{\textsf{true}}$ }
	\end{subfigure}
	~ %add desired spacing between images, e. g. ~, \quad, \qquad, \hfill etc.
	%(or a blank line to force the subfigure onto a new line)
	\begin{subfigure}[b]{0.3\textwidth}
		\includegraphics[width=\textwidth]{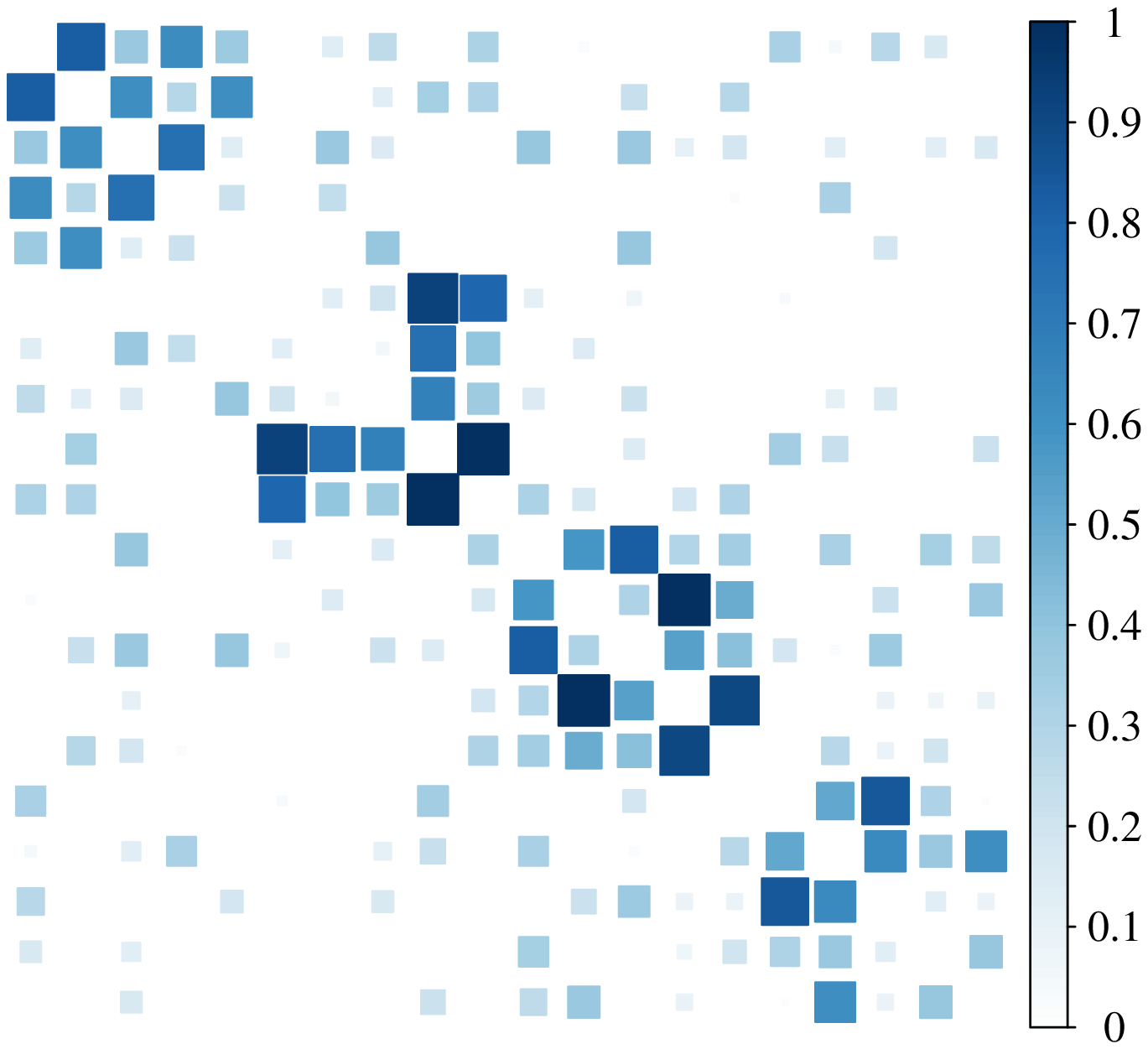}
		\caption{ $\Theta_{\textsf{noisy}}$ }
	\end{subfigure}
	~ %add desired spacing between images, e. g. ~, \quad, \qquad, \hfill etc.
	%(or a blank line to force the subfigure onto a new line)
	\begin{subfigure}[b]{0.3\textwidth}
		\includegraphics[width=\textwidth]{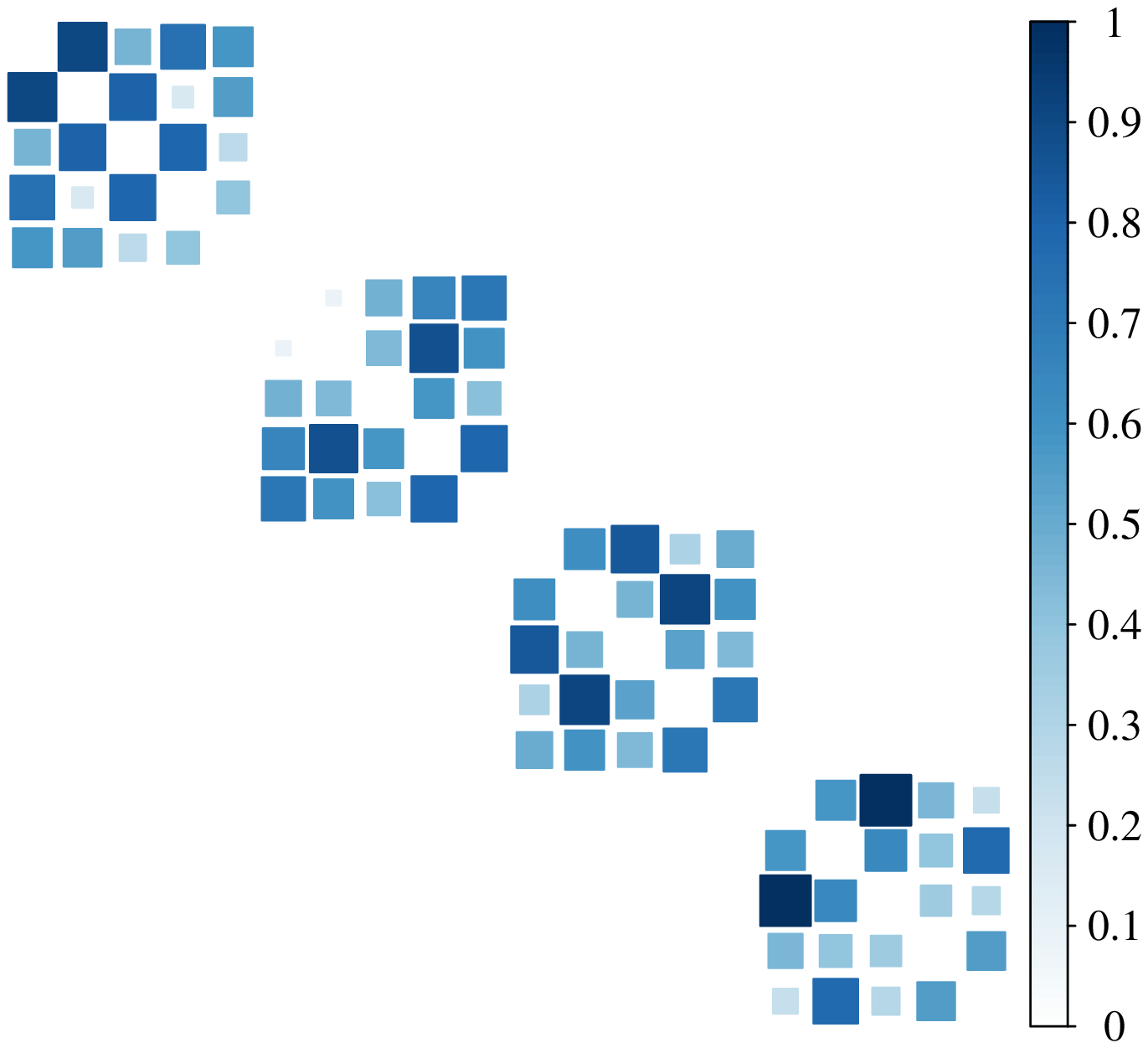}
		\caption{ $\Theta_{\textsf{learned}}$ }
	\end{subfigure}\\
	\vspace{1cm}
	\begin{subfigure}[b]{0.3\textwidth}
		\includegraphics[width=\textwidth]{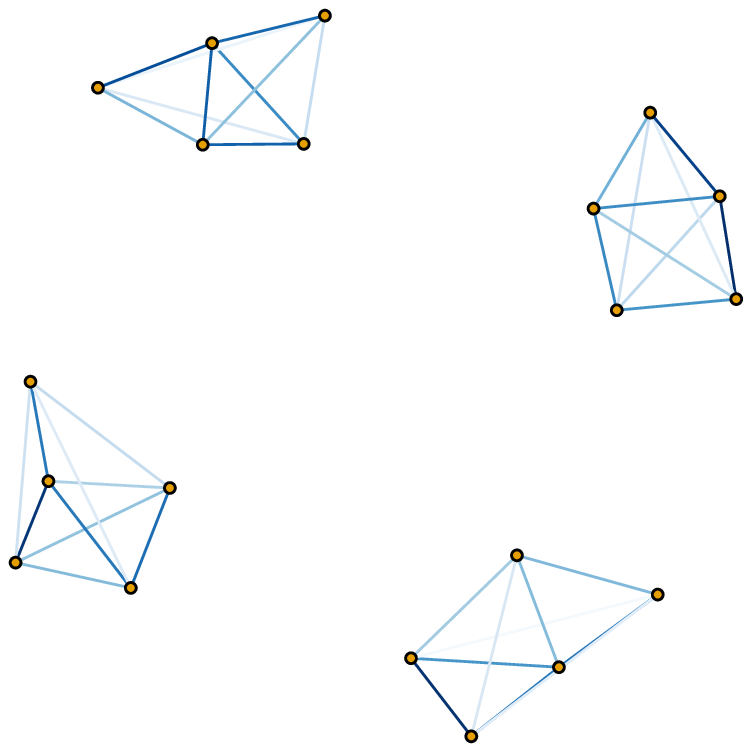}
	\caption{ $\G_{\textsf{true}}$ }
	\end{subfigure}
	~ %add desired spacing between images, e. g. ~, \quad, \qquad, \hfill etc.
	%(or a blank line to force the subfigure onto a new line)
	\begin{subfigure}[b]{0.3\textwidth}
		\includegraphics[width=\textwidth]{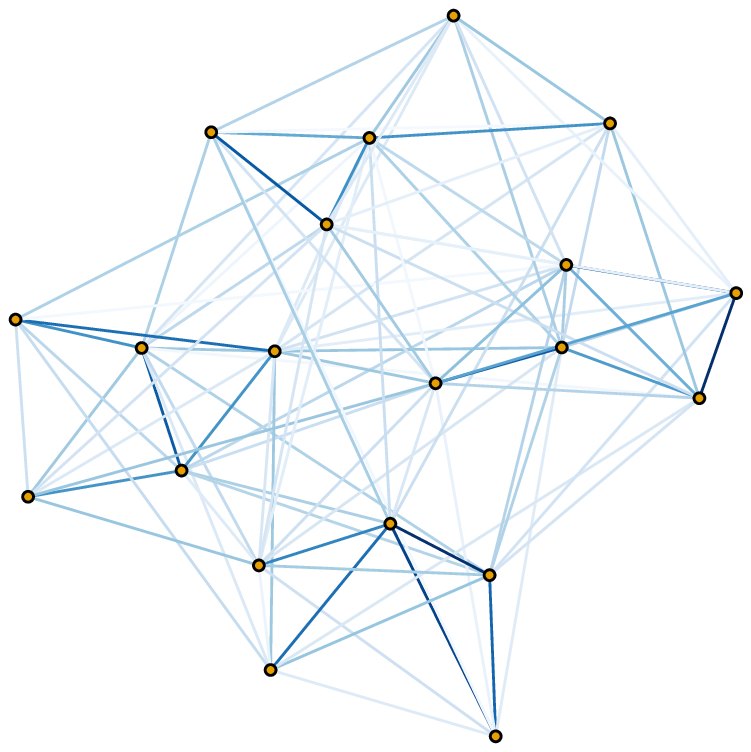}
	\caption{ $\G_{\textsf{noisy}}$ }
	\end{subfigure}
	~ %add desired spacing between images, e. g. ~, \quad, \qquad, \hfill etc.
	%(or a blank line to force the subfigure onto a new line)
	\begin{subfigure}[b]{0.3\textwidth}
		\includegraphics[width=\textwidth]{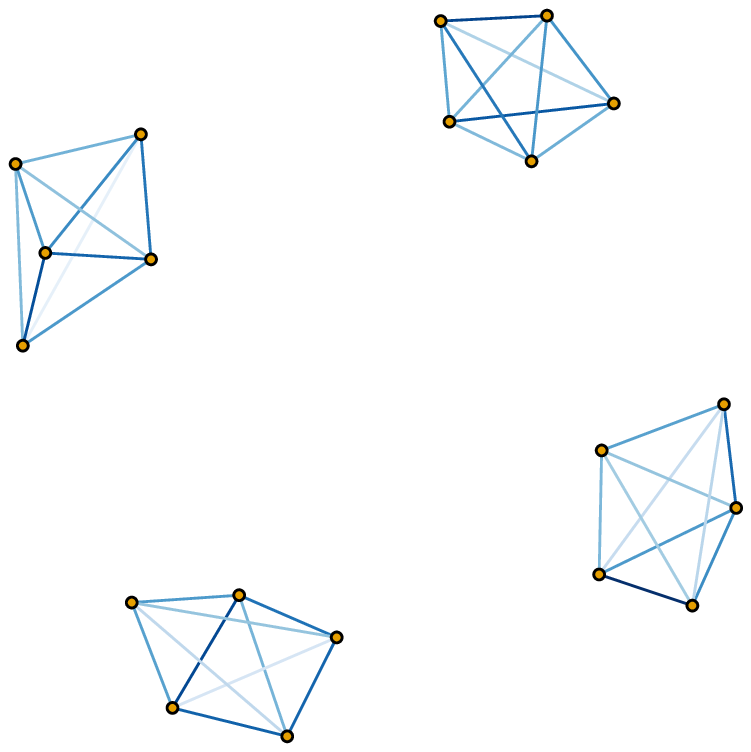}
	\caption{ $\G_{\textsf{learned}}$ }
	\end{subfigure}
	\caption{An example of estimating a 4-component graph. Heat maps of the graph matrices: (a) the ground truth graph Laplacian matrix $\bTheta_{\mathsf{true}}$, (b) $\bTheta_{\mathsf{noisy}}$ after being corrupted by noise, (c) $\Theta_{\textsf{learned}}$ the learned graph Laplacian with a performance of	$(\mathsf{RE}, \mathsf{FS}) = (0.210, 1)$, which means a perfect structure recovery even in a noisy setting that heavily suppresses the ground truth weights. The panels (d), (e), and (f) correspond to the graphs represented by the Laplacian matrices in (a), (b), and (c), respectively. }
	\label{fig:4-comp}
%	\label{fig:4-comp}
\end{figure}

Here we aim to learn a multi-component graph under noisy setting. At first we generate a 4 component graph $\G_{\textsf{mc}}(20,4, 1)$ with equal number of nodes in all the components, the nodes inside a component are fully connected and the edges are drawn randomly uniformly from $[0,1]$. Then we add random noise to all the in-component and out component edges. The noise is an Erdos-Renyi graph $\mathcal{G}_{\mathsf{ER}}(p,\wp)$, where $p=20$ is the number of nodes, $\wp=0.35$ is the probability of having an edge between any two pair of nodes, and edge weights are randomly uniformly drawn from $[0,\kappa]$. Specifically, we consider a scenario where each sample $\x_i \sim \mathcal{N}(\bzero, \bTheta_{\mathsf{noisy}}^\dagger)$ used for calculating SCM as in \eqref{scm} is drawn from the noisy precision matrix, 
\begin{align}\label{noisy-block}
\bTheta_{\mathsf{noisy}} = \bTheta_{\mathsf{true}} + \bTheta_{\mathsf{ER}},
\end{align}
 where $\bTheta_{\mathsf{true}}$ is the true Laplacian matrix and $\bTheta_{\mathsf{ER}}$ is the noise Laplacian matrix, which follows the ER graph structure. 
Figure~\ref{fig:4-comp} illustrates an instance of the \textsf{SGL} performance for noisy-multi component graph with fixed $n/p=30$, $\beta=400$, $\alpha=0.1$, and $\kappa=0.45$. 

%Figure~\ref{fig:performance:noise} depicts the average performance result for learning noisy multi-component graph matrix as a function of the noise factor. The value of $\kappa$ is responsible for the noise strength, higher values of $\kappa$ denotes highly noisy setting. The \textsf{SGL} algorithm is able to learn the graph structure faithfully even in highly noisy setting(e.g., \textsf{SGL} achieves a \textsf{FS} of $0.70$ for $\kappa=0.6$).
%\begin{figure}[!htb]
%		\begin{subfigure}[b]{0.45\textwidth}
%		\includegraphics[width=\textwidth]{examples/block-diagonal/relative_error_kappa.eps}
%		%	\caption{Relative error vs noise factor}
%	\end{subfigure}
%	\begin{subfigure}[b]{0.45\textwidth}
%		\includegraphics[width=\textwidth]{examples/block-diagonal/fscore_kappa.eps}
%	%	\caption{Average F-score vs noise factor}
%	\end{subfigure}
%	\caption{Average performance results, as a function of the noise factor, for learning noisy multi-component graph matrix, using the same settings of Figure \ref{fig:4-comp}, but for different $\kappa$. \textsf{SGL} is able to achieve a \textsf{FS} of $0.70$ even in highly noisy setting, i.e., $\kappa=0.6$}
%	\label{fig:performance:noise}
%\end{figure}

\subsubsection{Multi-component graph: components number mismatch}
For learning a multi-component graph structure,	\textsf{SGL} requires the knowledge of the number of components $k$, as a prior information, which is a common assumption for similar frameworks. If not available, one can infer $k$ by using existing methods for model selection e.g., cross validation, Bayesian information criteria (BIC), or Akaike information criteria (AIC). Furthermore, we also investigate the performance when accurate information about the true number of clusters is not available.

\begin{figure}[!htb]
	\centering
	\begin{subfigure}[b]{0.3\textwidth}
		\includegraphics[width=\textwidth]{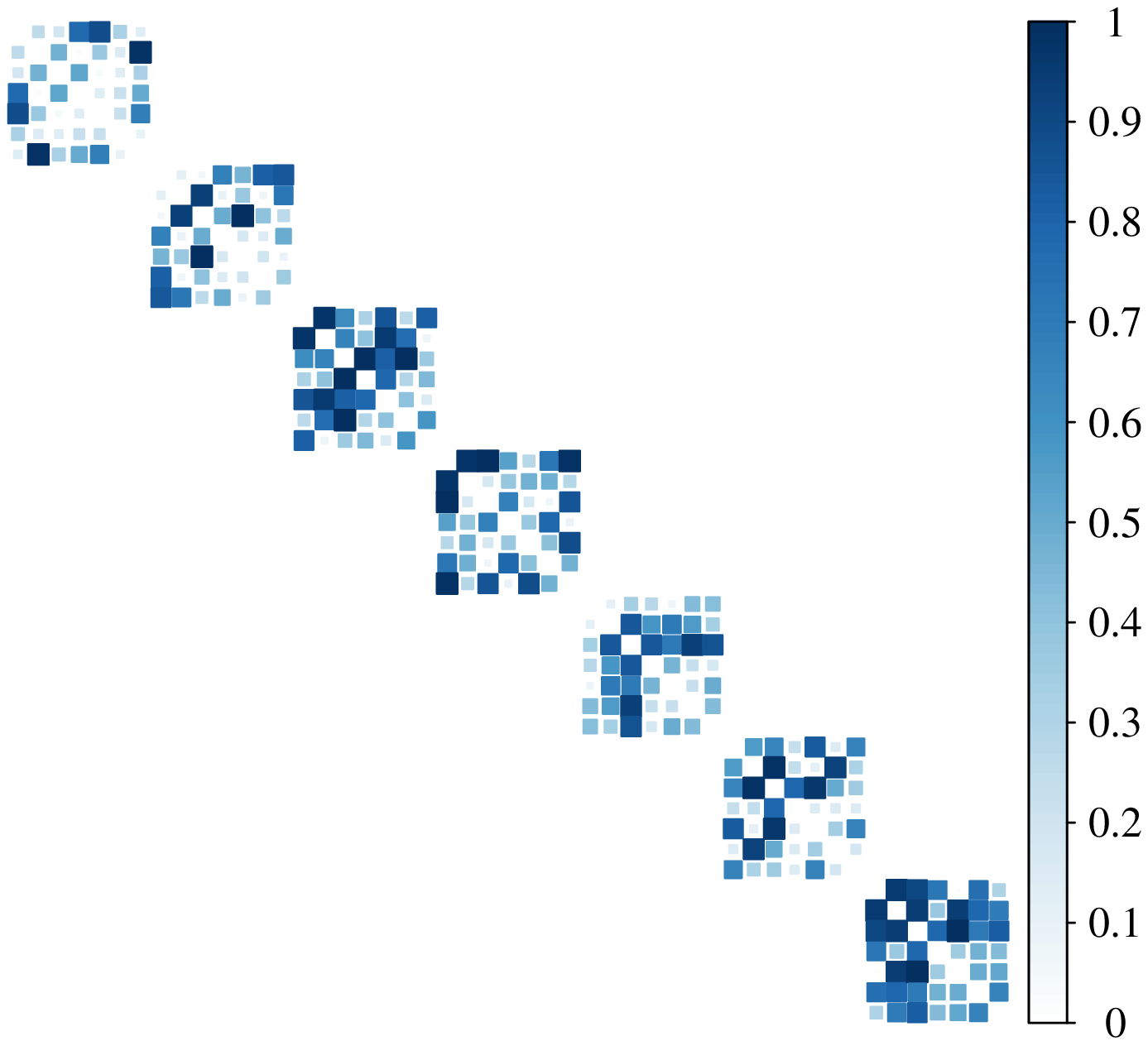}
		\caption{ $\Theta_{\textsf{true}}$ with $k=7$}
	\end{subfigure}
	~ %add desired spacing between images, e. g. ~, \quad, \qquad, \hfill etc.
	%(or a blank line to force the subfigure onto a new line)
	\begin{subfigure}[b]{0.3\textwidth}
		\includegraphics[width=\textwidth]{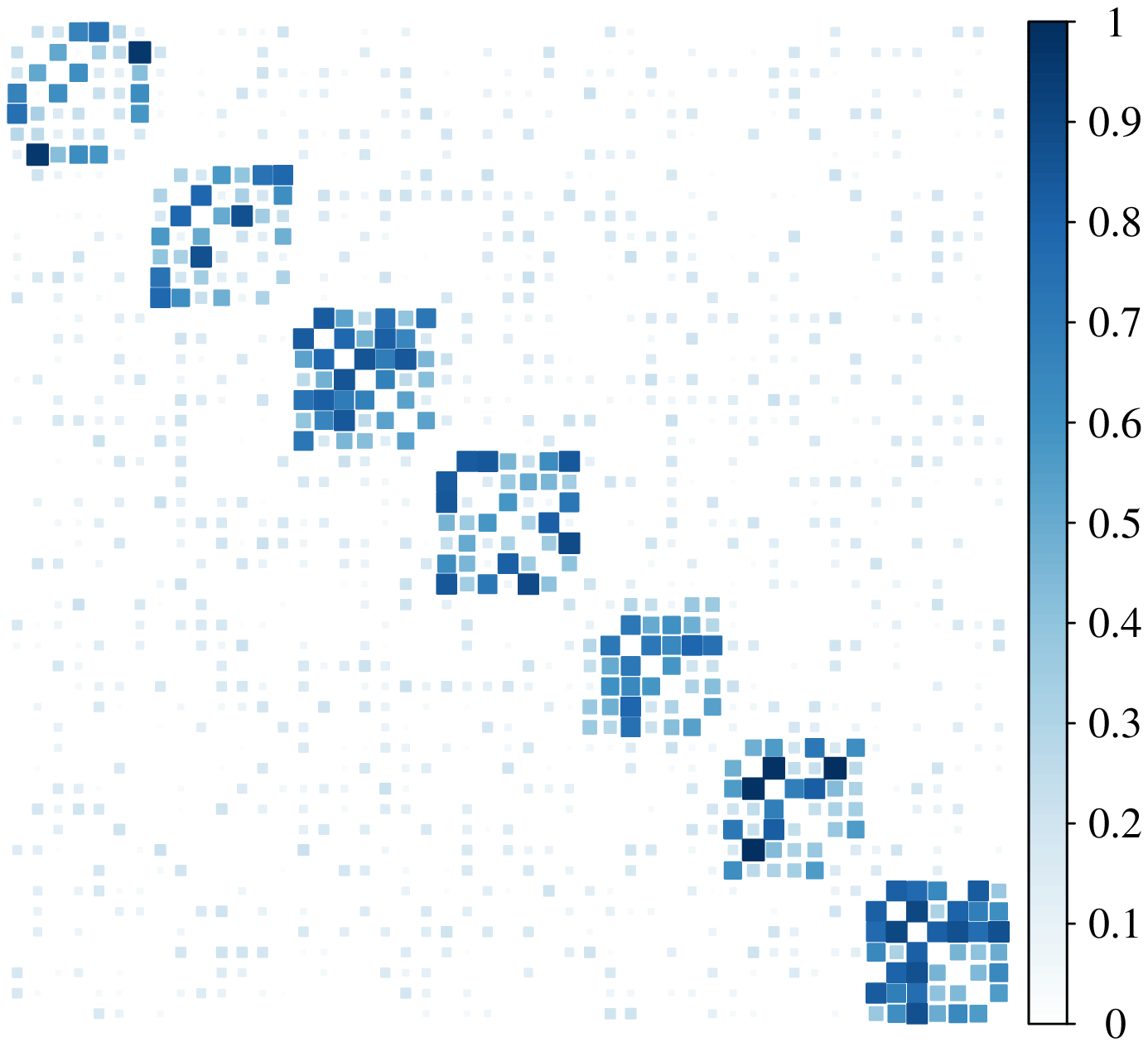}
	\caption{ $\Theta_{\textsf{noisy}}$}
	\end{subfigure}
	~ %add desired spacing between images, e. g. ~, \quad, \qquad, \hfill etc.
	%(or a blank line to force the subfigure onto a new line)
	\begin{subfigure}[b]{0.3\textwidth}
		\includegraphics[width=\textwidth]{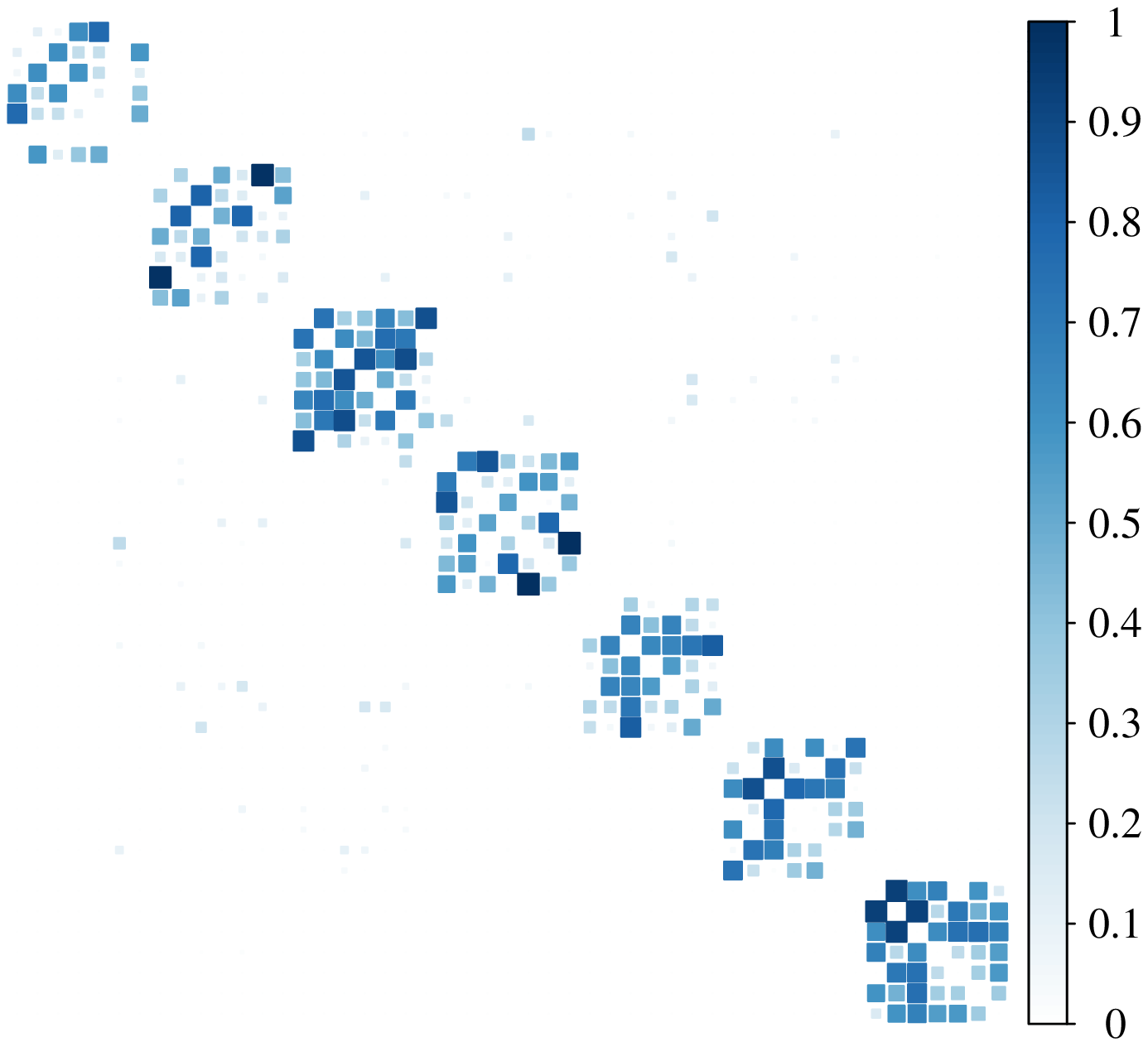}
	\caption{ $\Theta_{\textsf{learned}}$ with $k=2$}
	\end{subfigure}
	\\
	\vspace{1cm}	
	\begin{subfigure}[b]{0.3\textwidth}
		\includegraphics[width=\textwidth]{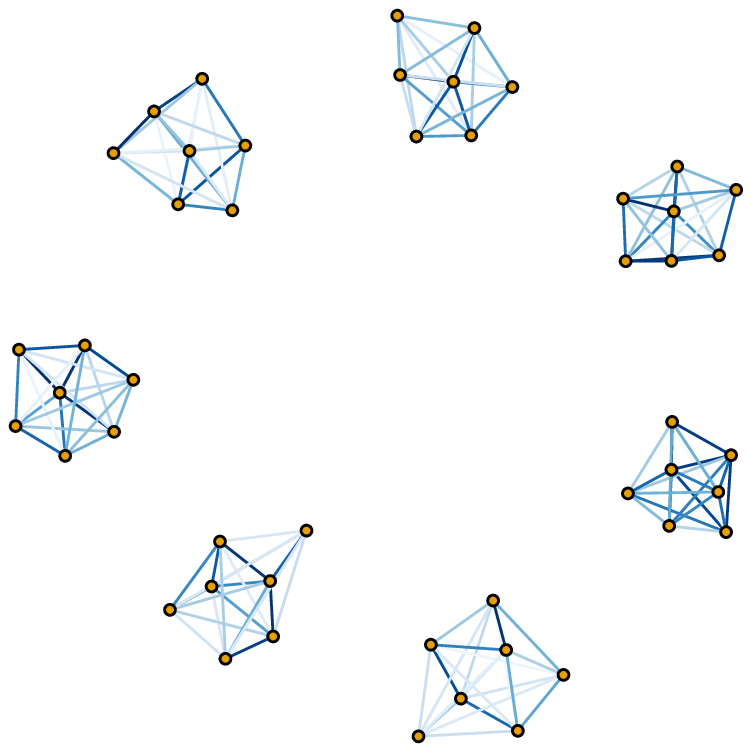}
	\caption{ $\G_{\textsf{true}}$ with $k=7$}
	\end{subfigure}
	~ %add desired spacing between images, e. g. ~, \quad, \qquad, \hfill etc.
	%(or a blank line to force the subfigure onto a new line)
	\begin{subfigure}[b]{0.3\textwidth}
		\includegraphics[width=\textwidth]{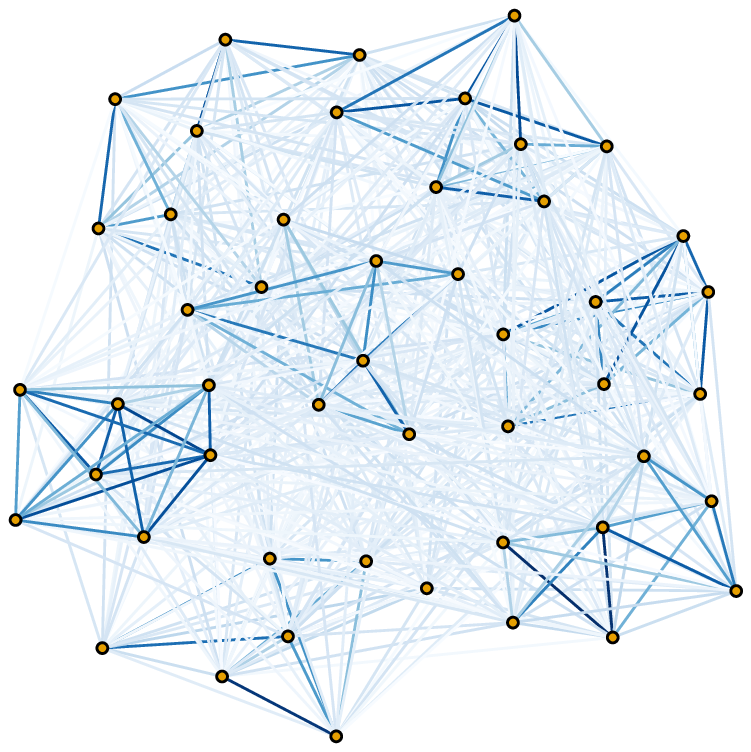}
		\caption{ $\G_{\textsf{noisy}}$ }
	\end{subfigure}
	~ %add desired spacing between images, e. g. ~, \quad, \qquad, \hfill etc.
	%(or a blank line to force the subfigure onto a new line)
	\begin{subfigure}[b]{0.3\textwidth}
		\includegraphics[width=\textwidth]{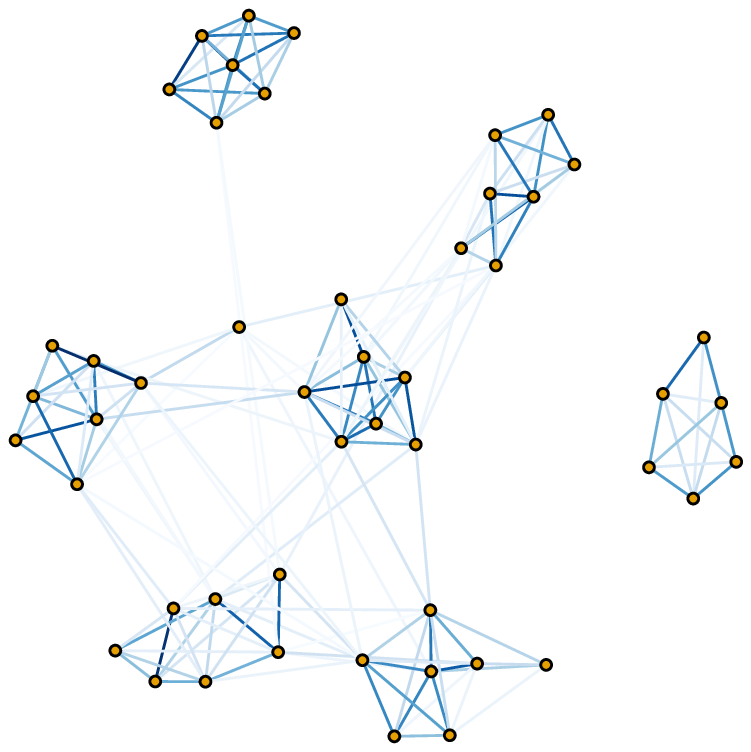}
	\caption{ $\G_{\textsf{learned}}$ with $k=2$}
	\end{subfigure}
\caption{Heat maps of the graph matrices: (a) the ground truth graph Laplacian of a seven-component graph $\bTheta_{\mathsf{true}}$, (b) $\bTheta_{\mathsf{noisy}}$ after being corrupted by noise, (c) $\bTheta_{\mathsf{learned}}$ the learned graph Laplacian with a performance of $(\mathsf{RE}, \mathsf{FS}) = (0.18, 0.81)$. The panels (d), (e), and (f) correspond to the graphs represented by the Laplacian matrices in (a), (b), and (c), respectively. For Figure \ref{fig:7-comp-graph} (c) and (f) we are essentially getting results corresponding to a two-component graph, which is imperative from the usage of spectral constraints of $k=2$. It is observed that the true graph (d) with $k=7$ components are contained exactly in the learned graph (f), the extra edges, which are due to the inaccurate spectral information when removed from \ref{fig:7-comp-graph} (f) can yield the true graph. One can use some simple post-processing techniques (e.g., thresholding of elements in the learned matrix $\Theta$), to recover the true component structure.}
	\label{fig:7-comp-graph}
\end{figure}

We consider an experiment involving model mismatch: the underlying Laplacian matrix that generates the data has $j$ number of components but we actually use $k$, $k \neq j$, number of components to estimate it. We generate a $k=7$ multi-component graph $\G_{\textsf{mc}}(49,7,1)$, the edge weights are randomly uniformly are drawn from $[0,1]$. Additionally, we consider a noisy model as in \eqref{noisy-block} i.e., $\bTheta_{\mathsf{noisy}} = \bTheta_{\mathsf{true}} + \bTheta_{\mathsf{ER}}$, where the noise is an Erdos-Renyi graph $\mathcal{G}_{\mathsf{ER}}(49,0.25)$ with edge randomly uniformly drawn from $[0, 0.45]$.
 Figure~\ref{fig:7-comp-graph} shows an example where the underlying graph has seven components, and we apply the \textsf{SGL} algorithm with $j = 2$. As we can see, even though the number of components is mismatched and the data is noisy, the \textsf{SGL} algorithm is still able to identify the true structure with a reasonable performance in terms of F-score and average relative error. 
 The take away from Figure~\ref{fig:7-comp-graph} is that, even in the lack of true information regarding the number of components in a graph, the graph learned from the \textsf{SGL} algorithm can yield an initial approximate graph very close to the true graph, which can be used as an input to other algorithms for post-processing to infer more accurate graph.
 % This can be beneficial in many applications, specially those in fully automated and unsupervised settings, the number of clusters is either unknown or it needs to be inferred from the data.

 \begin{figure}
	\centering
\includegraphics[width=.5\textwidth]{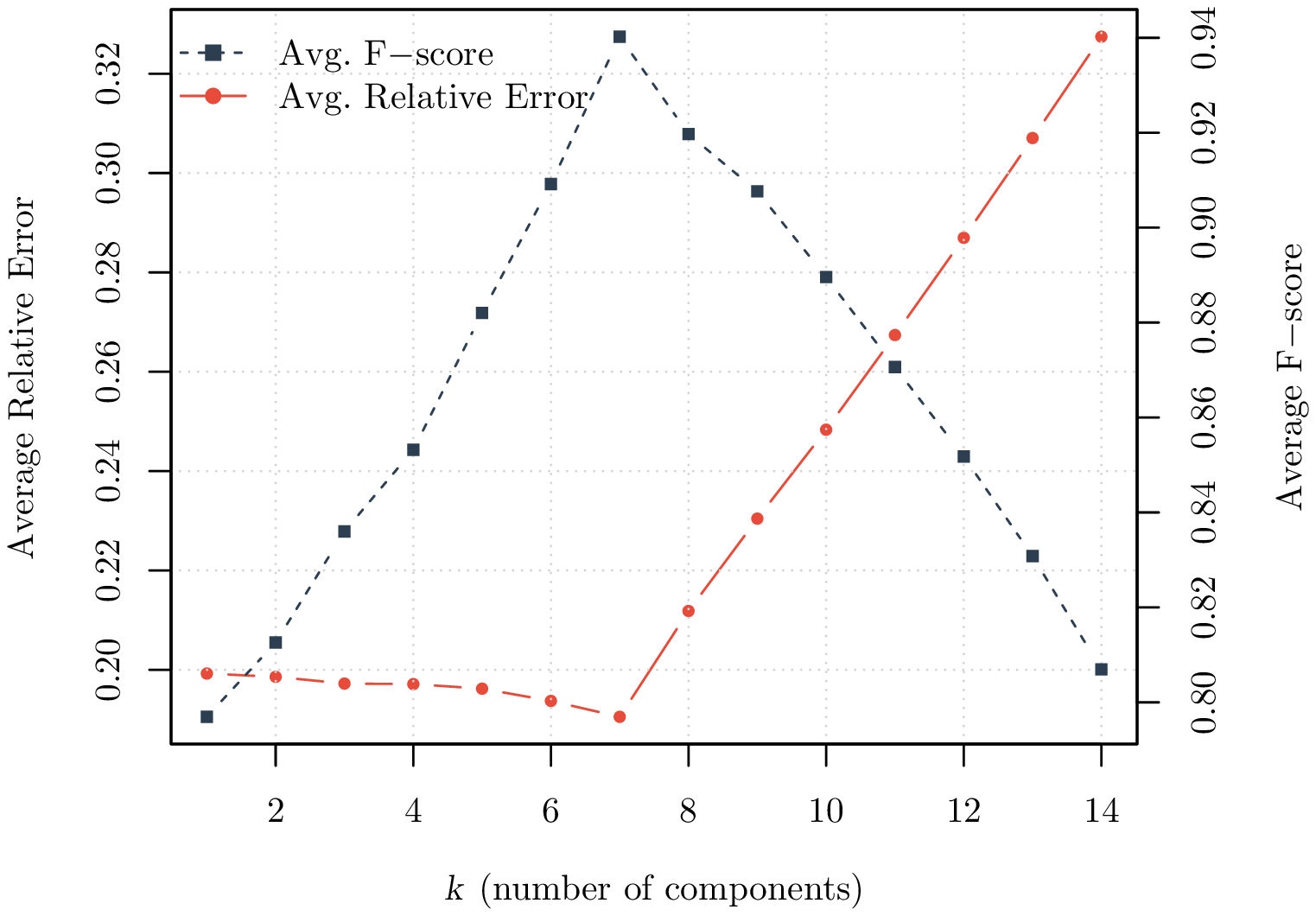}
\caption{Average performance results as a function of the number of components $k$: best results are obtained for true number of components. As we can note, the performance is monotonically increasing and eventually reaches a perfect F-score when $k = 7$. }
	\label{fig:fscore-k}
\end{figure}

Figure~\ref{fig:fscore-k} depicts the average performance of \textsf{SGL} as a function of $k$. The settings for the experiment is same as in Figure~\ref{fig:7-comp-graph}, except now we use different number of components information for each instances. It is observed that the \textsf{SGL} has its best performance when $k$ matches with the true number of the components in the graph. This also suggests that the \textsf{SGL} algorithm has the potential to be seamlessly integrated with model selection techniques to dynamically determine the number of clusters to use, in a single algorithm \citep{figueiredo2002unsupervised,schaeffer2007graph,fraley2007bayesian}.

\subsubsection{Popular multi-component structures}
\begin{figure}[!htb]
	\centering
\hspace{-1cm}	\begin{subfigure}[b]{0.3\textwidth}
		\includegraphics[width=\textwidth]{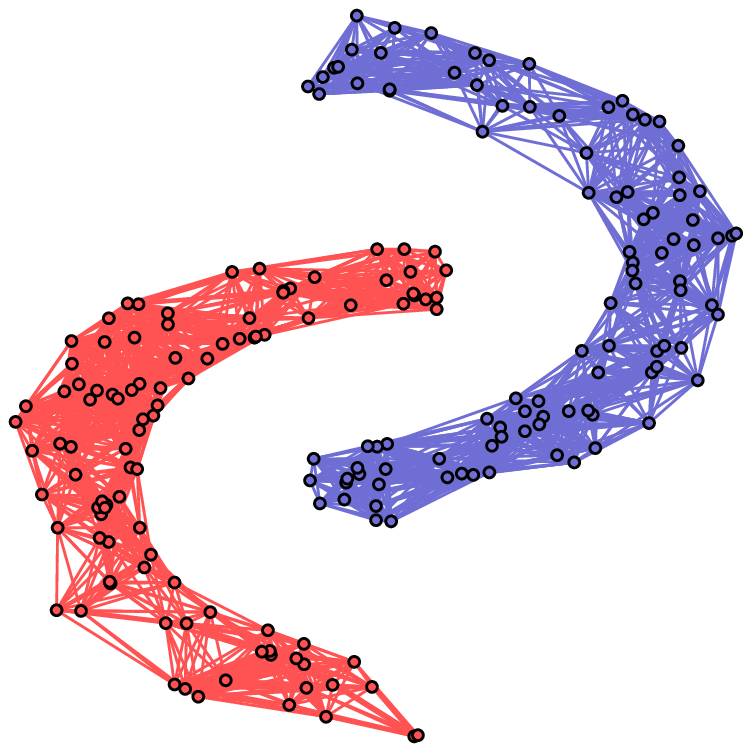}
		\caption{\textsf{two moons}}
	\end{subfigure}
	~ %add desired spacing between images, e. g. ~, \quad, \qquad, \hfill etc.
	%(or a blank line to force the subfigure onto a new line)
	\begin{subfigure}[b]{0.3\textwidth}
		\includegraphics[width=\textwidth]{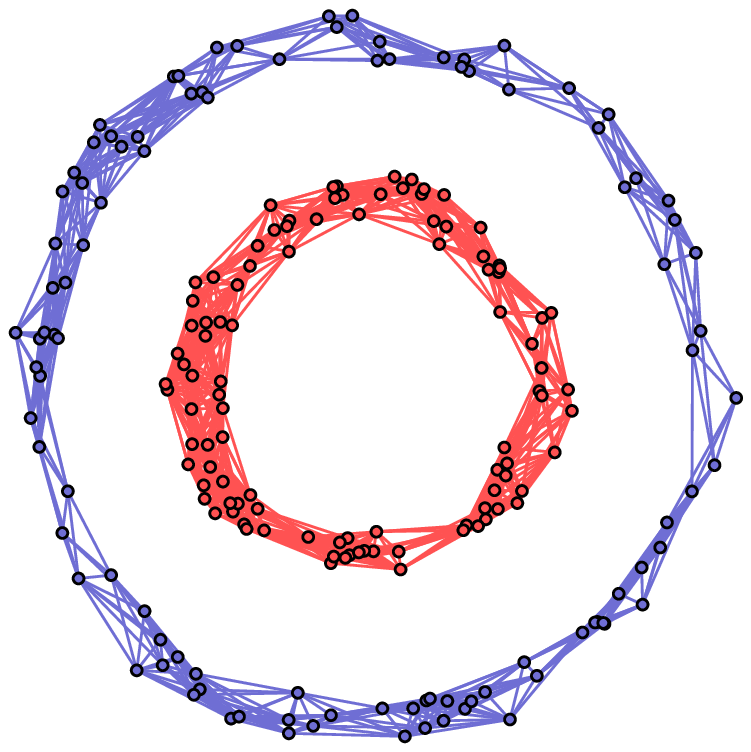}
		\caption{\textsf{two circles}}
	\end{subfigure}
	\begin{subfigure}[b]{0.3\textwidth}
	\includegraphics[width=\textwidth]{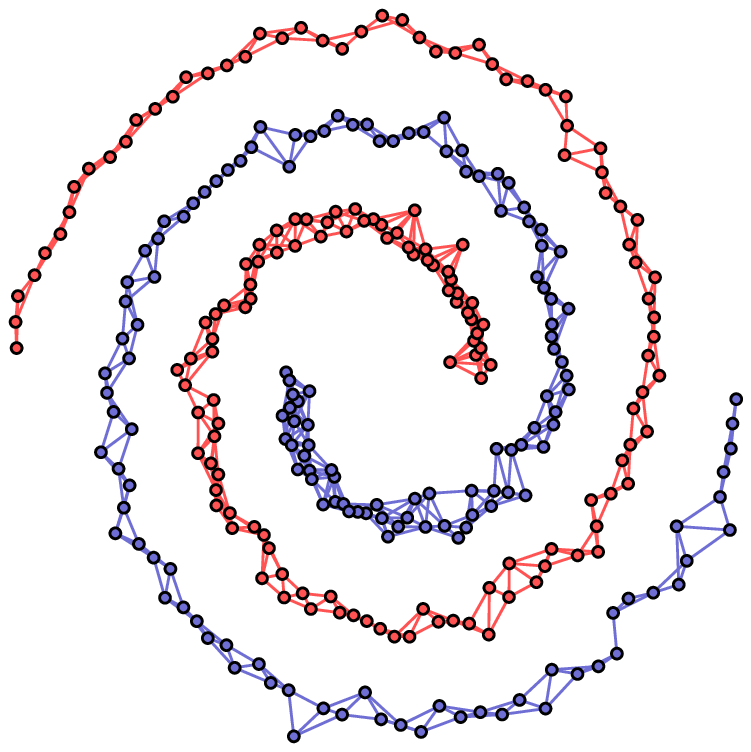}
	\caption{\textsf{three spirals}}
\end{subfigure}\\
\vspace{.4cm}
	\begin{subfigure}[b]{0.3\textwidth}
	\includegraphics[width=\textwidth]{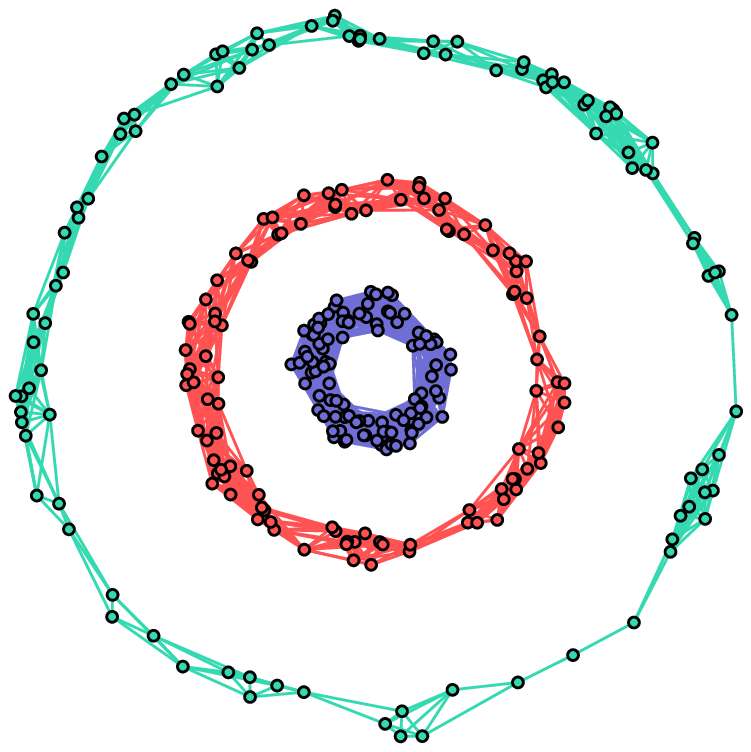}
	\caption{\textsf{three circles}}
\end{subfigure}
\begin{subfigure}[b]{0.3\textwidth}
	\includegraphics[width=\textwidth]{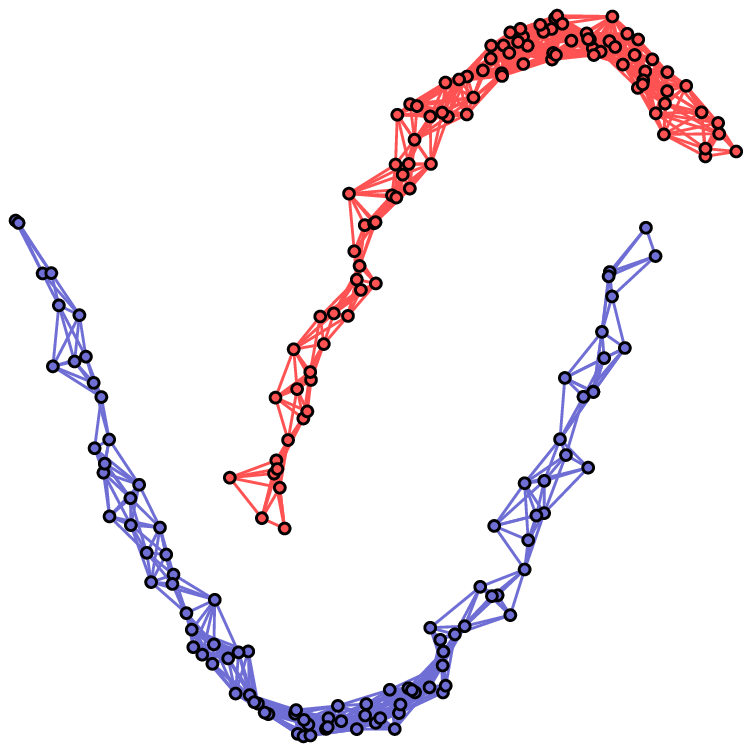}
	\caption{\textsf{worms}}
\end{subfigure}
\begin{subfigure}[b]{0.3\textwidth}
	\includegraphics[width=\textwidth]{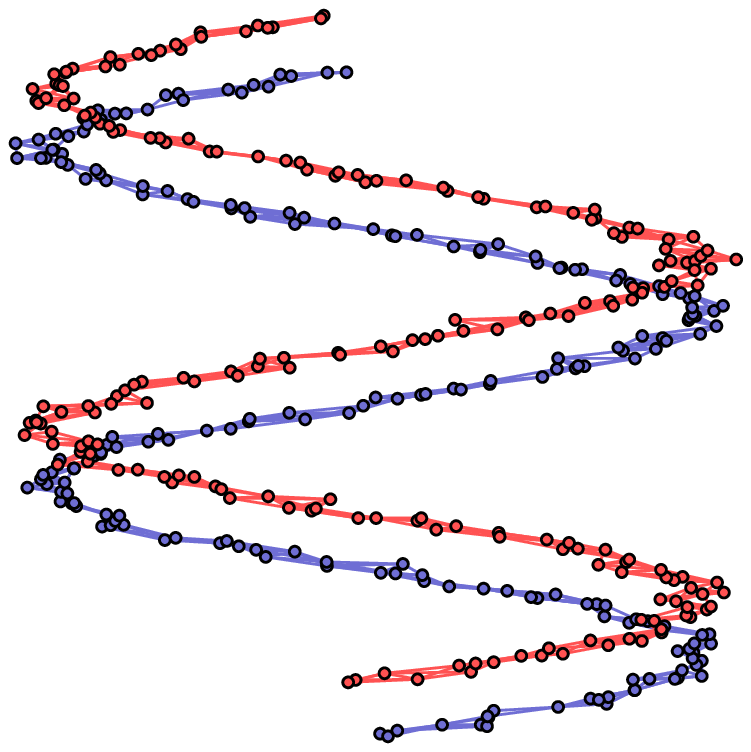}
	\caption{\textsf{helix 3d}}
	\label{fig:spiral_trend}
\end{subfigure}
\caption{\textsf{SGL} is able to perfectly classify the data points according to the cluster membership for all the structures.}
\label{fig:clusters}
\end{figure}

Here we consider the classical problem of clustering for some popular synthetic structures. To do that, we generate 100 nodes per cluster distributed according to structures colloquially known as \textit{two moons}, \textit{two circles}, \textit{three spirals}, \textit{three circles}, \textit{worms} and \textit{helix 3d}. Figure~\ref{fig:clusters} depicts the results of learning the clusters structures using the proposed algorithm \textsf{SGL}.

\subsubsection{Real data: animals data set}
\begin{figure}
	
		\begin{subfigure}[b]{0.45\textwidth}
		\includegraphics[width=1\textwidth]{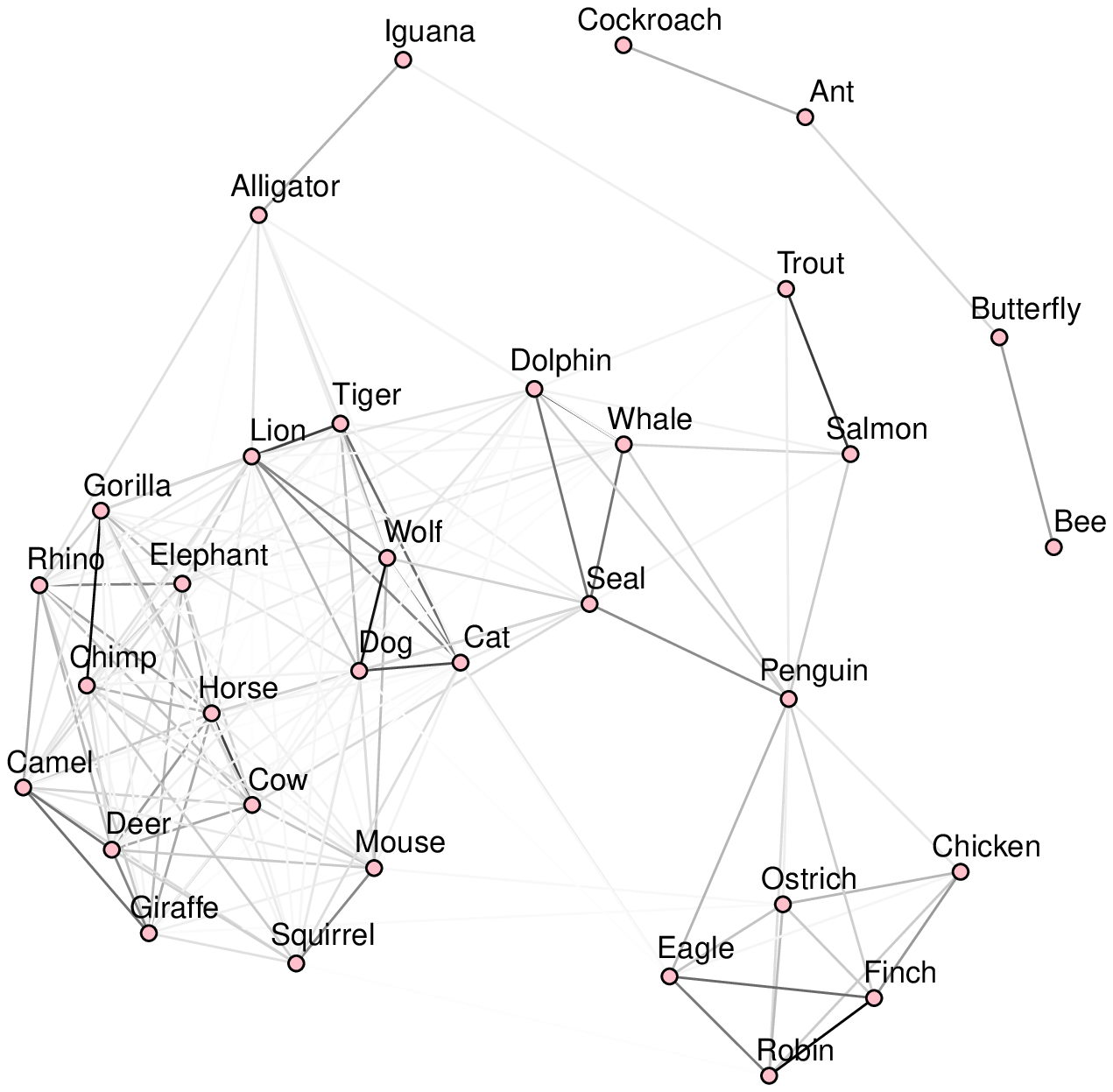}
		\caption{ \textsf{GLasso}\citep{friedman2008sparse}}
		\label{fig:animals-glasso}
	\end{subfigure} \qquad
~
	\begin{subfigure}[b]{0.45\textwidth}
		\includegraphics[width=1\textwidth]{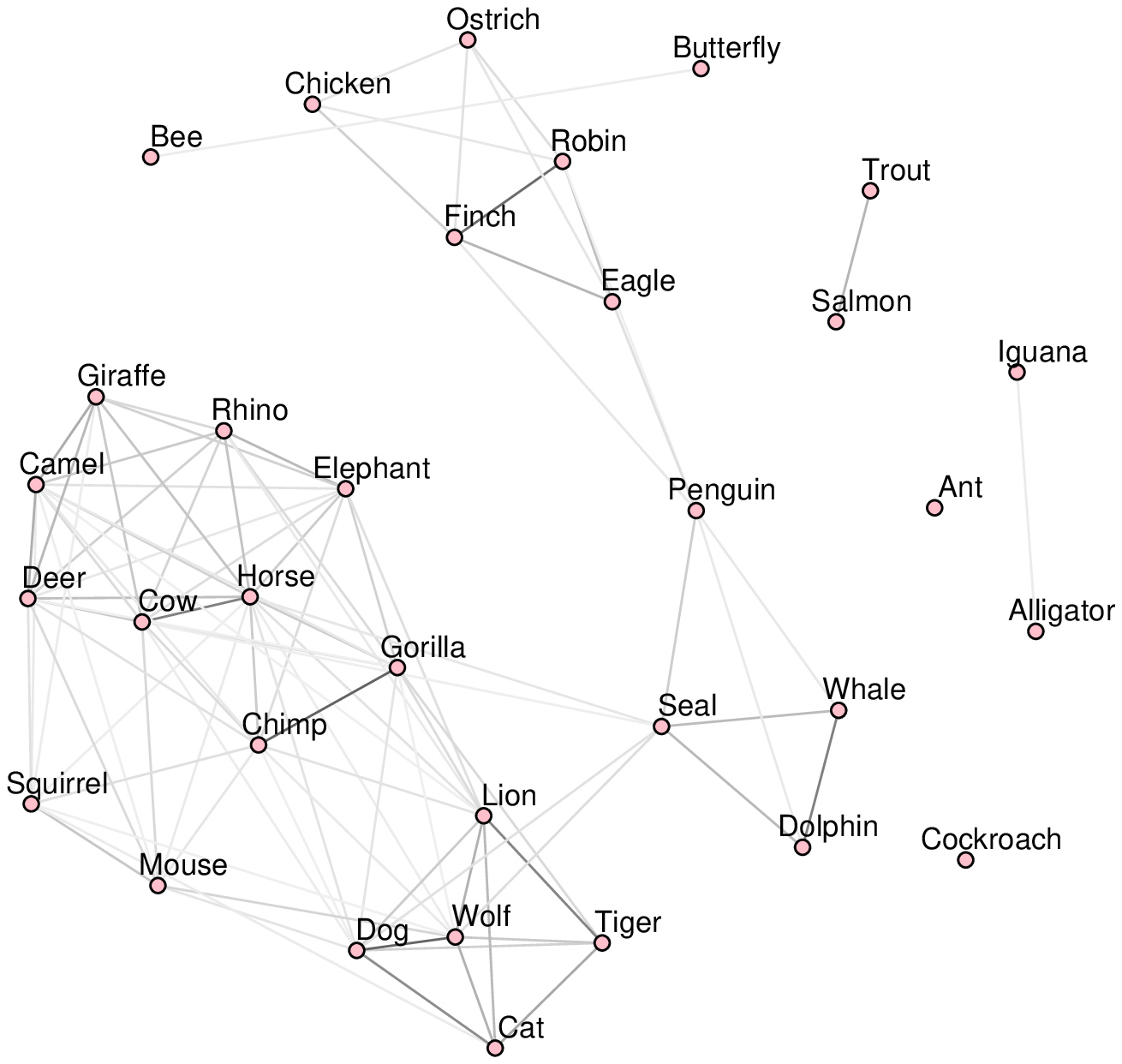}
		\caption{\textsf{GGL}\citep{egilmez2017graph}}
	\end{subfigure} 
	\\
	\vspace{.2cm}
	\begin{subfigure}[b]{0.45\textwidth}
		\includegraphics[width=1\textwidth]{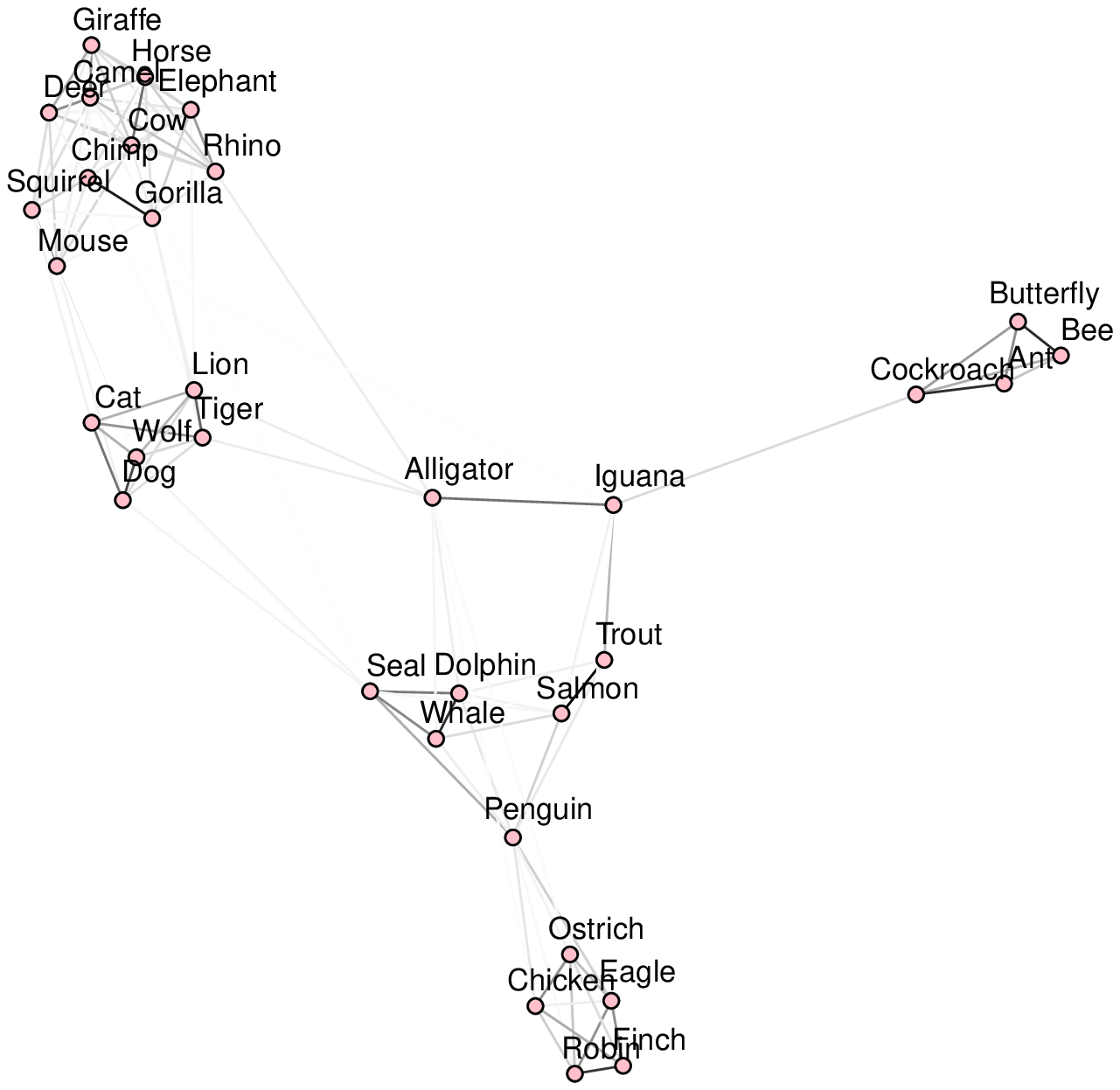}
		\caption{\textsf{SGL} with $k=1$(proposed)}
	\end{subfigure}\qquad 
~
	\begin{subfigure}[b]{0.45\textwidth}
	\includegraphics[width=1\textwidth]{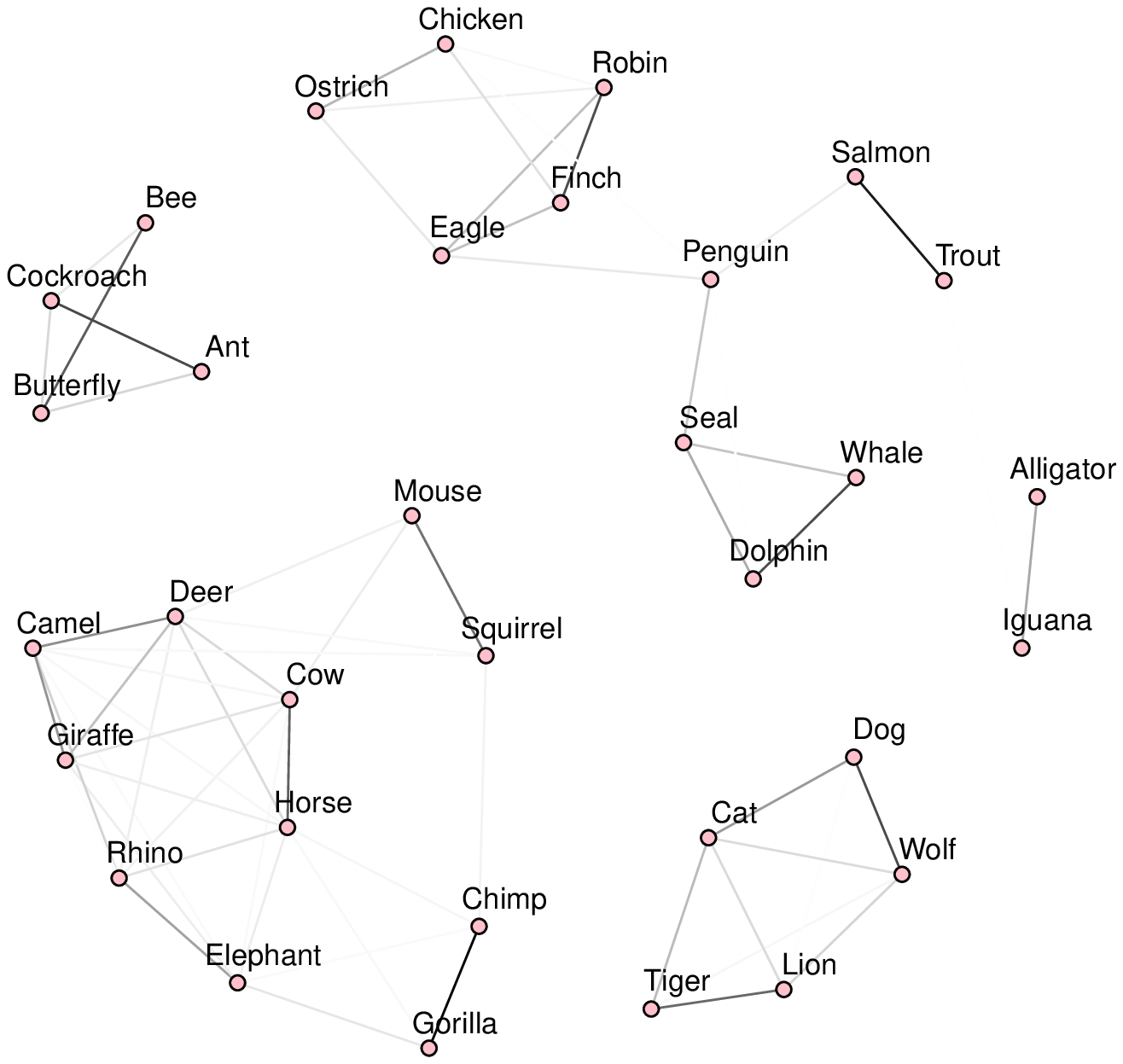}
	\caption{\textsf{SGL} with $k = 5$(proposed) }
 \end{subfigure}
\caption{Learning the connectivity of the \textsf{animals} data set with (a) \textsf{GLasso}, (b)\textsf{GGL}, (c) \textsf{SGL} with $k=1$ and (d) \textsf{SGL} with $k=5$. All methods obtain perceptual graphs of animal connections: darker edges denote stronger connections among animals. The methods (a) \textsf{GGL}, (b) \textsf{GLasso}, and (c) \textsf{SGL} $k=1$ were expected to obtain sparse-connected graphs. But, \textsf{GGL, GLasso} split the graph into multiple components due to the sparsity regularization. While \textsf{SGL} using sparsity regularization along with spectral constraint $k=1$(connectedness) yields a sparse-connected graph. (d) \textsf{SGL} with $k=5$ obtains a graph with component which depicts a more fine-grained representation of animal connection by grouping similar animals in respective components, highlighting the fact that the explicit control of the number of components may yield an improved visualization. Furthermore, the animal data is categorical (non-Gaussian) which does not follow the IGMRF assumption, the above result also establishes the capability of \textsf{SGL} under mismatch of the data model.}
\label{fig:animals}
\end{figure}

Herein, \textsf{animals} data set \citep{osherson1991default,lake2010discovering} is taken into consideration to learn weighted graphs. Graph vertices denote animals, and edge weights representing similarity among them. The data set consists of binary values (categorical non-Gaussian data) which are the answers to questions such as ``is warm-blooded?," ``has lungs?", etc. There are a total of 102 such questions, which make up the features for 33 animal categories. Figure~\ref{fig:animals} shows the results of estimating the graph of the \textsf{animals} data set using the \textsf{SGL} algorithm with $\beta = 1/2$ and $\alpha = 0$, \textsf{GGL} with $\alpha = 0.05$ and \textsf{GLasso} $\alpha = 0.05$. The input for all the algorithms is the sample covariance matrix plus an identity matrix scaled by $1/3$ (see \citet{egilmez2017graph}). The evaluation of the estimated graphs is based on the visual inspection. It is expected that similar animals such as (\textit{ant}, \textit{cockroach}), (\textit{bee}, \textit{butterfly}), and (\textit{trout}, \textit{salmon}) would be grouped together. Based on this premise, it can be seen that the \textsf{SGL} algorithm yields a more clear graph than the ones learned by \textsf{GGL} and \textsf{GLasso}. 

\begin{figure}[!htb]
	\centering
\begin{subfigure}[b]{0.375\textwidth}
	\includegraphics[width=1\textwidth]{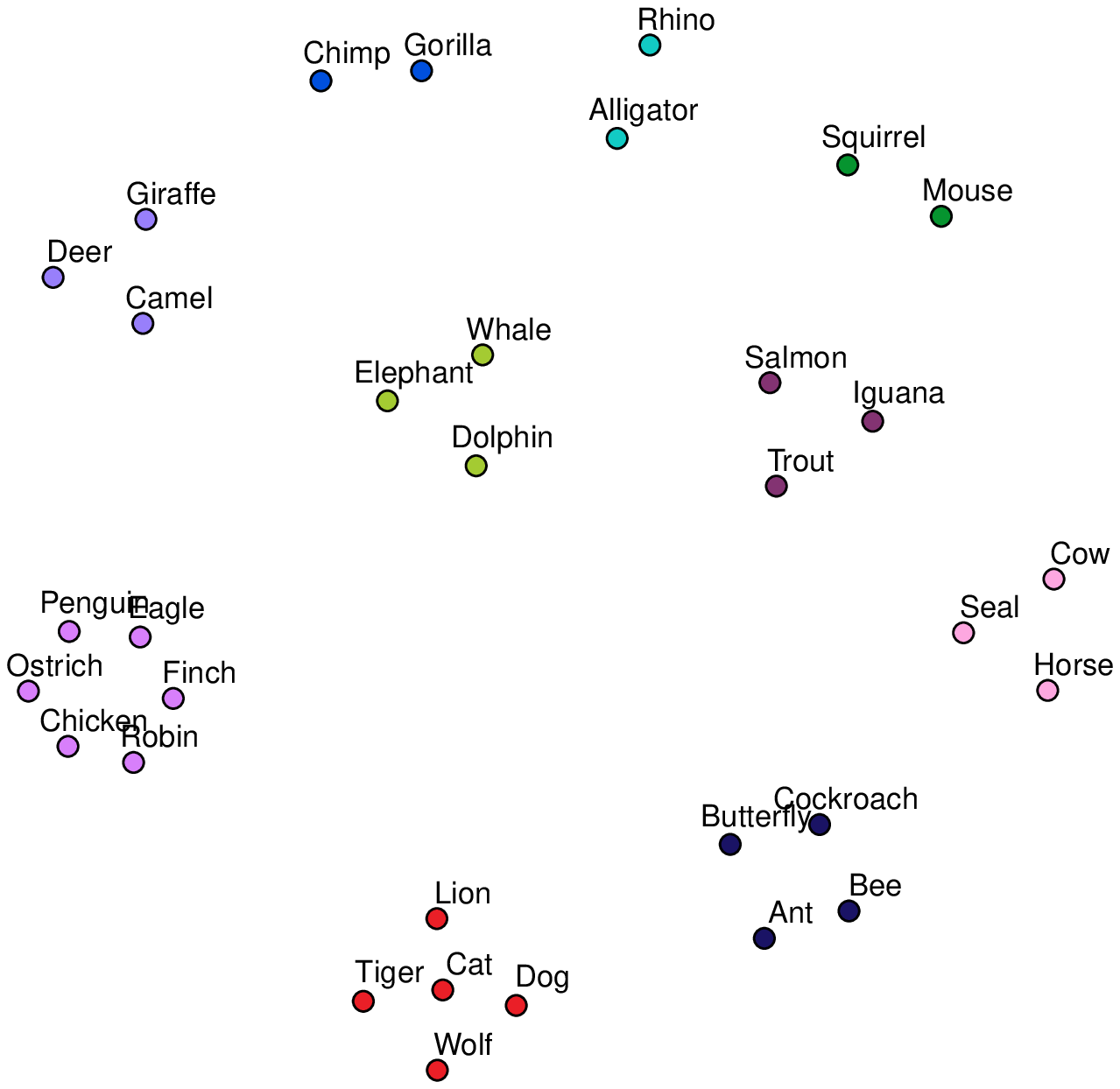}
	\caption{$k-$\textsf{means} }
\end{subfigure}
\hspace{1.6cm}
~
	\begin{subfigure}[b]{0.375\textwidth}
	\includegraphics[width=1\textwidth]{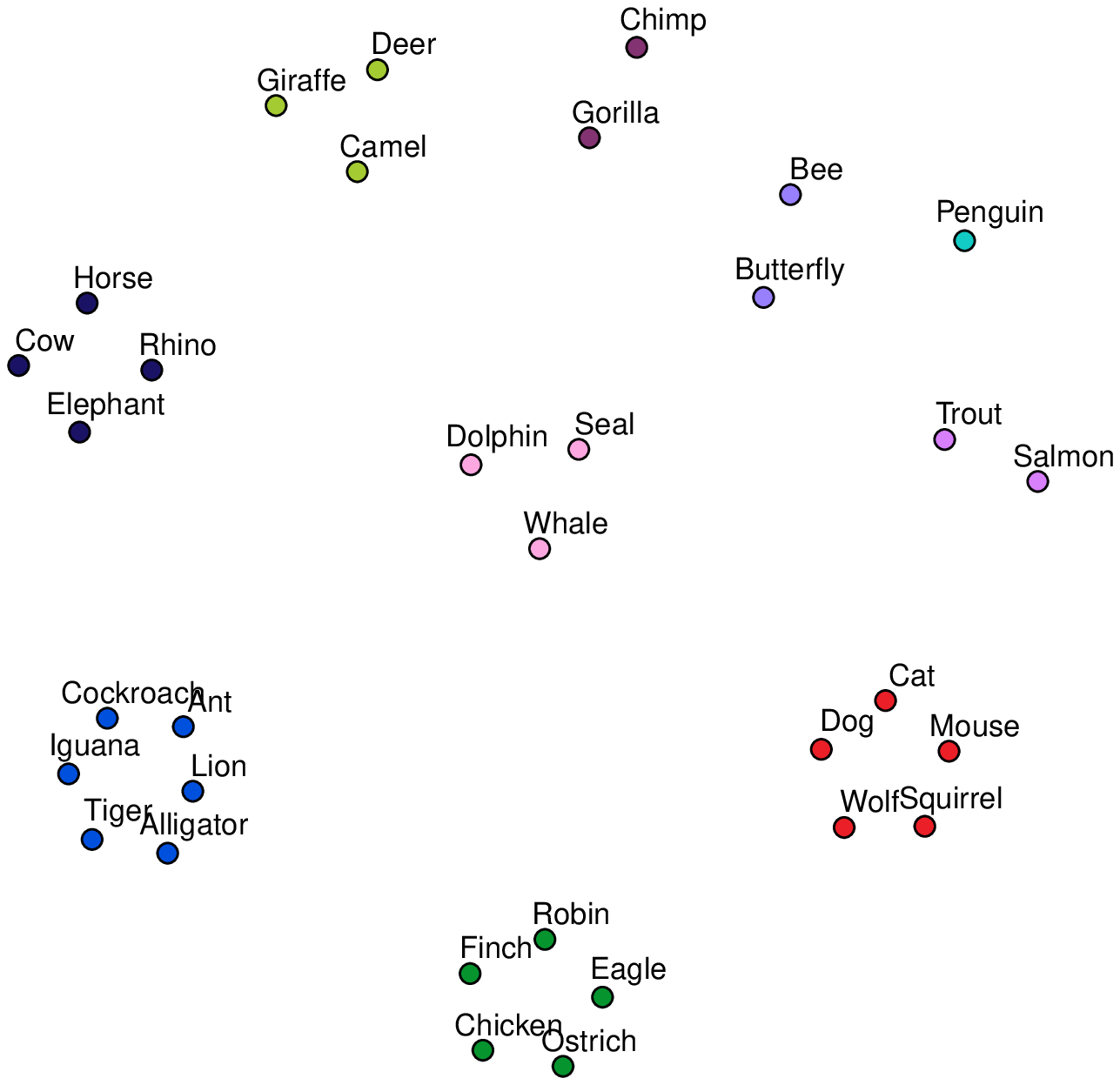}
	\caption{\textsf{Spectral} \citep{ng2002spectral}}
\end{subfigure}\\
\vspace{.5cm}
\begin{subfigure}[b]{0.375\textwidth}
	\includegraphics[width=1\textwidth]{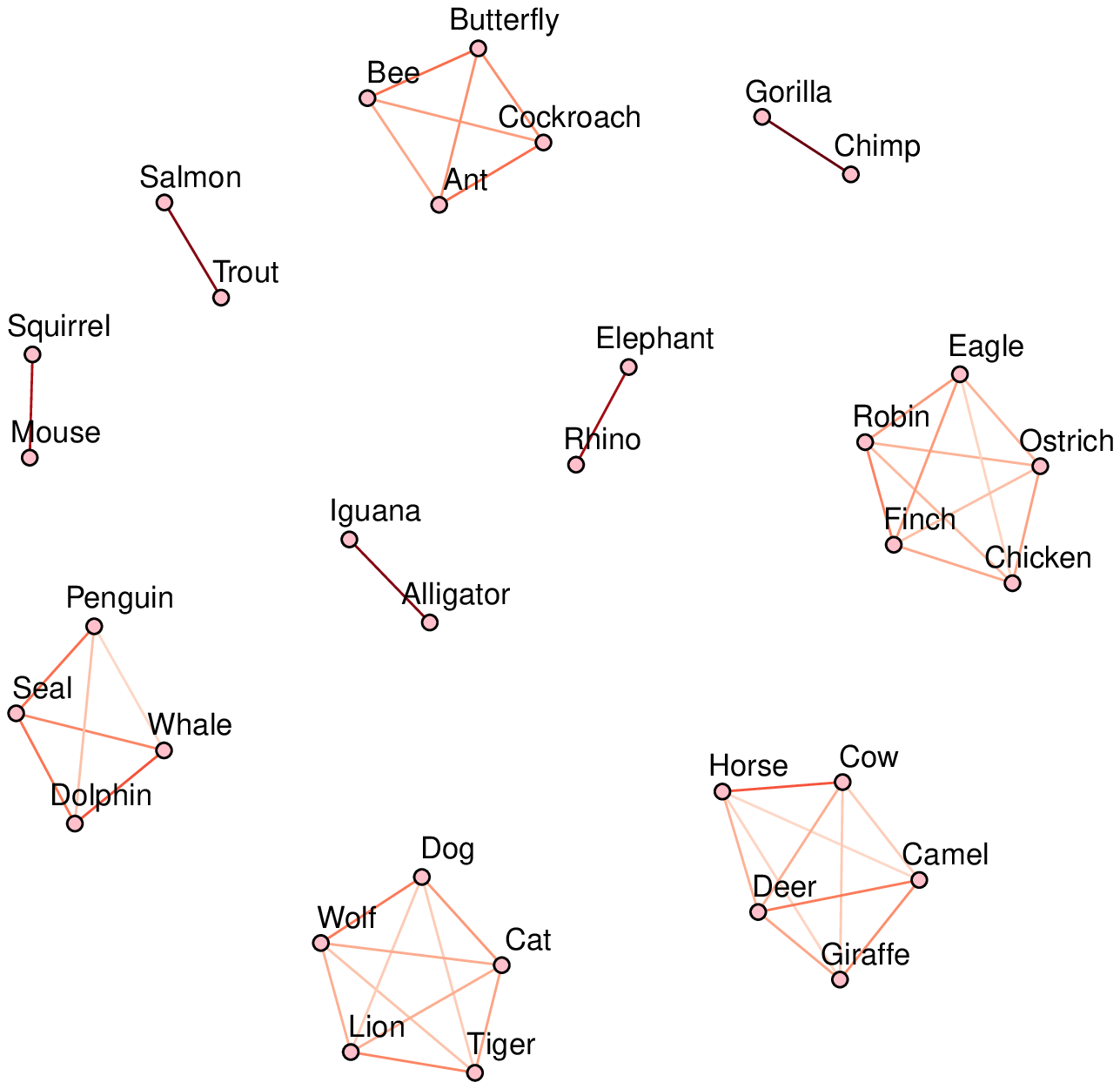}
	\caption{\textsf{CLR} \citep{nie2016constrained} }
\end{subfigure}
\hspace{1.6cm}
~
	\begin{subfigure}[b]{0.375\textwidth}
		\includegraphics[width=1\textwidth]{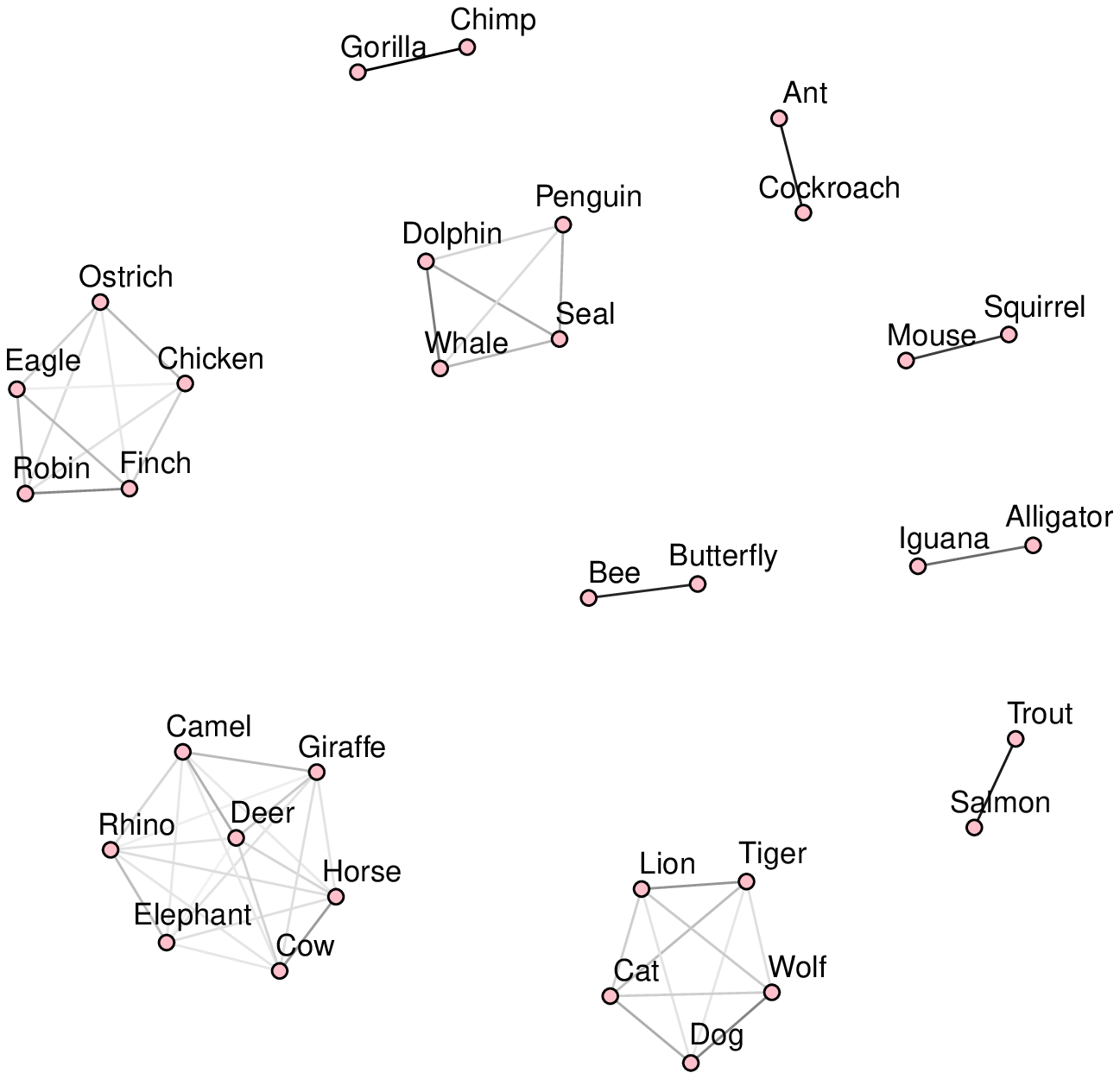}
		\caption{\textsf{SGL} with $k = 10$(proposed)}
	\end{subfigure}
\caption{All the methods obtain 10 components intending to group similar animals together. Clustering with \textsf{$k-$means} and \textsf{spectral} methods yield components with un-common(possibly wrong) groupings. For example, in (a) {\it seal, cow, horse} are grouped together: characteristics of seal does not seem to fit with the cow and horse, and in (b) {\it cockroach, lion, iguana, tiger, ant, alligator} are grouped together which is also in contrary to the expectation. On the other hand, it is observed that both \textsf{CLR} (c) and \textsf{SGL} (d) are able to obtain groupings of animal adhering to our general expectation of the animal behaviors. Although both the results vary slightly, the final results from both the methods are meaningful. For example, \textsf{CLR} groups all the insects ({\it bee, butterfly, cockroach, ant}) together in one group while \textsf{SGL} splits them into two groups, one with {\it ant, cockroach} and another with {\it bee, butterfly}. On the other hand, \textsf{SGL} groups the herbivore mammals ({\it horse, elephant, giraffe, deer, camel, rhino, cow}) together in one group, while \textsf{CLR} splits these animals into two groups, one containing {\it rhino, elephant} and another group containing the rest.}
\label{fig:animals-K}
\end{figure}

Figure~\ref{fig:animals-K} compares the clustering performance of the \textsf{SGL} method for $k=10$ clusters against the state-of-the-art clustering algorithms: (a) \textsf{$k-$means} clustering, (b) \textsf{spectral} clustering \footnote{ Code for \textsf{spectral} and $k-$\textsf{means} is available at https://cran.r-project.org/web/packages/kernlab \citep{karatzoglou2004kernlab}}, (c) \textsf{CLR}, and (d) \textsf{SGL} with $k=10$. It is remarked that all the algorithms except the \textsf{SGL} are designed only for clustering (grouping) task and are not capable of specifying further connectivity inside a group while \textsf{SGL} is capable of doing both the task of clustering and connectivity(edge weights) estimation jointly.

\subsubsection{Real data: Cancer Genome data set}
\begin{figure}[!htb]
	%	\centering
	\begin{subfigure}[b]{0.45\textwidth}
		\includegraphics[width=\textwidth]{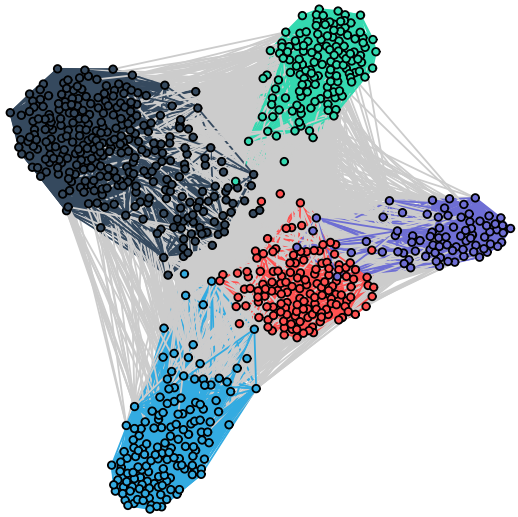}
		\caption{\textsf{GLasso}\citep{friedman2008sparse}}
		\label{fig:cancer-gene-glasso}
	\end{subfigure}
	\hspace{1cm}
	\vspace{1cm}
	~
	\begin{subfigure}[b]{0.45\textwidth}
		\includegraphics[width=\textwidth]{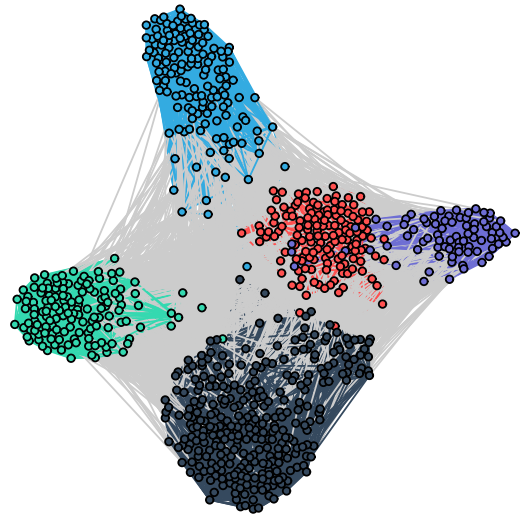}
		\caption{ \textsf{GGL}\citep{egilmez2017graph}}
		\label{fig:cancer-gene-ggl}
	\end{subfigure}
	\\
	\begin{subfigure}[b]{.45\textwidth}
		\includegraphics[width=\textwidth]{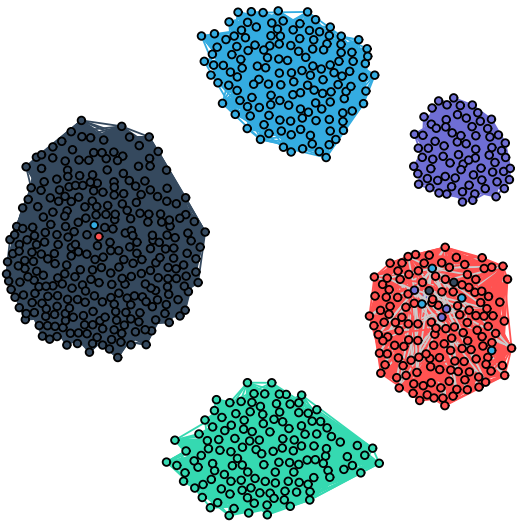}
		\caption{\textsf{CLR}\citep{nie2016constrained}}
	\end{subfigure}
	\hspace{1cm}
	~
	\begin{subfigure}[b]{.45\textwidth}
		\includegraphics[width=\textwidth]{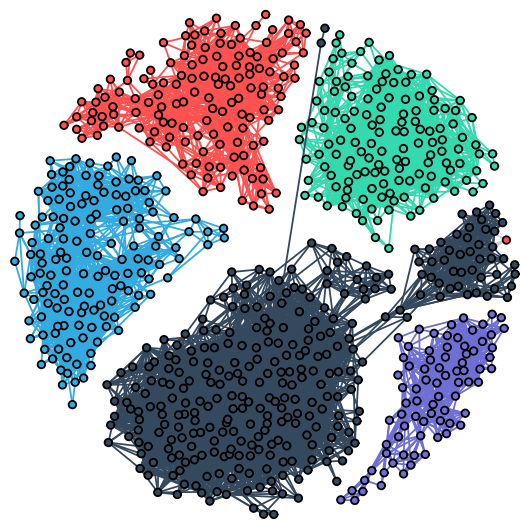}
		\caption{\textsf{SGL}(proposed)}
	\end{subfigure}
	\caption{Learning the clustered graph for the full 801 genetic samples with 20531 features of the \textsf{PANCAN} data set. Please see the plots in color version. Graph learned with (a) \textsf{GLasso}, (b) \textsf{GGL} method: grey colored edges indicate wrong connections, (c) \textsf{CLR} method: there are two misclassified points in the {\it black} group and 10 misclassified points in the {\it red} group, and (d) Graph learned with proposed \textsf{SGL} method. Herein, the label information is not taken into consideration. The result for \textsf{SGL} is consistent with the label information: samples belonging to each set are connected together and the components are fairly separated with only few wrongly inter component connections. Furthermore, the graph for the BRCA (black) data sample highlights an inner sub-grouping: suggesting for further biological investigation. This indicates that \textsf{SGL} has the potential to perform both the clustering and multiple graph learning simultaneously. }
	\label{fig:cancer-gene-sgl-full}
\end{figure}

For a big-data application, we consider the RNA-Seq Cancer Genome Atlas Research Network \citep{weinstein2013cancer} data set available at the UC-Irvine Machine Learning Database \citep{Dua:2017}. This data set consists of genetic features which map 5 types of cancer namely: breast carcinoma (BRCA), kidney renal clear-cell carcinoma (KIRC), lung adenocarcinoma (LUAD), colon adenocarcinoma (COAD), and prostate adenocarcinoma (PRAD). In Figure~\ref{fig:cancer-gene-sgl-full}, they are labeled with colors\footnote{Please see the plots in color version}, \emph{black, blue, red, violet}, and \emph{green}, respectively. The data set consists of 801 labeled samples, in which every sample has 20531 genetic features. An important goal with this data set is to classify and group the samples, according to their tumor type, on the basis of those genetic features. We apply the \textsf{SGL} for $k = 5, \beta = 5$ algorithm that exploits the spectral constraints to obtain $5$ component graph structure: we do not consider the label information.

We also compare the \textsf{SGL} performance against the existing state-of-the-art methods for graph learning algorithms namely \textsf{GLasso} with $\alpha=1.71$ and \textsf{GGL} with $\alpha=1.71$ algorithms, and also with the graph-based clustering algorithm \textsf{CLR} with $m=10$, where $m$ is the number of neighbors taken into the consideration for creating affinity matrix. We also conducted \textsf{CLR} experiments with different choices of $m=3,5,7$, and we observed similar performances\footnote{The authors in \citet{nie2016constrained} report that the \textsf{CLR} is insensitive with the choice of $m$} for all the values of $m$. See Figure~\ref{fig:cancer-gene-sgl-full} \textsf{GLasso} and \textsf{GGL} are not able to enforce component structure and learn a connected graph with many wrong undesirable edges. 
\textsf{SGL} method outperforms \textsf{CLR} for the clustering task as well, even though the later is a specialized clustering algorithm. 

The improved performance of the \textsf{SGL} can be attributed to two main reasons i) \textsf{SGL} is able to estimate the graph structure and weight simultaneously, which is essentially an optimal joint procedure, ii) \textsf{SGL} is able to capture the conditional dependencies (i.e., the precision matrix entries) among nodes which consider a global view of relationships, while the \textsf{CLR} encodes the connectivity via the direct pairwise distances. The conditional dependencies relationships are expected to give improved performance for clustering tasks \citep{JMLR:v18:17-019}. The performance with \textsf{SGL} shows a perfect clustering, which indicates that \textsf{SGL} may be used to perform simultaneous clustering and graph learning.

\subsubsection{Effect of the parameter $\beta$}
\begin{figure}[!htb]	
	\begin{subfigure}[b]{0.45\textwidth}
			\includegraphics[width=1\textwidth]{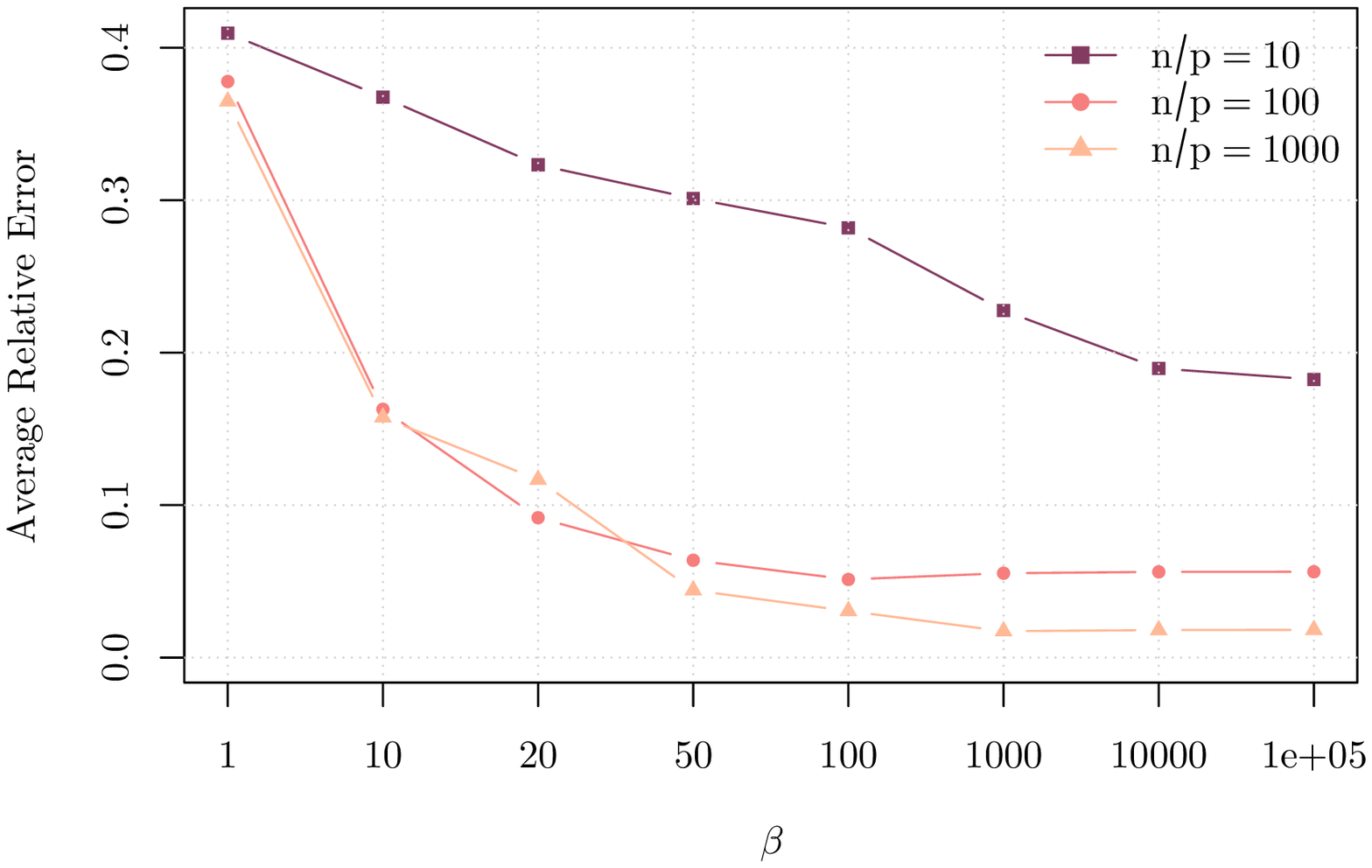}
		\end{subfigure}\qquad 
		\begin{subfigure}[b]{0.45\textwidth}
			\includegraphics[width=1\textwidth]{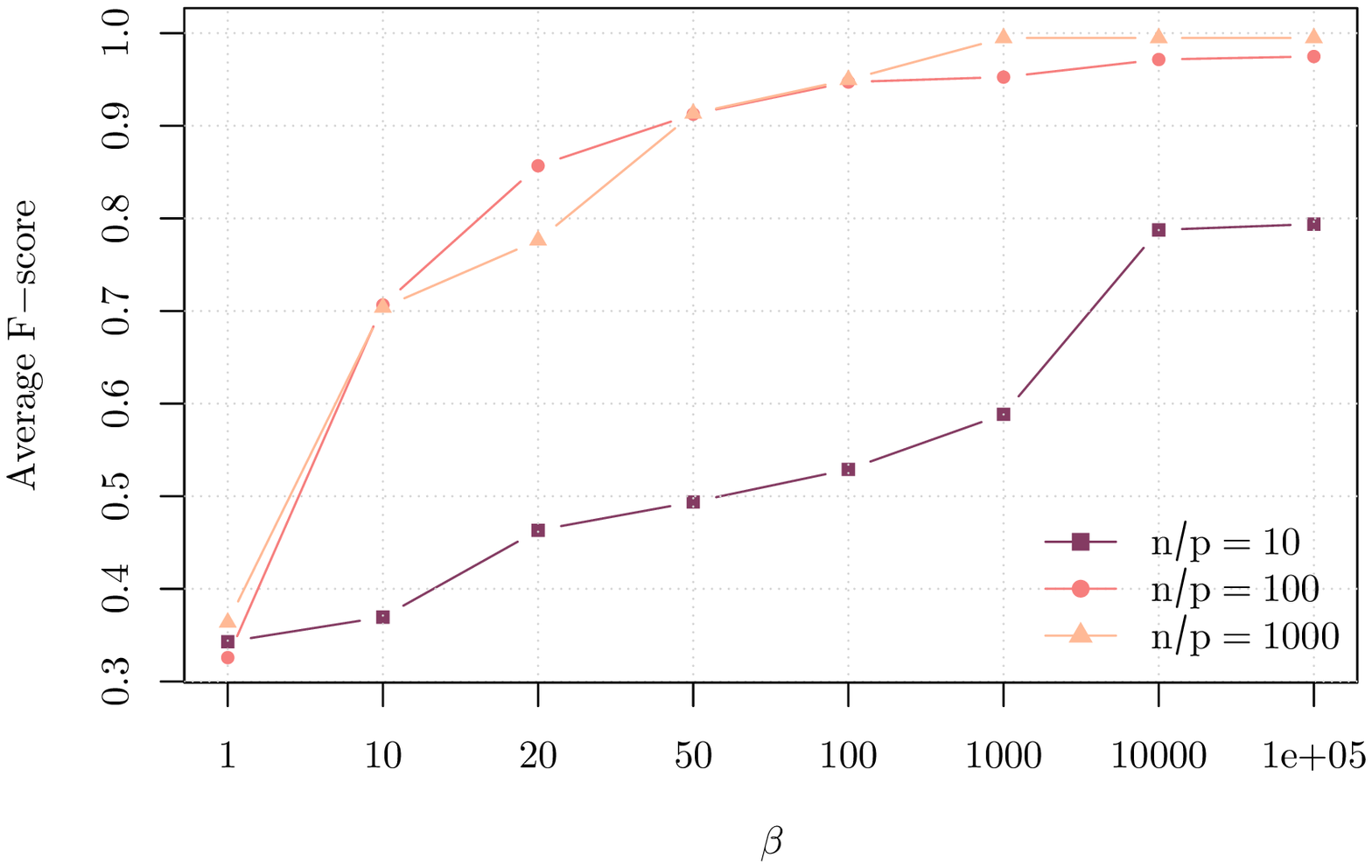}
		\end{subfigure}
\caption{Effect of the parameter $\beta$ on the \textsf{SGL} algorithm. We consider here estimating of a multi-component graph structure $\G_{\mathsf{mc}}(32,4,0.5)$ edge weights drawn randomly uniformly from [0,1]. It is observed from that a large $\beta$ enables the \textsf{SGL} to obtain a low \textsf{RE} and high \textsf{FS}.} 
\label{fig:beta_affect}
\end{figure}

In the current subsection, we study the effect of the parameter $\beta$ on the algorithm performance in terms of \textsf{RE} and \textsf{FS}. It is observed from Figure~\ref{fig:beta_affect} that a large $\beta$ enables the \textsf{SGL} to obtain a low \textsf{RE} and a high \textsf{FS}. For a large $\beta$, the formulation puts more weight on the relaxation term so as to make it closer to the spectral constraints. In addition, along with the increase of $\beta$ the \textsf{RE} and \textsf{FS} tend to be stable. But empirically it is observed that a huge $\beta$ slows down the convergence speed of the algorithm. Similar observations are also made regarding the parameter $\gamma$ for \textsf{SGA} and \textsf{SGLA} algorithms. 

\subsection{Performance evaluation for \textsf{SGA} Algorithm }\label{simulation-algo2}
\begin{figure}[!htb]
	\begin{subfigure}[b]{0.45\textwidth}
		\includegraphics[width=1\textwidth]{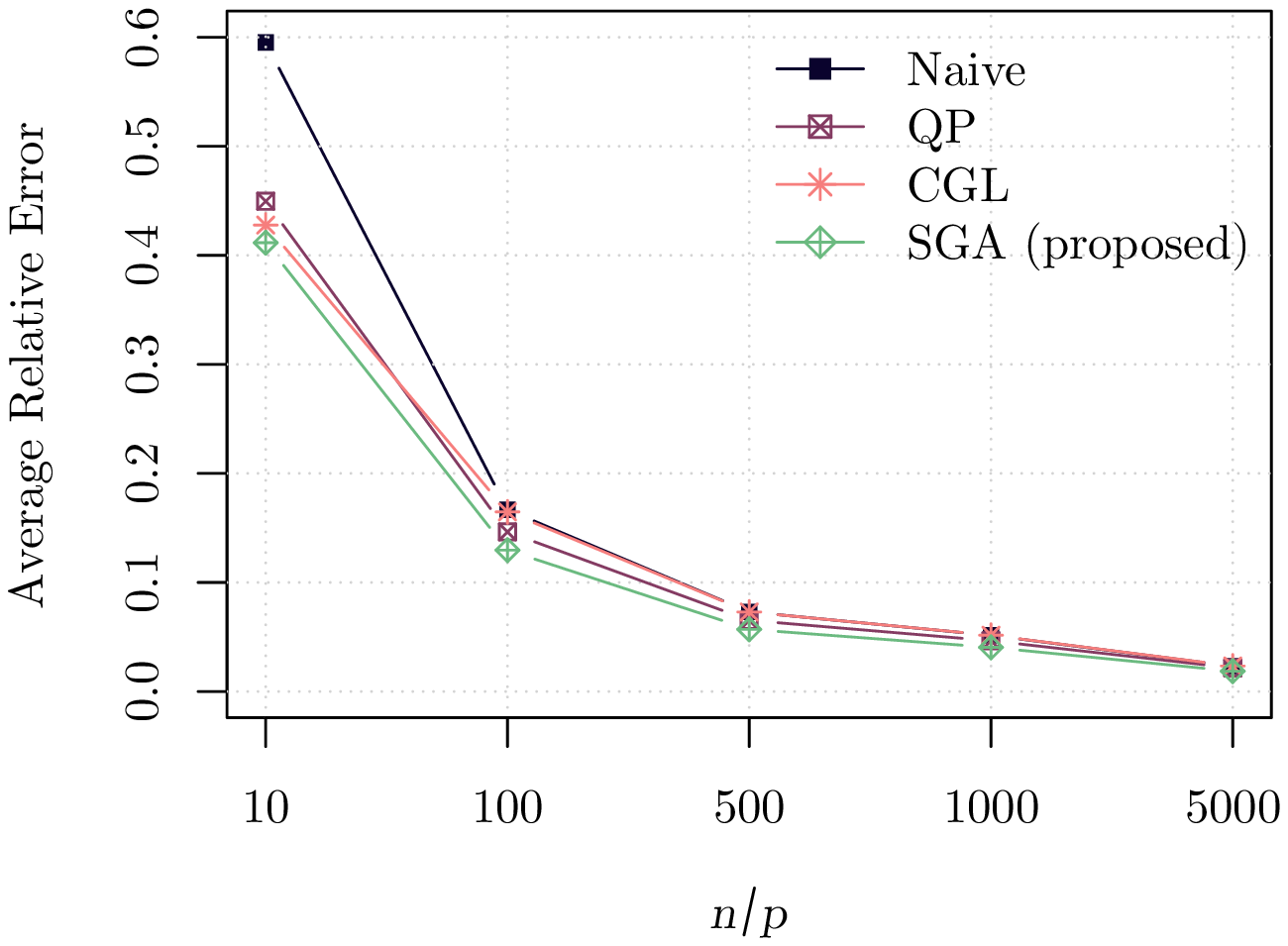}
		%		\caption{Average Relative Error vs $n/p$}
	\end{subfigure}\qquad \qquad
	~
	\begin{subfigure}[b]{0.45\textwidth}
		\includegraphics[width=1\textwidth]{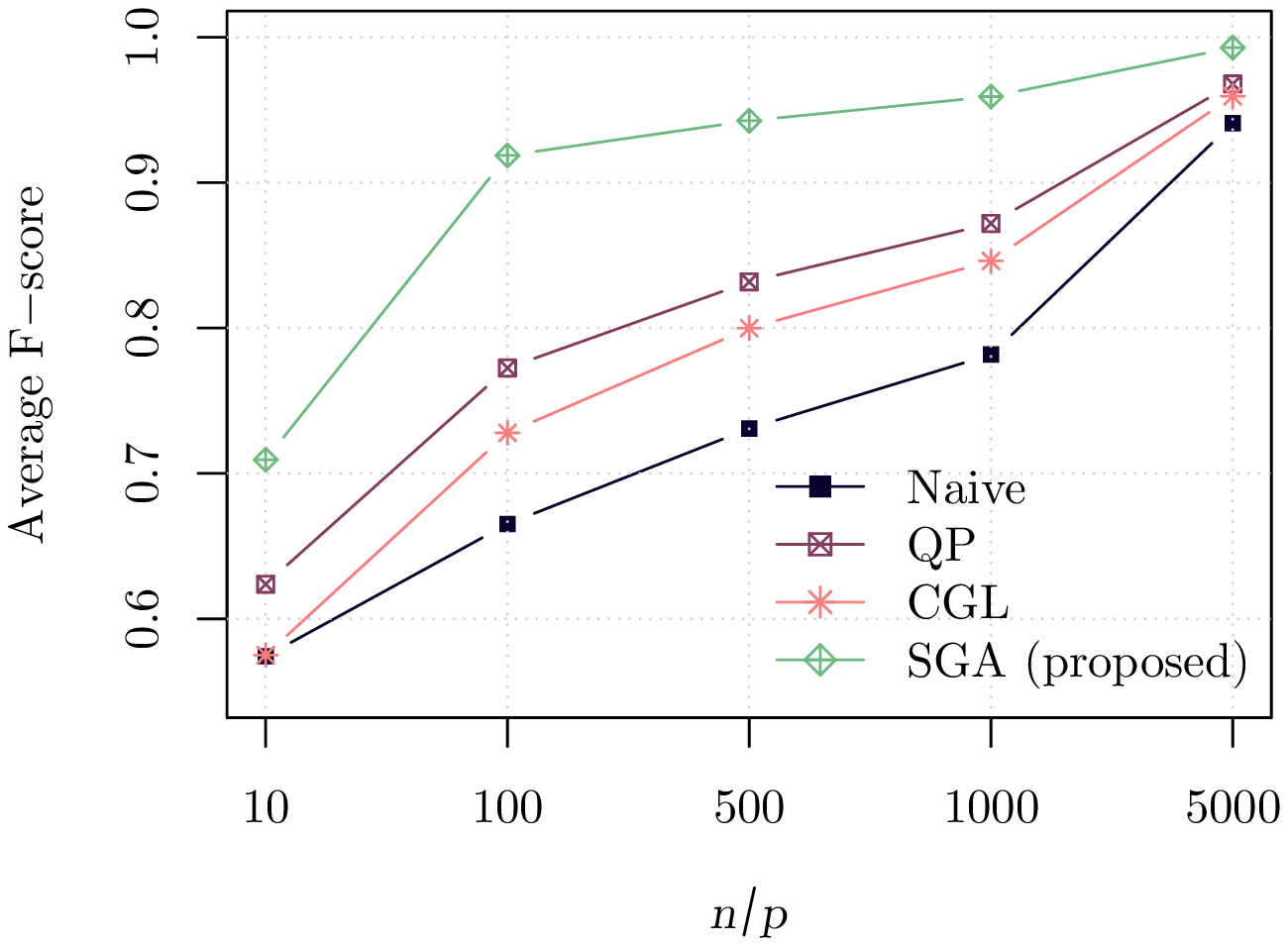}
		%		\caption{Average F-score vs $n/p$}
	\end{subfigure}
	%	\begin{subfigure}[b]{0.3\textwidth}
	%	\includegraphics[width=\textwidth]{examples/bipartite/connected/27march/accuracy_bipartite.eps}
	%%	\caption{Average F-score vs $n/p$}
	%\end{subfigure}
	\caption{Average performance results for learning Laplacian matrix of a bipartite graph structure $\mathcal{G}_{\mathsf{bi}}$. The proposed $ \textsf{SGA}$ method significantly outperforms the base line approaches, \textsf{Naive, QP} and \textsf{CGL}.}
	\label{connected-bipartite:fig:performance:n/p}
\end{figure}

A bipartite graph is denoted as $ \G_{\mathsf{bi}}(p_1,p_2, \wp)$, having two disjoint subsets with $p_1$ vertices in one subset and $p_2$ vertices in the another subset, with $\wp$ as the probability of having an edge between the nodes of two disjoint subsets. 
Figure~\ref{connected-bipartite:fig:performance:n/p} depicts the average performance of the algorithms for different sample size regimes for the following bipartite graph structure $ \G_{\mathsf{bi}}(p_1=40,p_2=24, \wp=0.6)$ with edge weights are randomly uniformly selected from $[1,3]$. We set $\gamma=10^5$. Here we consider a connected bipartite graph, thus \textsf{SGA} is also compared against \textsf{CGL}.

\subsubsection{Bipartite graph leaning: noisy setting }
\begin{figure}[!htb]
	\centering
	\begin{subfigure}[b]{0.3\textwidth}
		\includegraphics[width=\textwidth]{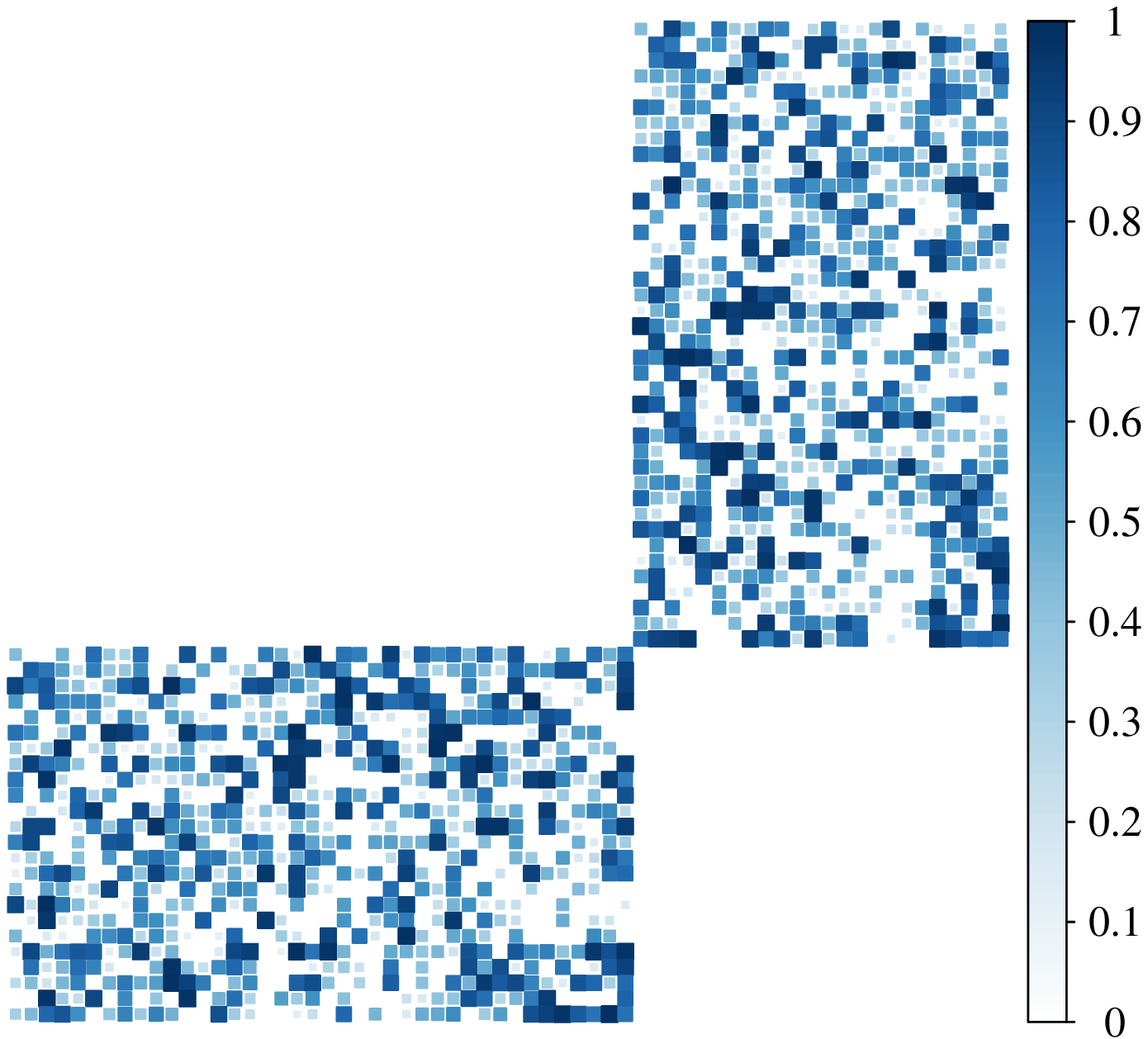}
	\caption{ $\Theta_{\textsf{true}}$ }
	\end{subfigure}
	~ %add desired spacing between images, e. g. ~, \quad, \qquad, \hfill etc.
	%(or a blank line to force the subfigure onto a new line)
	\begin{subfigure}[b]{0.3\textwidth}
		\includegraphics[width=\textwidth]{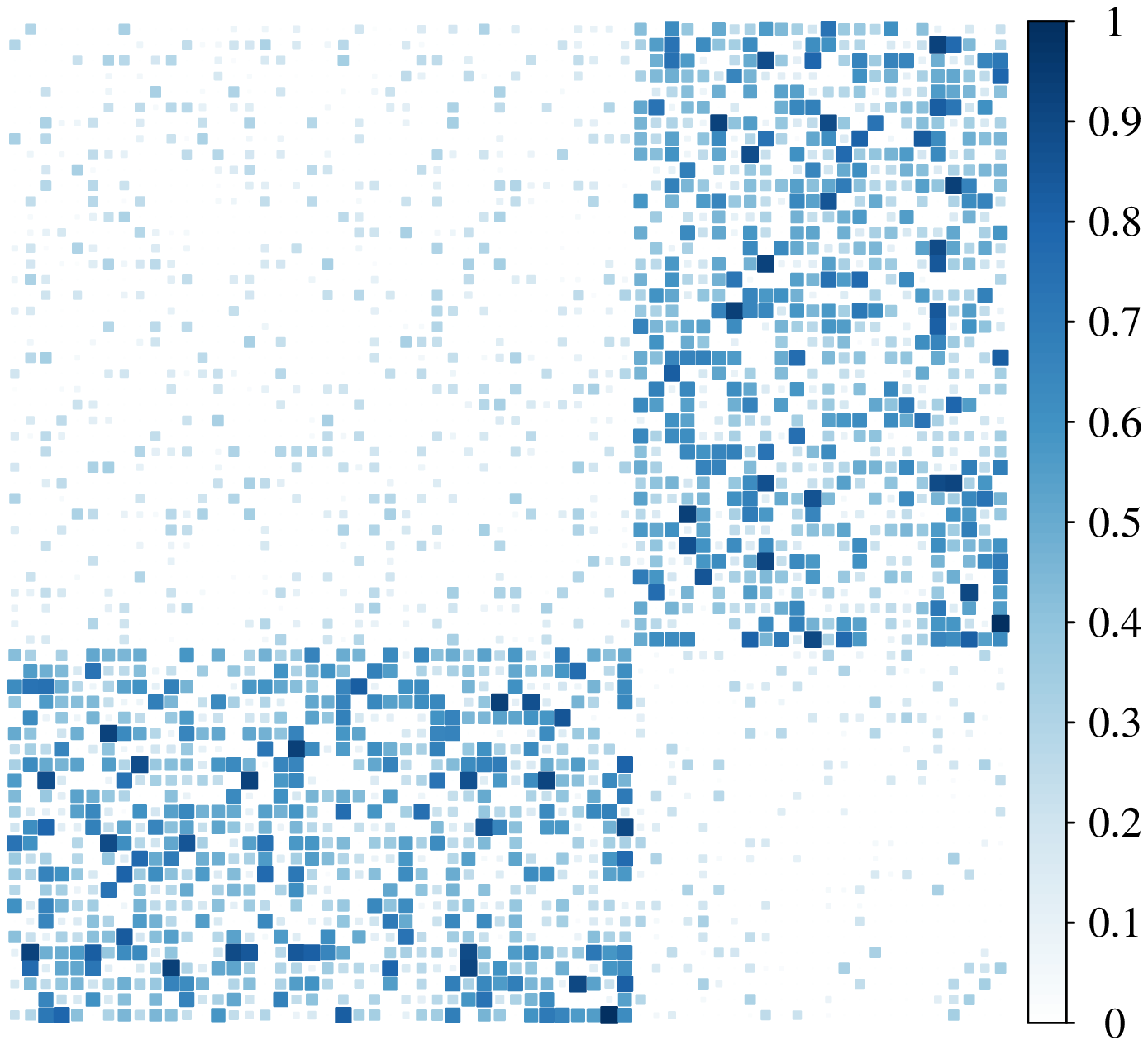}
	\caption{ $\Theta_{\textsf{noisy}}$}
	\end{subfigure}
	~ %add desired spacing between images, e. g. ~, \quad, \qquad, \hfill etc.
	%(or a blank line to force the subfigure onto a new line)
	\begin{subfigure}[b]{0.3\textwidth}
		\includegraphics[width=\textwidth]{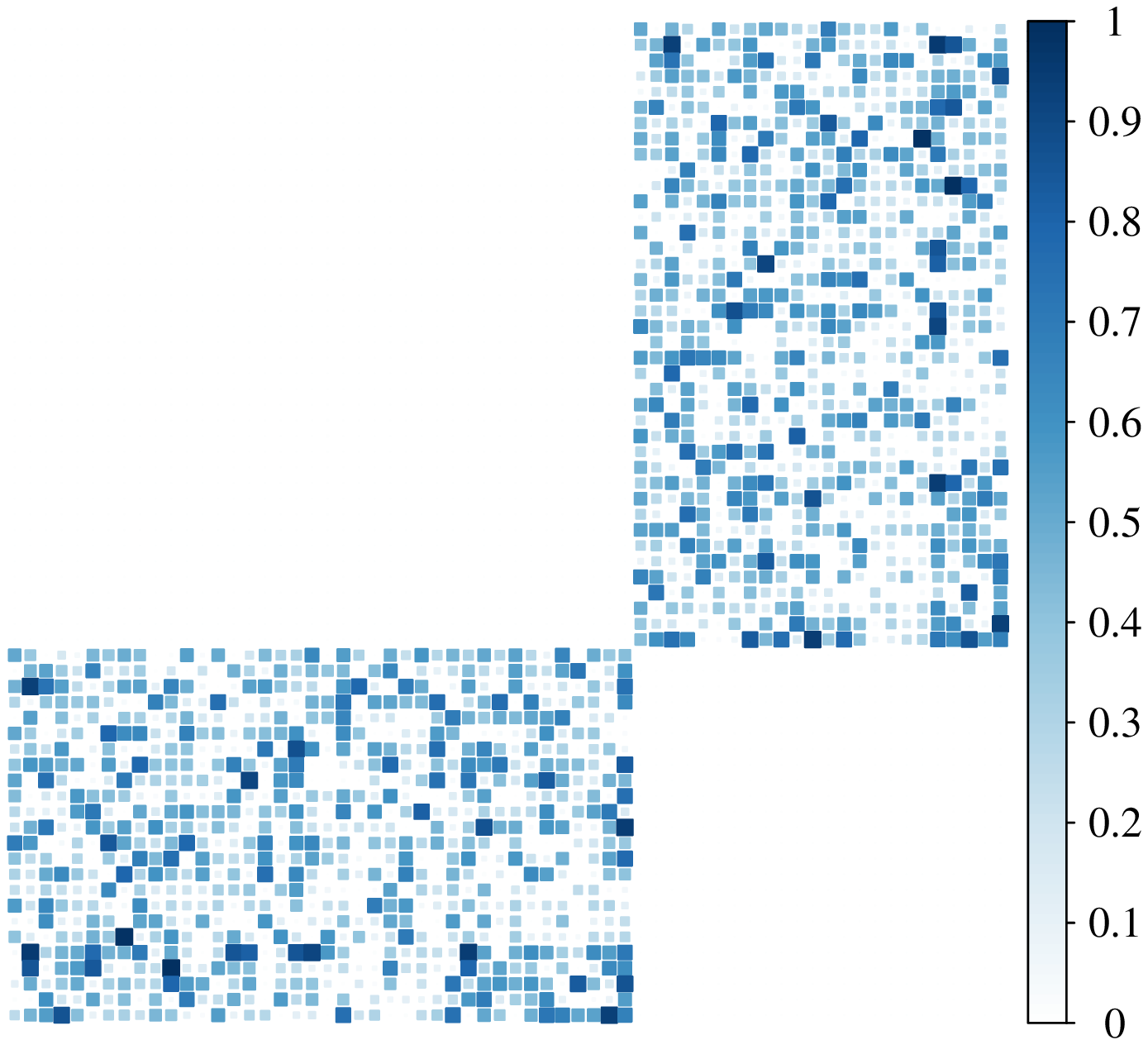}
	\caption{ $\Theta_{\textsf{learned}}$}
	\end{subfigure}
	\caption{An instance of bipartite graph learning with SGA algorithm with data generated from noisy graph Laplacian, and fixing $n/p=500$ and $\gamma=10^5$. (a) the ground truth Laplacian matrix ($\bTheta_{\mathsf{true}}$),
	(b) $\bTheta_{\mathsf{noisy}}$ after being corrupted by noise, (c) the learned graph Laplacian with a performance of $(\mathsf{RE}=0.219, \mathsf{FS} = 0.872)$.}
\label{bipartite-1}	
\end{figure}

We consider here learning of a bipartite graph structure under the noisy setting \eqref{noisy-block}, i.e., the samples used for calculating the SCM is obtained from noisy precision matrix, for which, the ground truth precision matrix corresponds to a bipartite graph. At first we generate $\G_{\textsf{bi}}(40,24,0.70)$ and the edge weights are drawn from $[0.1,1]$. Then we add random noise to all the possible connections, by adding edge weights following the ER graph $\mathcal{G}_{\mathsf{ER}}(64,0.35)$ graph model. Figure~\ref{bipartite-1} illustrates an instance of \textsf{SGA} algorithm performance for learning bipartite graph structure from noisy sample data.

\subsection{Performance evaluation for \textsf{SGLA} Algorithm }\label{simulation-algo3}
Herein, we consider learning of a multi-component bipartite graph structure. This structure is widely used in a lot of applications including medicine and biology \citep[see][]{nie2017learning,pavlopoulos2018bipartite}, which makes it appealing from both the practical as well as theoretical and algorithmic perspective. To learn multi-component bipartite graph structure from the SCM of data $S$, we plugin the Laplacian spectral properties ($\S_{\lambda}$ as in \eqref{Eig_set_K-connected}) corresponding to the multi-component structure along with the adjacency spectral constraints corresponding to the bipartite structure (i.e., $\S_{ \psi}$ as in \eqref{bipartite-eig}) in Algorithm \ref{algo-3} [cf. step 8, 9]. 
\begin{figure}[!htb]
	\begin{subfigure}[b]{0.45\textwidth}
		\includegraphics[width=1\textwidth]{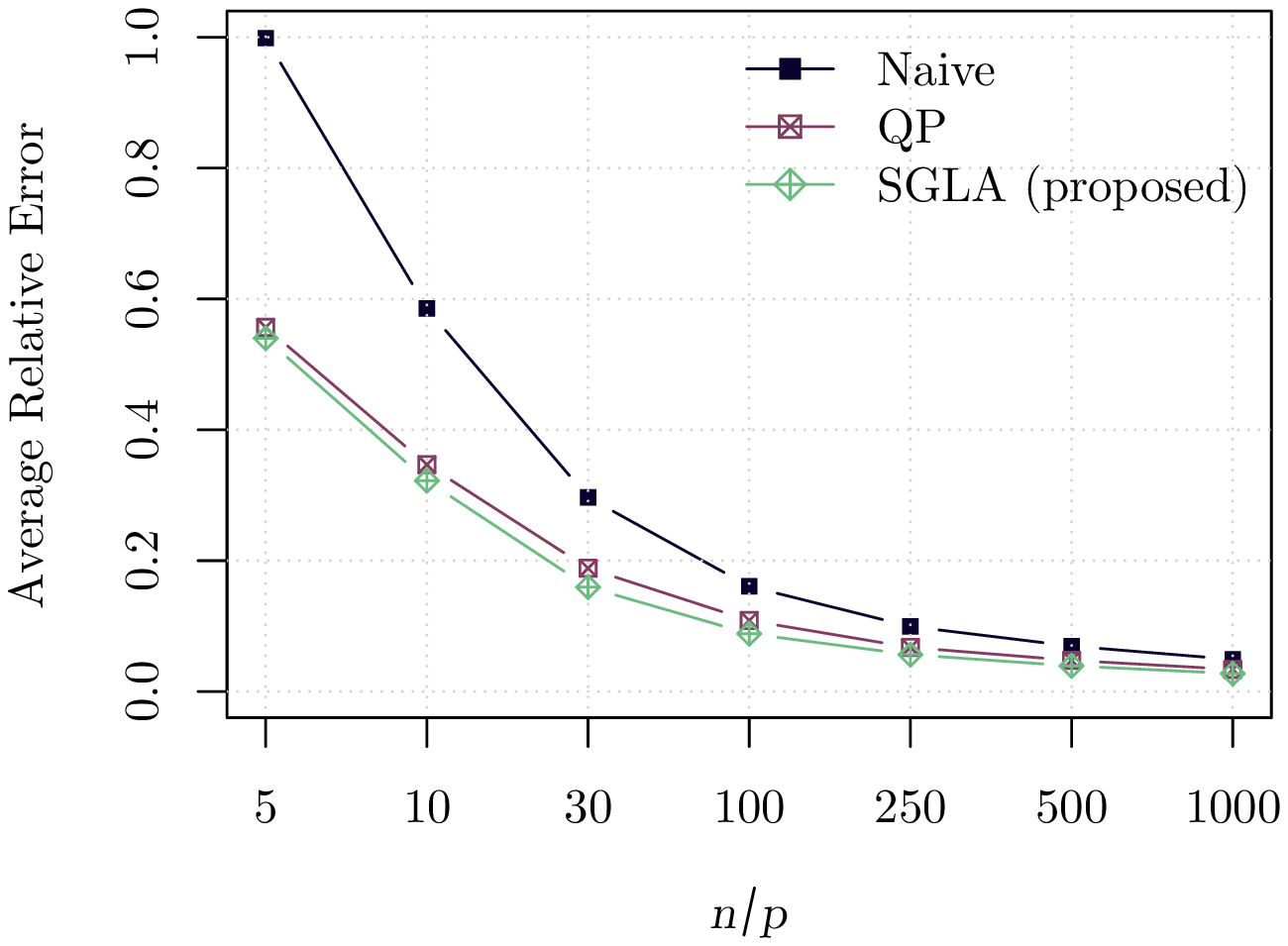}
		%		\caption{Average Relative Error vs $n/p$}
	\end{subfigure}\hspace{.5cm}
	\begin{subfigure}[b]{0.45\textwidth}
		\includegraphics[width=1\textwidth]{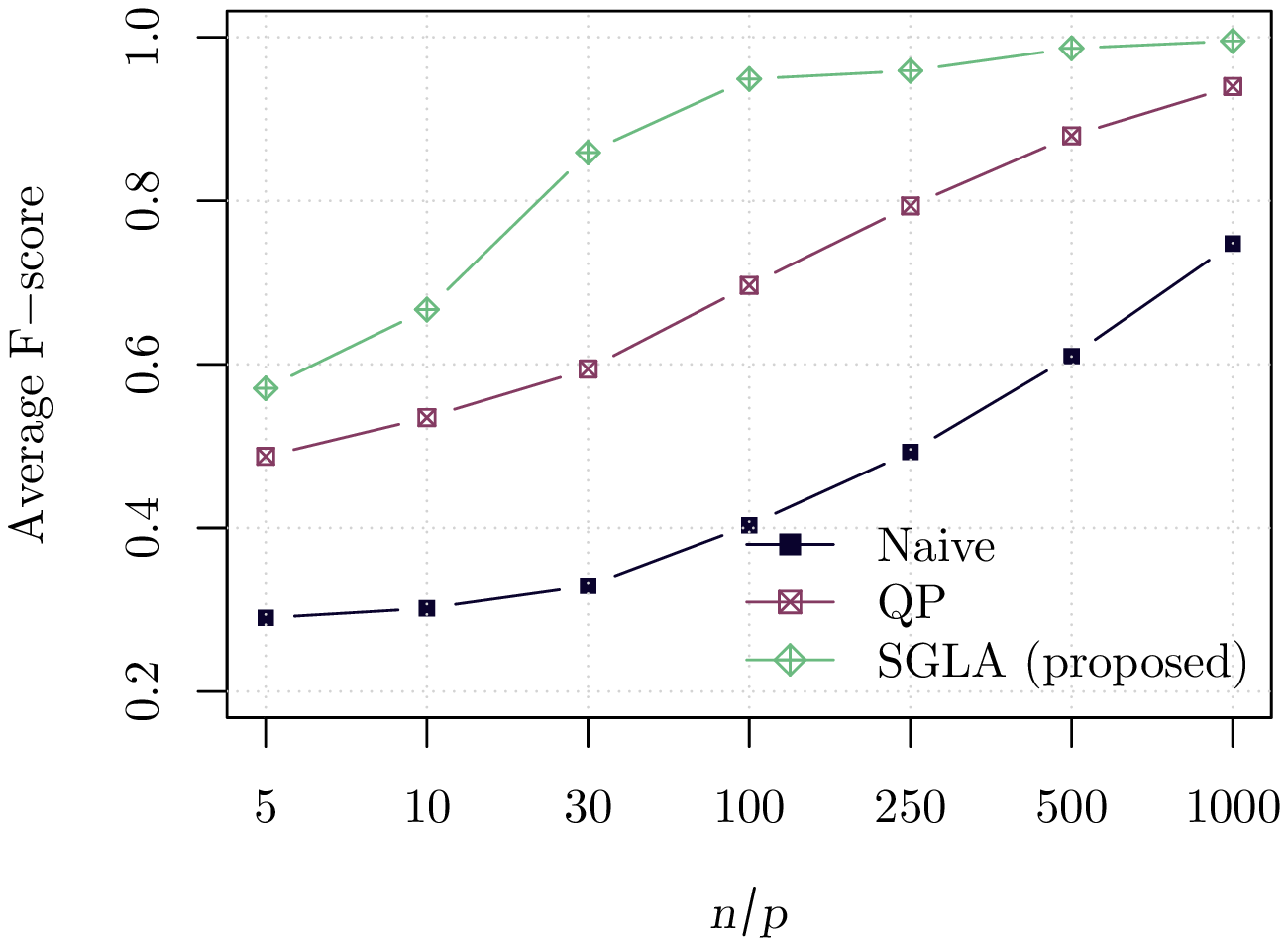}
		%	\caption{Average F-score vs $n/p$}
	\end{subfigure}
	\caption{Average performance results for learning Laplacian matrix of a $\mathcal{G}_{\mathsf{bi}}$. The proposed \textsf{SGLA} method significantly outperforms the base line approaches, \textsf{QP} and \textsf{Naive}.}
	\label{block-bipartite:fig:performance:n/p}
\end{figure}
Figure~\ref{block-bipartite:fig:performance:n/p} depicts the average performance of the algorithms for different sample size regimes for a block- bipartite graph structure three components $ k=3$ with unequal number of nodes, where each component represents a bipartite graph structure, denoted by $\G^1_{\textsf{bi}}(20,8,0.5)$, $\G^2_{\textsf{bi}}(12,8,0.6)^2$, and $\G^3_{\textsf{bi}}(8,8,0.7)$, bipartite edge weights are randomly uniformly drawn from $[0.1,3]$. We fix $\beta=10^3$ and $\gamma=10^3$. Here we consider a multi-component bipartite graph, therefore we can only compare against, \textsf{QP} and \textsf{Naive}. In terms of \textsf{RE}, \textsf{QP} performance is comparable with \textsf{SGLA} but in terms of \textsf{FS}, \textsf{SGLA} significantly outperforms the baseline approaches.

\subsubsection{Multi-component bipartite graph: noisy setting}
\begin{figure}[!htb]
	\centering
	\begin{subfigure}[b]{0.3\textwidth}
		\includegraphics[width=\textwidth]{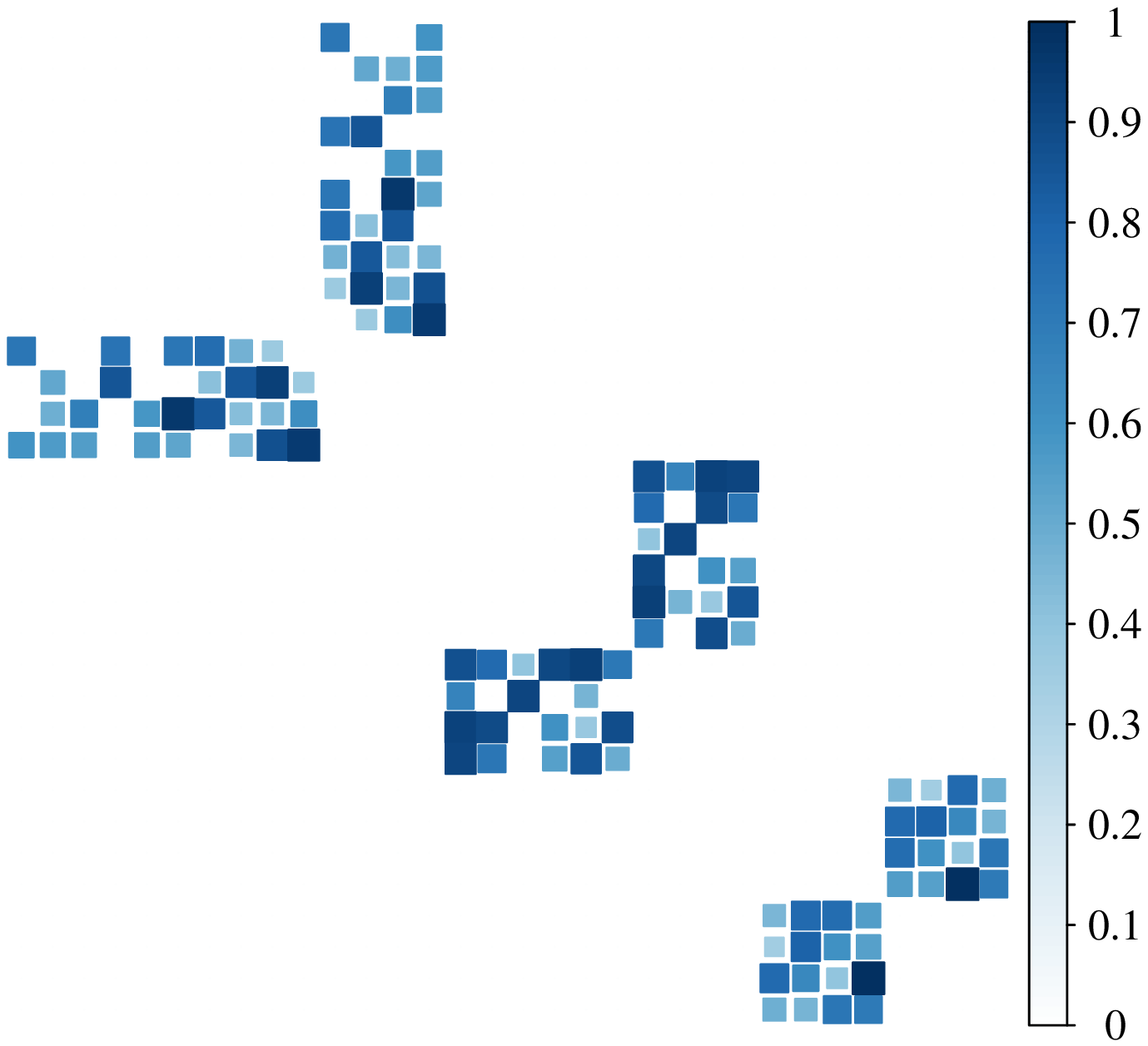}
	\caption{ $\Theta_{\textsf{true}}$}
	\end{subfigure}\quad
	~ %add desired spacing between images, e. g. ~, \quad, \qquad, \hfill etc.
	%(or a blank line to force the subfigure onto a new line)
	\begin{subfigure}[b]{0.3\textwidth}
		\includegraphics[width=\textwidth]{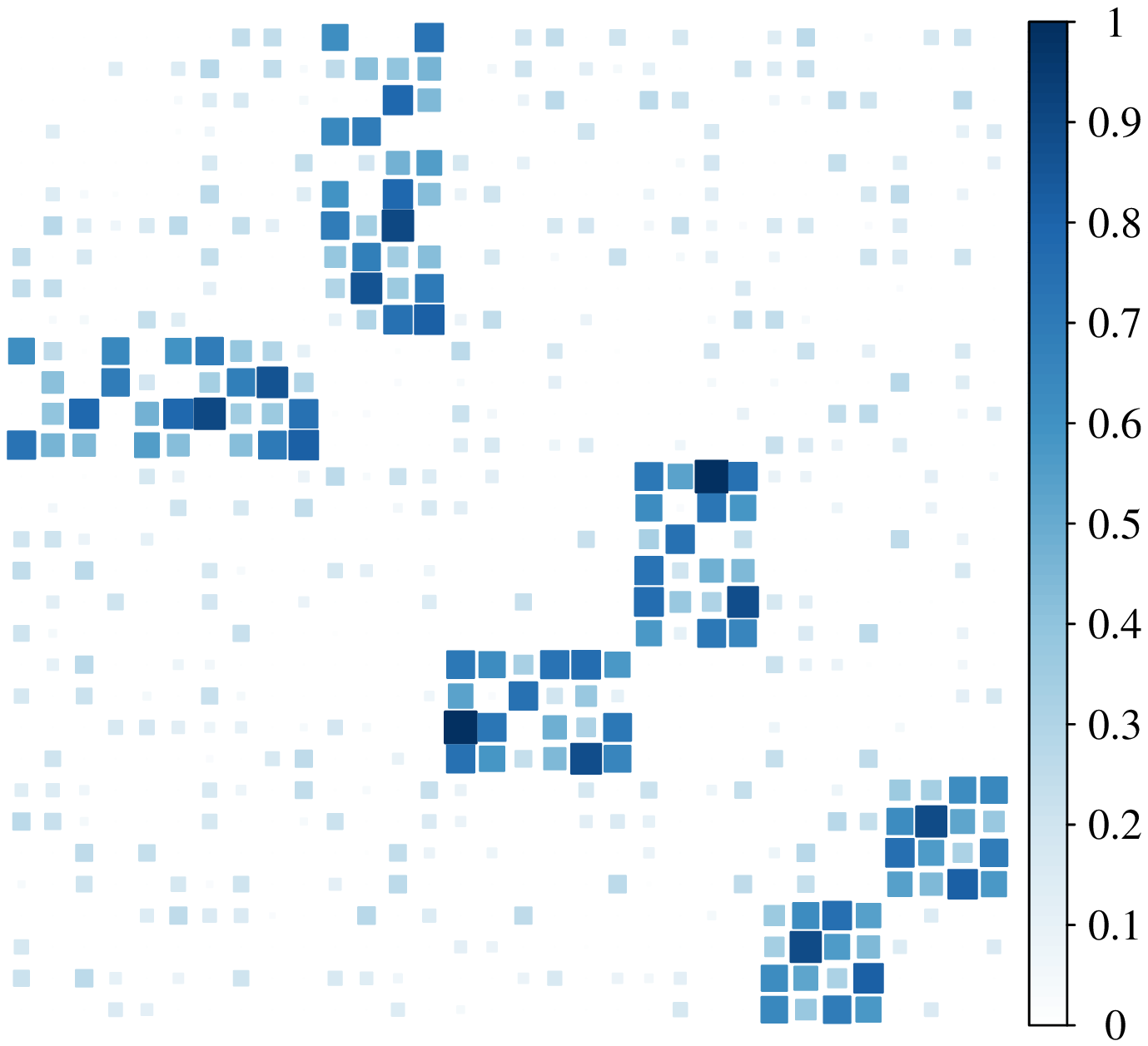}
	\caption{ $\Theta_{\textsf{noisy}}$ }
	\end{subfigure}\quad 
	~ %add desired spacing between images, e. g. ~, \quad, \qquad, \hfill etc.
	%(or a blank line to force the subfigure onto a new line)
	\begin{subfigure}[b]{0.3\textwidth}
		\includegraphics[width=\textwidth]{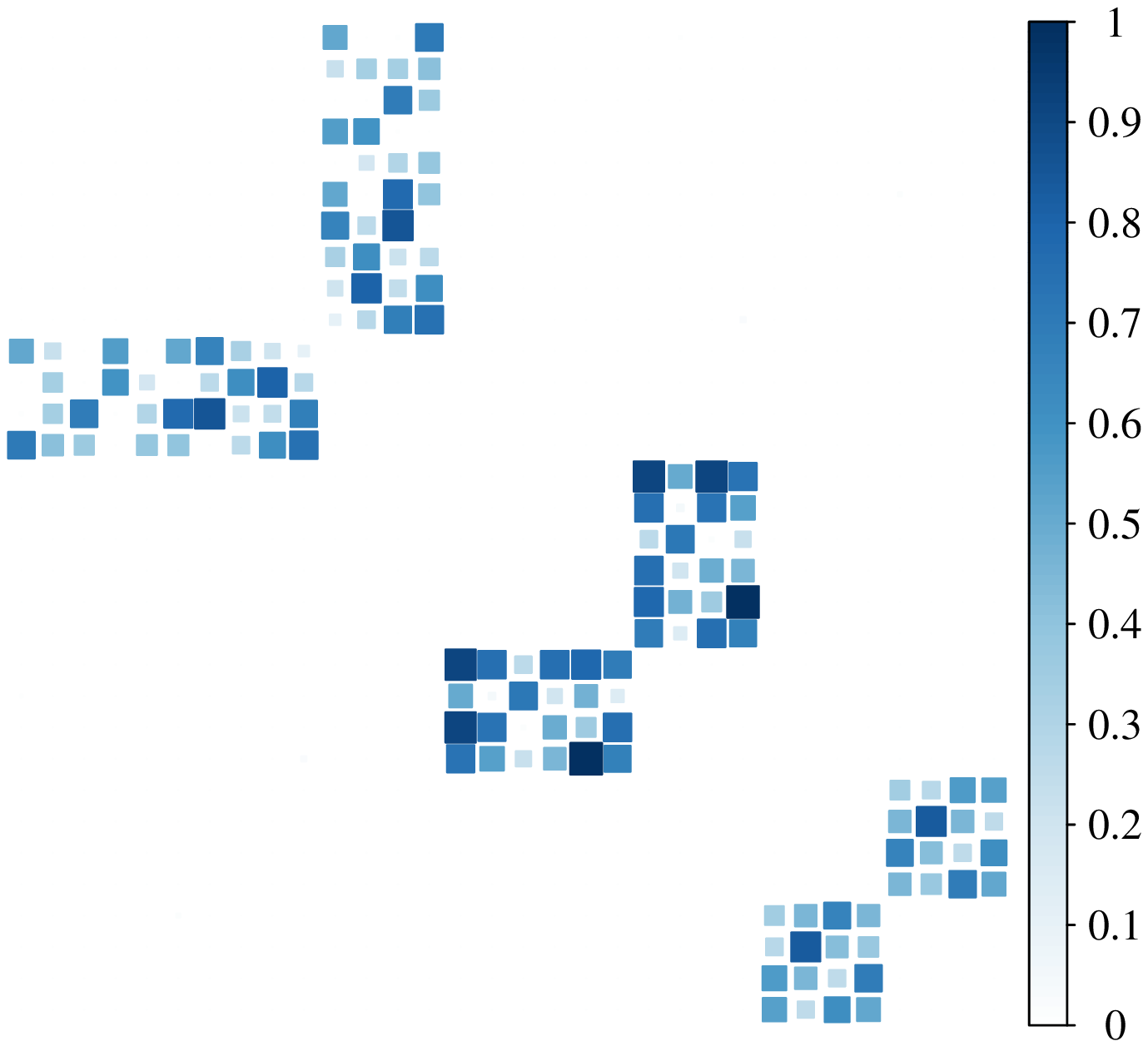}
	\caption{ $\Theta_{\textsf{learned}}$}
	\end{subfigure}
\caption{An example of learning a 3-component bipartite graph structure with \textsf{SGLA} algorithm, the samples used for calculating the SCM is obtained from noisy Laplacian precision matrix. Heat maps of the graph matrices: (a) $\bTheta_{\mathsf{true}}$ the ground truth Laplacian matrix, (b) $\bTheta_{\mathsf{noisy}}$ after being corrupted by noise, (c) $\Theta_{\textsf{learned}}$ the learned graph Laplacian with a performance of $(\mathsf{RE}=0.225, \mathsf{FS} = 0.947)$, which means a perfect structure recovery even in a noisy model that heavily suppresses the ground truth weights. }\label{fig:3-comp-bip-graph}
\end{figure}

We consider here learning of a block-bipartite graph structure under the noisy setting \eqref{noisy-block}, i.e., the samples used for calculating the SCM is obtained from noisy precision matrix, for which, the ground truth precision matrix corresponds to a multi-component bipartite graph. We first generate a graph of $p=32$ nodes and three components $ k=3$ with unequal number of nodes, where each component represents a bipartite graph structure, denoted by $\G^1_{\textsf{bi}}(10,4,0.7)$, $\G^2_{\textsf{bi}}(6,4,0.8)^2$, and $\G^3_{\textsf{bi}}(4,4,0.9)$, bipartite edge weights are drawn from $[1,3]$. Then we add random noise to all the possible edges between any two vertices by adding edge weights following the ER graph $\mathcal{G}_{\mathsf{ER}}(32, .35)$ with edges drawn from $[0,1]$, and we fix $n/p=250$, $\gamma=10^5$ and $\beta=10^5$. 
Figure~\ref{fig:3-comp-bip-graph} depicts one instance of the performance of \textsf{SGLA} for noisy bipartite graph structure.

\section{Discussion and Conclusion}\label{conclusion}
In this paper, for learning graphs with specific structures, we have considered the spectral constraints on the eigenvalues of graph matrices under the GGM setting. We have developed a unified optimization framework which is very general and putsforth plausible directions for future works. 

\subsection{Discussion} 
The extensions of this paper include considering more specific prior information on eigenvalues constraints, obtaining structured graph transform for graph signal processing applications and, extending the framework by considering other statistical models. 
\subsubsection{Other spectral prior information}
In addition to the spectral properties considered in this paper, there are numerous interesting results in the literature relating other properties to some specific graph structures, we refer readers to the comprehensive exposition in \citep[see for Laplacian eigenvalues,][]{das2005sharp,yu2014lower,farber2011upper,berman2011lower, chung1997spectral,spielman2011spectral} see \citep{chung1997spectral,mohar1997some,yang2003bounds,van2010graph,schulte1996constructing,lin2014some}. The proposed framework has the potential to be extended for undertaking those prior spectral information.

\subsubsection{Structured statistical models}
 Although, the focus of the current paper is on the unification of spectral graph theory with the GGM framework, but the proposed spectral constraints and developed algorithms are very general and can be integrated with other statistical models such as structured Ising model \citep{ravikumar2010high}, structured Gaussian covariance graphical models \citep{drton2008graphical}, structured Gaussian graphical models with latent variables \citep{chandrasekaran2010latent}, least-square formulation for graph learning \citep{nie2016constrained}, structured linear regression, and vector auto-regression models \citep{basu2015network}.

\subsubsection{Structured graph signal processing }
One of the motivations of the present work comes from Graph signal processing (GSP), which provides a promising tool for the representation, processing and analysis of complex networked-data (e.g., social, energy, finance, biology and, etc.), where the data are defined on the vertices of a graph \citep{6494675,ortega2018graph}. The eigenvector of graph matrices are used to define graph Fourier transform also known as dictionaries, used in a variety applications including, graph-based filtering, graph-based transforms, and sampling for graph signals. More recently it is realized that the graph eigenvectors with additional properties (e.g., sparsity) could be instrumental in doing some complex tasks. For example, sparse eigenvectors of the graph matrices are important to investigate the uncertainty principle over graphs \citep{teke2017uncertainty}. Few other examples include sparsifying eigenvectors \citep{sardellitti2018graph} and robust eigenvectors\citep{maretic2017graph}. Within the proposed formulation, one can easily enforce desired properties on the eigenvectors by pairing the optimization step of \eqref{CGL_A}, \eqref{CGL_L_A} and \eqref{CGL_L:0} with specific constraints on the eigenvectors (e.g., through regularization). Joint learning structured graphs with specific properties on the eigenvectors have significant potential for GSP applications, which have not been investigated yet, and constitute a promising research direction.

%The eigenvector of graph matrices are used to define graph Fourier transform also known as dictionaries, used in a variety applications including, graph-based filtering, graph-based transforms, and sampling for graph signals.
%
%The eigenvector of graph matrices (adjacency and Laplacian), respectively, play crucial role in the GSP applications.
%
%
%
%These are essentially used to define graph Fourier transform also known as dictionaries. These graph transforms are being used in a variety applications including, graph-based filtering, graph-based transforms, and sampling for graph.

% which is an active area of research having immense potential in numerous applications of social, energy, finance, biology and, etc.

\subsection{Conclusion}
Summarizing, in this paper, we have shown how to convert the combinatorial constraints of structured graph learning into analytical constraints of the graph matrix eigenvalues. We have developed three algorithms \textsf{SGL}, \textsf{SGA}, and \textsf{SGLA} that can learn structured graphs from a large class of graph families. The algorithms are capable of learning the graph structure and their weights simultaneously by utilizing the spectral constraints of graph matrices directly into the Gaussian graphical modeling framework. The algorithms enjoy comprehensive theoretical convergence properties along with low computational complexity. Extensive numerical experiments with both synthetic and real data sets demonstrate the effectiveness of the proposed methods. The paper also puts forth several extensions worthy of possible future research direction.

%\newpage

\section{Appendix}

\subsection{Proof of Lemma \ref{lem-lap-norm}}\label{proof:lem-lap-norm}

\begin{proof}
We define an index set $\Omega_t$:
\begin{align}\label{omega}
\Omega_t := \left \{ l \ |[\L \x]_{tt} = \sum_{l \in \Omega_t} x_l \right \}, \quad t \in [1, p].
\end{align} 

For any $\x \in \Rn ^{\frac{p(p-1)}{2}}$, we have
\begin{align}
\norm{\L \x}_F^2 &= 2\sum_{k=1}^{\frac{p(p-1)}{2}}x_k^2 + \sum_{i=1}^{p}([\L \x]_{ii})^2 \\
&= 4\sum_{k=1}^{\frac{p(p-1)}{2}}x_k^2 + \sum_{t=1}^n \sum_{i, j \in \Omega_t, \ i \neq j} x_i x_j \\
& \leq 4\sum_{k=1}^{\frac{p(p-1)}{2}}x_k^2 + \frac{1}{2}\sum_{t=1}^n \sum_{i, j \in \Omega_t, \ i \neq j} x_i^2 + x_j^2 \\
& = (4 + 2(|\Omega_t| -1))\sum_{k=1}^{\frac{p(p-1)}{2}}x_k^2 \\
& = 2p\norm{\x}^2,
\end{align}
where the second equality is due to the fact that each $x_k$ only appears twice on the diagonal; the first inequality achieves equality when each $x_k$ is equal; the last equality follows the fact that $|\Omega_t |=p-1$.

Therefore, by the definition of operator norm, we can obtain
\begin{align}
\norm{\L}_2 = \sup_{\norm{\x}=1}\norm{\L \x}_F =\sqrt{2p}, 
\end{align}
concluding the proof.

\end{proof}

\subsection{Proof for Lemma \ref{alg1}}\label{apendix-lambda}
The Lagrangian of the optimization \eqref{sub-lambda} is
\begin{align}
L(\blambda, \bm \mu)&=-\sum_{i=1}^q \log \lambda_i +\frac{\beta}{2} \|\blambda -\b d\|_2^2 \\
&+ \mu_1(c_1 - \lambda_1) + \sum_{i=2}^q \mu_i(\lambda_{i-1}-\lambda_i)+\mu_{q+1}(\lambda_q-c_2). \nonumber
\end{align}
The KKT optimality conditions are derived as:
\begin{align}
-\frac{1}{\lambda_i} + \beta(\lambda_i-d_i)-\mu_i+\mu_{i+1}&=0, \ i=1,\cdots, q; \label{k1} \\
c_1-\lambda_1 &\leq 0; \label{k2} \\
\lambda_{i-1}-\lambda_i &\leq 0, \ i=2,\cdots, q; \label{k3}\\
\lambda_q - c_2 &\leq 0; \label{k4} \\
\mu_i &\geq 0, \ i=1, \cdots, q+1; \label{k5} \\
\mu_1(c_1-\lambda_1)&=0; \label{k6} \\
\mu_i(\lambda_{i-1}-\lambda_i)&=0, \ i=2,\cdots, q; \label{k7} \\
\mu_{q+1}(\lambda_q-c_2)&=0; \label{k8}
\end{align}

\begin{Lem1}\label{lem_c} 
	The solution of the KKT system \eqref{k1}-\eqref{k8} is $\lambda_i=(d_i+\sqrt{d_i^2+4/\beta})/2$, for $i=1,\cdots, q$, if $c_1 \leq \lambda_1 \leq \cdots \leq c_q \leq c_2$ hold.
	%\begin{align}
	%d_1 &\geq \alpha_1-\frac{1}{\alpha_1\beta}; \label{c1} \\
	%d_1 & \leq d_2 \leq \cdots \leq d_q; \label{c2} \\
	%d_q & \leq \alpha_2-\frac{1}{\alpha_2\beta}. \label{c3}
	%\end{align}
\end{Lem1}

\begin{proof} It is obvious that it conditions $c_1 \leq \lambda_1 \leq \cdots \leq \lambda_q \leq c_2$ hold, then the solutions of the primal and dual variables satisfy all equations.
\end{proof}
%If the conditions in Lemma \ref{lem_c} did not hold, the solution $\lambda_i=(d_i+\sqrt{d_i^2+4/\beta})/2$, $i=1,\cdots, q$, will violate some inequality constraints. In this case, we need to update $\b d$ so that the solution $\lambda_i=(d_i+\sqrt{d_i^2+4/\beta})/2$ still obeys the inequality constraints.

We start from the corresponding unconstrained version of the problem \eqref{sub-lambda} whose solution is
\begin{align}
\lambda_i^{(0)}=\left (d_i+\sqrt{d_i^2+4/\beta}\right )/2.
\end{align}
If this solution satisfies all the KKT conditions \eqref{k1}-\eqref{k8}, then it is also the optimal. Otherwise, each $\lambda_i^{(0)}$ that violates the conditions $c_1 \leq \lambda_1^{(0)} \leq \cdots \leq \lambda_q^{(0)} \leq c_2$ needs to be updated. % we need to update $\b \lambda^{(0)}$ iteratively. In the $k^{\rm th}$ iteration,

\textbf{Situation 1:} $c_1 \geq \lambda_1^{(0)} \geq \cdots \geq \lambda_r^{(0)}$, implying $c_1-\frac{1}{c_1\beta} \geq d_1 \geq \cdots \geq d_r$, where at least one inequality is strict and $r\geq 1$. Without loss of generality, let the $j$-th inequality is strict with $1\leq j \leq r$, i.e. $d_j > d_{j+1}$. The KKT optimality conditions for this pare are:
\begin{align}
-\frac{1}{\lambda_j} + \beta(\lambda_j-d_j)-\mu_j+\mu_{j+1}&=0; \label{c1} \\
-\frac{1}{\lambda_{j+1}} + \beta(\lambda_{j+1}-d_{j+1})-\mu_{j+1}+\mu_{j+2}&=0; \label{c2}\\
\lambda_{j}-\lambda_{j+1} &\leq 0; \label{c3}\\
\mu_i &\geq 0, \ i=j, j+1, j+2; \label{c4} \\
\mu_{j+1}(\lambda_{j}-\lambda_{j+1})&=0; \label{c5} 
\end{align}
We subtract the first two equations and obtain:
\begin{align}
2\mu_{j+1}=\mu_{j+2}+\mu_j+(\frac{1}{\lambda_j}-\frac{1}{\lambda_{j+1}})+\beta(\lambda_{j+1}-\lambda_j)+\beta(d_j-d_{j+1})> 0, \label{mu}
\end{align}
due to the fact that $d_j > d_{j+1}$ and $\lambda_j \leq \lambda_{j+1}$. Since $\mu_{j+1} >0$, we also have
\begin{align}
2\mu_{j}=\mu_{j+1}+\mu_{j-1}+(\frac{1}{\lambda_{j-1}}-\frac{1}{\lambda_{j}})+\beta(\lambda_{j}-\lambda_{j-1})+\beta(d_{j-1}-d_{j})> 0, \label{mu1}
\end{align}
where $d_{j-1} \geq d_j$ and $\lambda_{j-1} \leq \lambda_j$. Similarly, we can obtain $\mu_j > 0$ with $2 \leq j \leq r$. In addition,
\begin{align}
\mu_1 &= - \frac{1}{\lambda_1} + \beta(\lambda_1 - d_1)+ \mu_2 \\
& -\frac{1}{c_1}+\beta(c_1 - d_1) +\mu_2 >0.
\end{align}
Totally, we have $\mu_j > 0$ with $1 \leq j \leq r$. By \eqref{k6} and \eqref{k7}, we obtain $\lambda_1=\cdots =\lambda_r=c_1$. Therefore, we update
\begin{align}
\lambda_1^{(1)}= \cdots = \lambda_r^{(1)} =c_1.
\end{align}

\textbf{Situation 2:} $\lambda_s^{(0)} \geq \cdots \geq \lambda_q^{(0)} \geq c_2$, implying $d_s \geq \cdots \geq d_q \geq c_2-\frac{1}{c_2 \beta}$, where at least one inequality is strict and $s \leq q$.

Similar to situation 1, we can also obtain $\mu_j > 0$ with $s+1 \leq j \leq m+1$ and thus $\lambda_s = \cdots = \lambda_q = c_2$. Therefore, we update $\lambda_s^{(0)}, \cdots, \lambda_q^{(0)}$ by $\lambda_s^{(1)} = \cdots = \lambda_q^{(1)} = c_2$.

\textbf{Situation 3:} $\lambda_i^{(0)} \geq \cdots \geq \lambda_m^{(0)}$, implying $d_i \geq \cdots \geq d_m$, where at least one inequality is strict and $1 \leq i\leq m \leq q$. Here we assume $\lambda_{i-1}^{(0)} < \lambda_i^{(0)}$ ($c_1 <\lambda_1^{(0)}$ if $i=1$) and $\lambda_m^{(0)} < \lambda_{m+1}^{(0)}$ ($\lambda_q^{(0)}<c_2$ if $m=q$). Otherwise, this will be reduced to situation 1 or 3. 

%($c_1 <\lambda_1$ if $j=1$) and $\lambda_m < \lambda_{m+1}$ ($\lambda_q< c_2$ if $m=q$)} 

Similar to situation 1, we can also obtain $\mu_j > 0$ with $i+1 \leq j \leq m$ and thus $\lambda_i^{(1)}= \lambda_{i+1}^{(1)} = \cdots = \lambda_m^{(1)}$.

We sum up equations \eqref{c1} with $i \leq j \leq m$ and obtain
\begin{align}\label{sum_j}
-\frac{1}{\lambda_j} + \beta\lambda_j-\frac{1}{m-i+1}(\beta\sum_{j=i}^{m}d_j+\mu_i-\mu_{m+1})=0, \quad j=i, \cdots, m. 
\end{align}

Here we need to use iterative method to find the solution that satisfies KKT conditions. It is easy to check that $\mu_i = \mu_{m+1} = 0$ when $\lambda_{i-1}^{(0)} < \lambda_{i}^{(0)}$ and $\lambda_{m}^{(0)}<\lambda_{m+1}^{(0)}$. In that case, according to \eqref{sum_j}, we have
\begin{align}
\lambda_j = \left( \bar{d}_{i \to m } + \sqrt{\bar{d}^2_{i \to m} + 4/\beta} \right)/2, \quad j=i, \cdots, m.
\end{align}
where $\bar{d}_{i \to m} = \frac{1}{m-i+1} \sum_{j=i}^m d_j$. Therefore, we update $\lambda_i^{(0)}, \cdots, \lambda_m^{(0)}$ by
\begin{align} \label{up}
\lambda_i^{(1)}=\cdots =\lambda_m^{(1)} = \left( \bar{d}_{i \to m} + \sqrt{\bar{d}^2_{i \to m} + 4/\beta} \right)/2.
\end{align}
If there exists the case that $\lambda_{i-1}^{(1)} > \lambda_{i}^{(1)} $, we need to further update $\lambda_{i-1}^{(1)}, \lambda_{i}^{(1)}, \cdots, \lambda_m^{(1)}$ in the next iteration. It will include two cases to discuss:

1. $\lambda_{i-1}^{(1)}$ has not been updated by \eqref{up}, implying that $\lambda_{i-1}^{(1)} = \lambda_{i-1}^{(0)} = \left( \bar{d}_{i-1} + \sqrt{d^2_{i-1 } + 4/\beta} \right)/2$. So $\lambda_{i-1}^{(1)} > \lambda_{i}^{(1)}$ means $d_{i-1} > \bar{d}_{i \to m}$. KKT conditions for this pare are:
\begin{align}
-\frac{1}{\lambda_{i-1}} + \beta(\lambda_{i-1}-d_{i-1})-\mu_{i-1}+\mu_{i}&=0; \label{q1} \\
-\frac{1}{\lambda_{i}} + \beta(\lambda_{i}-\bar{d}_{i \to m})-\mu_{i}+\mu_{m+1}&=0; \label{q2}\\
\lambda_{i-1}-\lambda_{i} &\leq 0; \label{q3}\\
\mu_p &\geq 0, \ p=i-1, i, m+1; \label{q4} \\
\mu_{i}(\lambda_{i-1}-\lambda_{i})&=0; \label{q5} 
\end{align}
We subtract the first two equations and obtain
\begin{align}
2\mu_{i}=\mu_{i-1}+\mu_{m+1}+(\frac{1}{\lambda_{i-1}}-\frac{1}{\lambda_{i}})+\beta(\lambda_{i}-\lambda_{i-1})+\beta(d_{i-1}-\bar{d}_{i \to m})> 0, \label{mu}
\end{align}
and thus $\lambda_{i-1} = \lambda_i = \cdots = \lambda_m$. Then the equation \eqref{sum_j} can be written as
\begin{align}
-\frac{1}{\lambda_j} + \beta\lambda_j-\frac{1}{m-i+2}(\beta\sum_{j=i-1}^{m}d_j+\mu_{i-1}-\mu_{m+1})=0, \quad j=i-1, \cdots, m.
\end{align}
Hence, we update
\begin{align} \label{up2}
\lambda_{i-1}^{(2)} = \cdots = \lambda_{m}^{(2)}= \left( \bar{d}_{(i-1) \to m} + \sqrt{\bar{d}_{(i-1) \to m}^2 + 4/\beta} \right)/2.
\end{align}

2. $\lambda_{i-1}^{(1)}$ has been updated by \eqref{up}, implying that $\lambda_{t}^{(1)} = \cdots = \lambda_{i-1}^{(1)} = \left( \bar{d}_{t \to (i-1)} + \sqrt{d^2_{t \to (i-1)} + 4/\beta} \right)/2$ with $t < i-1$. Then $\lambda_{i-1}^{(1)} > \lambda_{i}^{(1)}$ means $\bar{d}_{t \to (i-1)} > \bar{d}_{i \to m}$. Similarly, we can also obtain $\lambda_t = \lambda_{t+1} = \cdots = \lambda_m$ by deriving KKT conditions. We sum up equations \eqref{k1} over $t \leq j \leq m$ and obtain
\begin{align}
-\frac{1}{\lambda_j} + \beta\lambda_j-\frac{1}{m-t+1}(\beta\sum_{j=t}^{m}d_j+\mu_t-\mu_{m+1})=0, \quad j=t, \cdots, m. \label{sum_t}
\end{align}
So we update
\begin{align} \label{up3}
\lambda_{t}^{(2)} = \cdots = \lambda_{m}^{(2)}= \left( \bar{d}_{t \to m} + \sqrt{\bar{d}_{t \to m}^2 + 4/\beta} \right)/2.
\end{align}

For the case that $\lambda_m^{(1)} > \lambda_{m+1}^{(1)}$, the update strategy is similar to the case $\lambda_{i-1}^{(1)} > \lambda_{i}^{(1)} $.

We iteratively check each situation and update the corresponding $\lambda_i$ accordingly. We can check that the algorithm will be terminated with the maximum number of iterations $q + 1$ and $c_1 \leq \lambda_1^{(q+1)} \leq \cdots \leq \lambda_q^{(q+1)} \leq c_2$ holds for all variables. Because each updating above is derived by KKT optimality conditions for Problem \eqref{sub-lambda}, the iterative-update procedure summarized in Algorithm \ref{algo-lambda} converges to the KKT point of Problem \eqref{sub-lambda}.

\subsection{Proof for Theorem \ref{thm:conv}}\label{apendix-thm}
\begin{proof} The proof of algorithm convergence is partly based on the proof of BSUM in \citet{razaviyayn2013unified}. We first show the linear independence constraint qualification on unitary constraint set $\S_{{U}} \triangleq \{{U} \in \mathbb{R}^{p \times q} | {U}^T{U}={I}_q \}$.

\begin{Lem1}\label{lem:conv}
	Linear independence constraint qualification (LICQ) holds on each ${U} \in \S_{{U}} $.
\end{Lem1}

\begin{proof}
	
We rewrite $\S_{{U}}$ as
\begin{align}
\{ {U} \in \mathbb{R}^{p \times q} | g_{ij}( U)= \sum_{k=1}^p u_{ki}u_{kj} - I_{ij}, \forall 1 \leq i \leq j \leq q \},
\end{align}
where $u_{ij}$ and $I_{ij}$ are the elements of $ U$ and identity matrix $ I$ in $i$-th row and $j$-th column, respectively. It is observed that
\begin{equation} 
\nabla g_{ij}( U)=
\left\{ 
\begin{array}{lc} 
{[ 0_{p \times (i-1)}; 2 u_i; 0_{p \times (q-i)} ],} & {\rm if}\ i=j ;\\ 
{[ 0_{p \times (i-1)}; u_j; 0_{p \times (j-i-1)}; u_i; 0_{p \times (q-j)} ],} & {\rm otherwise}. 
\end{array} 
\right. 
\end{equation}
We can see $ u_i$ from $\nabla g_{ii}( U)$ will only appear in $i$-th column, but $ u_i$ from $\nabla g_{ij}( U)$ with $i \neq j$ will not appear in $i$-th column. Consequently, each $\nabla g_{ij}( U)$ cannot be expressed as a linear combination of the others, thus each $\nabla g_{ij}( U)$ is linear independent. 
\end{proof}

Now we prove Theorem \ref{thm:conv}. It is easy to check that the level set $\{(\w, U, \blambda)|f(\w, U, \blambda) \leq f(\w^{(0)}, U^{(0)}, \blambda^{(0)})\}$ is compact, where $f(\w, U, \blambda)$ is the cost function in Problem \eqref{CGL_L}. Furthermore, the sub-problems \eqref{sub-x} and \eqref{sub_lambda} have unique solutions since they are strictly convex problems and we get the global optima. According to Theorem 2 in \citet{razaviyayn2013unified}, we obtain that the sequence $(\w^{(t)}, U^{(t)}, \blambda^{(t)})$ generated by Algorithm \ref{algo-1} converges to the set of stationary points. Note that $ U$ is constrained on the orthogonal Stiefel manifold that is nonconvex, while BSUM framework does not cover nonconvex constraints. But the subsequence convergence can still be established \citep{fu2017scalable} due to the fact that the cost function value here is non-increasing and bounded below in each iteration.

Next we will further show that each limit of the sequence $(\w^{(t)}, U^{(t)}, \blambda^{(t)})$ satisfies KKT conditions of Problem \eqref{CGL_L_cost}. Let $({\w}, { U}, { \blambda})$ be a limit point of the generated sequence. The Lagrangian function of \eqref{CGL_L_cost} is
\begin{align}
L(\w, U, \blambda, \bm \mu_1,\bm \mu_2, M)= &- \log \text{gdet} (\Di(\blambda))+\tr{ K \L \w}+\frac{\beta}{2}\| \L \w - U \Di(\blambda) U^T\|_F^2 \nonumber \\ 
&- \bm \mu_1^T\w + \bm \mu_2^T h( \blambda) + \tr{ M^T( U^T U - I_q)},
\end{align}
where $\bm \mu_1$, $\bm \mu_2$ and $M$ are dual variables, and $\bm \mu_2^T h( \blambda)= \mu_{2,1}(\alpha_1 - \lambda_1) + \sum_{i=2}^q \mu_{2,i}(\lambda_{i-1}-\lambda_i)+\mu_{2,q+1}(\lambda_q-\alpha_2)$ with $ \bm \mu_2=[\mu_{2,1}, \cdots, \mu_{2,q+1}]^T$.

\vspace{3ex}

\noindent (1) we can see ${ \blambda}$ is derived from KKT conditions of sub-problem \eqref{sub_lambda}. Obviously, ${ \blambda}$ also satisfies KKT conditions of Problem \eqref{CGL_L_cost}. 

\vspace{3ex}

\noindent (2) we show ${\w}$ satisfies KKT conditions \eqref{CGL_L_cost}. The KKT conditions with $\w$ can be derived as:
\begin{align}
 \L^*\L \w - \L^*( U \Di(\blambda) U^T - \beta^{-1} K) -\beta^{-1} \bm \mu_1 = 0; \label{m1} \\
 \bm \mu_1^T \w =0; \label{m2} \\
\w \geq 0; \label{m3} \\
 \bm \mu_1 \geq 0; \label{m4} 
\end{align}
Know ${\w}$ is derived by KKT system see Lemma \ref{lem-kkt-w}, we obtain
\begin{align}
{\w} - ({\w} - \frac{1}{L_1}(\L^*\L {\w}- c))- \mu = 0,
\end{align}
and $ c=\L^*({ U}{\Di(\blambda)}{ U}^T-\beta^{-1} K)$. So we have
\begin{align}
 \L^*\L {\w}-\L^*({ U}{ \Di(\blambda)}{ U}^T-\beta^{-1} K) - \frac{1}{L_1} \mu = 0,
\end{align}
Therefore, ${\w}$ also satisfies KKT conditions \eqref{CGL_L_cost}.

\vspace{3ex}

\noindent (3) KKT conditions with respect to $ U$ are as below:
\begin{align}
\L \w U\Di(\blambda) - \frac{1}{2} U(\Di(\blambda)^2+\beta^{-1}(M+M^T)) = 0; \label{d1}\\
 U^T U = I_q. \label{d2}
\end{align}
Since ${ U}$ admits the first order optimality condition on orthogonal Stiefel manifold, we have 
\begin{align}
\L {\w }{ U} \Di(\blambda) - { U}( { U}^T \L{\w} { U} { \Di(\blambda)} - \frac{1}{2} [{ U}^T \L {\w} { U}, { \Di(\blambda)}])= 0,
\end{align}
where $[A,B]=AB-BA$. Note that ${ U}^T \L{\w} { U}$ is a diagonal matrix according to the update of $U$. So there must exist a $ M$ such that ${ U}$ satisfies \eqref{d1}. Therefore, $({\w}, { U}, { \blambda})$ satisfies KKT conditions of Problem \eqref{CGL_L_cost}.

\end{proof}

\subsection{Proof for Lemma \ref{iso-adj}}\label{appendix-isotonic}
\begin{proof}
Note that both $\boldsymbol{\psi}$ and ${\b e}$ are diagonal, and additionally we require the values for $\boldsymbol{\psi}$ to be symmetric across zero. We can express the least square $ \| \boldsymbol{\psi}-{\b e} \|_2^2$ in \eqref{sub_psi} as 
\begin{align}
\| \boldsymbol{\psi}-{\b e} \|_2^2=&\sum_{i=1}^{b/2}(\psi_i-e_i)^2+(\psi_i-e_{b-i+1})^2 \nonumber \\
&=2\sum_{i=1}^{b/2}\left(\psi_i^2-2\psi_i\frac{e_i+e_{b-i+1}}{2} + \frac{e_i^2+e_{b-i+1}^2}{2}\right)\\
&= 2\sum_{i=1}^{b/2}\left(\left(\psi_i- \frac{e_i+e_{b-i+1}}{2} \right)^2 + \left(\frac{e_i+e_{b-i+1}}{2} \right)^2 +\frac{e^2_i+e_{b-i+1}^2}{2} \right)
\end{align}

Thus, we can write
\begin{align}\label{sub_psi-app}
\| \boldsymbol{\psi}-{\b e} \|_2^2=\text{constant}+ \| \tilde{\boldsymbol{\psi}}-\tilde{{\b e}} \|_2^2
\end{align}
 where $\tilde{\boldsymbol{\psi}}=[\psi_1,\psi_2,\cdots, \psi_{b/2}]$ are the first half elements of $ {\boldsymbol{\psi}}$, and $\tilde{\b e}= [\tilde{e}_1,\tilde{e}_2,\cdots,\tilde{e}_{b/2}]$, with $\{\tilde{e}_i=\frac{e_i+e_{b-i+1}}{2}\}_{i=1}^{b/2} $. 
\end{proof}

\subsection{Proof for Theorem \ref{thm:conv:lap-adj}}\label{appendix-conv:lap-adj}
\begin{proof}
	We cannot directly apply the established convergence results of BSUM in \citep{razaviyayn2013unified} to prove our algorithm, because there are two variables $U$ and $V$ that may not have unique solutions in iterations while the paper \citep{razaviyayn2013unified} only allows one variable to enjoy multiple optimal solutions. However, we can still establish the subsequence convergence as below by following the proof in \citep{razaviyayn2013unified} together with the fact that $\blambda$ and $U$ are updated independently with $\bpsi$ and $V$.
	
	For the convenience of description, let $\x = (\x_1, \x_2, \x_3, \x_4, \x_5)=(\w, \blambda, U, \bpsi, V)$ with each $\x_i \in \mathcal{X}_i$. Since the iterates $\x^k$ are in a compact set, there must be a limit point for the sequence $\{ \x^k \}$. Now we need to show every limit point of the iterates is a stationary point of \eqref{CGL_A-Rlx}.
	
	Let $\bar{\x} = (\bar{\x}_1, \bar{\x}_2, \bar{\x}_3, \bar{\x}_4, \bar{\x}_5)$ is a limit point of $\{ \x^k \}$, and $\{ \x^{k_j} \}$ be the subsequence converging to $\bar{\x}$. Without loss of generality, we can assume that
	\begin{align}\label{sub x1}
	\x_1^{k_j} = \arg \min_{\x_1} g(\x_1 | \x^{k_j-1}),
	\end{align}
	where $g(\x_1 | \x^{k_j-1})$ is the majorized function defined in \eqref{sur}.
	
	For the convenience of proof, we change the updating order along with $(\w, \blambda, U, \bpsi, V)$ and have the following updating procedure:
	\begin{align}\label{updating}
	&\x_1^{k_j} = \arg \min_{\x_1} g(\x_1 | \x^{k_j-1}),\\
	&\x^{k_j}=[\x_1^{k_j}, \x_2^{k_j-1}, \x_3^{k_j-1}, \x_4^{k_j-1}, \x_5^{k_j-1}]^T;\\
	&\x_2^{k_j+1} = \arg \min_{\x_2} F_2(\x_2 | \x^{k_j}),\\
	&\x^{k_j+1}=[\x_1^{k_j}, \x_2^{k_j+1}, \x_3^{k_j}, \x_4^{k_j}, \x_5^{k_j}]^T;\\
	&\x_3^{k_j+2} = \arg \min_{\x_3} F_3(\x_3 | \x^{k_j+1}),\\
	&\x^{k_j+2}=[\x_1^{k_j+1}, \x_2^{k_j+1}, \x_3^{k_j+2}, \x_4^{k_j+1}, \x_5^{k_j+1}]^T;\\
	&\x_4^{k_j+3} = \arg \min_{\x_4} F_4(\x_4 | \x^{k_j+2}),\\
	&\x^{k_j+3}=[\x_1^{k_j+2}, \x_2^{k_j+2}, \x_3^{k_j+2}, \x_4^{k_j+3}, \x_5^{k_j+2}]^T;\\
	&\x_5^{k_j+4} = \arg \min_{\x_5} F_5(\x_5 | \x^{k_j+3}),\\
	&\x^{k_j+4}=[\x_1^{k_j+3}, \x_2^{k_j+3}, \x_3^{k_j+3}, \x_4^{k_j+3}, \x_5^{k_j+4}]^T;
	\end{align}
	where $F_i(\x_i | \x^{k_j+i-1})$ is the cost function $F (\x)$ in \eqref{CGL_A-Rlx} with respect to $\x_i$ and other variables fixed, i.e., $F_i(\x_i | \x^{k_j+i-1}) = F (\x_1^{k_j+i-1}, \cdots, \x_{i-1}^{k_j+i-1}, \x_i, \x_{i+1}^{k_j+i-1}, \x_5^{k_j+i-1})$, $i \in \{2, 3, 4, 5\}$. Note that here we increase the iteration number $k$ when updating each variable, which is different with the notation $t$ in Algorithm \ref{algo-3} where we increase $t$ only after updating all the variables. 
	
	We further restrict the subsequence such that
	\begin{align}\label{limit}
	\lim_{j \to \infty} \x^{k_j+i} = \b{z}^i, \quad \forall i=-1, 0, 1, \cdots, 4,
	\end{align}
	where $\z^0=\bar{\x}$. 
	
	Since the cost function $F (\x)$ in \eqref{CGL_A-Rlx} is continuous and nonincreasing, we have
	\begin{align}\label{limit_F}
	F(\z^{-1})=F(\z^0)= \cdots = F(\z^4).
	\end{align}
	
	By the majorized function property \eqref{bsum-maj-1}, we have
	\begin{align}\label{upt_w}
	F(\x^{k_j}) \leq g(\x_1^{k_j} | \x^{k_j-1}) \leq g(\x_1^{k_j-1} | \x^{k_j-1}) = F(\x^{k_j-1}).
	\end{align}
	By the continuity of $g (\cdot)$ according to \eqref{bsum-maj-4}, we take the limit $j \to \infty$ and obtain
	\begin{align}\label{limit_w}
	g(\z^0_1 | \z^{-1}) = g(\z^{-1}_1 | \z^{-1}),
	\end{align}
	Since $\z_1^0$ is the minimizer of $g(\x_1 | \z^{-1})$ and $g(\x_1 | \z^{-1})$ has the unique minimizer, we have $\z_1^0 = \z_1^{-1}$. We can see only $\z_1$ is updated from $\z^{-1}$ to $\z^0$, so we further obtain $\z^0 = \z^{-1}$.
	
	Regarding $\x_2$, we have
	\begin{align}\label{upt_x2}
	F(\x^{k_j+1}) \leq F_2(\x_2^{k_j+1} | \x^{k_j}) \leq F_2(\x_2^{k_j} | \x^{k_j}) = F(\x^{k_j}).
	\end{align}
	By the continuity of $F_2 (\cdot)$, we take the limit $j \to \infty$ and get
	\begin{align}\label{limit_x2}
	F_2(\z^1_2 | \z^{0}) = F_2(\z^{0}_2 | \z^{0}).
	\end{align}
	Considering $F_2(\x_2 | \z^{0})$ has the unique minimizer, we have $\z_2^1 = \z_2^{0}$. Similarly, only $\z_2$ is updated from $\z^{0}$ to $\z^1$ and thus $\z^1 = \z^{0}$. We can also obtain $\z^3 = \z^2$ by the uniqueness of the minimizer of $F_4(\cdot)$.
	
	Since $\x_2$ and $\x_3$ are updated independently with $\x_4$ and $\x_5$, we have
	\begin{align}\label{upt_x4}
	F_4(\x_4^{k_j+3} | \x^{k_j+2}) &= F_4(\x_4^{k_j+3} | \x^{k_j}) - (F(\x^{k_j+1})-F(\x^{k_j+2})) - (F(\x^{k_j})-F(\x^{k_j+1})) \\
	& \leq F_4(\x_4^{k_j} | \x^{k_j}) - (F(\x^{k_j+1})-F(\x^{k_j+2})) - (F(\x^{k_j})-F(\x^{k_j+1})) \\
	& = F(\x^{k_j}) - (F(\x^{k_j+1})-F(\x^{k_j+2})) - (F(\x^{k_j})-F(\x^{k_j+1})).
	\end{align}
	and
	\begin{align}\label{limit_x4q}
	F_4(\x_4^{k_j+3} | \x^{k_j+2}) \geq F(\x^{k_j+3}).
	\end{align}
	Then we take the limit $j \to \infty$ and obtain
	\begin{align}\label{limit_x4q1}
	F(\z^3) \leq F_4(\z_4^{3} | \z^{2}) = F_4(\z_4^{3} | \z^{0}) \leq F_4(\z_4^{0} | \z^{0}) = F(\z^{0}).
	\end{align}
	Together with $F(\z^3)=F(\z^0)$, we have
	\begin{align}\label{limit_x4q2}
	F_4(\z_4^{3} | \z^{0}) = F_4(\z_4^{0} | \z^{0}).
	\end{align}
	Since $\z_4^{3}$ is the minimizer of $F_4(\x_4 | \z^{2})$ as well as $F_4(\x_4 | \z^{0})$, and the minimizer of $F_4(\x |\z^{0})$ is unique, we can get
	\begin{align}\label{limit_x4q3}
	\z_4^{3} = \z_4^{0}.
	\end{align}
	
	By the fact that $\z_1^0$ is the minimizer of $g(\x_1 | \z^{-1})$ and $\bar{\x} = \z^0 = \z^{-1}$, we have
	\begin{align}\label{st1}
	g(\bar{\x}_1 |\bar{\x}) \leq g(\x_1 | \bar{\x}), \quad \forall \ \x_1 \in \mathcal{X}_1.
	\end{align}
	which implies
	\begin{align}\label{stationary1}
	g'(\x_1 |\bar{\x}) |_{\x_1 = \bar{\x}_1} = \b{0}.
	\end{align}
	By the majorized function property \eqref{bsum-maj-3} and $F(\cdot)$ is differentiable, we can get
	\begin{align}\label{stationary11}
	F'(\x_1, \bar{\x}_2, \bar{\x}_3, \bar{\x}_4, \bar{\x}_5) |_{\x_1 = \bar{\x}_1} = \b{0}.
	\end{align}

	Similarly, we can also obtain
	\begin{align}\label{st2}
	F_2(\bar{\x}_2 |\bar{\x}) \leq F_2(\x_2 | \bar{\x}), \quad \forall \ \x_2 \in \mathcal{X}_2.
	\end{align}
	implying
	\begin{align}\label{stationary2}
	F'(\bar{\x}_1, \x_2, \bar{\x}_3, \bar{\x}_4, \bar{\x}_5)|_{\x_2 = \bar{\x}_2} = \b{0}.
	\end{align}
	
	By the fact that $\z_4^{3}$ is the minimizer of $F_4(\x_4 | \z^{0})$, and Eq. \eqref{limit_x4q3}, we get
	\begin{align}\label{st3}
	F_4(\bar{\x}_4 |\bar{\x}) \leq F_4(\x_4 | \bar{\x}), \quad \forall \ \x_4 \in \mathcal{X}_4.
	\end{align}
	and thus
	\begin{align}\label{stationary3}
	F'(\bar{\x}_1, \bar{\x}_2, \x_3, \bar{\x}_4, \bar{\x}_5)|_{\x_3 = \bar{\x}_3} = \b{0}.
	\end{align}

	For $\x_3$, we have
	\begin{align}\label{upt_x3}
	F(\x^{k_j+2}) \leq F_3(\x_3^{k_j+2} | \x^{k_j+1}) \leq F_3(\x_3^{k_j+1} | \x^{k_j+1}) = F(\x^{k_j+1}).
	\end{align}
	By the continuity of $F_3(\cdot)$, we take the limit $j \to \infty$ and get
	\begin{align}\label{limit_x3}
	F_3(\z^2_3 | \z^{1}) = F_3(\z^{1}_3 | \z^{1}).
	\end{align}
	Since $\z_3^{1}$ is the minimizer of $F_3(\x_3|\z^1)$. By $\z^1 = \z^0 = \bar{\x}$, we obtain
	\begin{align}\label{st4}
	F_3(\bar{\x}_3 |\bar{\x}) \leq F_3(\x_3 | \bar{\x}), \quad \forall \ \x_3 \in \mathcal{X}_3.
	\end{align}
	implying
	\begin{align}\label{stationary4}
	F'(\bar{\x}_1, \bar{\x}_2, \bar{\x}_3, \x_4, \bar{\x}_5)|_{\x_4 = \bar{\x}_4} = \b{0}.
	\end{align}
	
	Similarly, we can also have 
	\begin{align}\label{st5}
	F_5(\bar{\x}_5 |\bar{\x}) \leq F_5(\x_5 | \bar{\x}), \quad \forall \ \x_5 \in \mathcal{X}_5.
	\end{align}
	and thus
	\begin{align}\label{stationary5}
	F'(\bar{\x}_1, \bar{\x}_2, \bar{\x}_3, \bar{\x}_4, \x_5)|_{\x_5 = \bar{\x}_5} = \b{0}.
	\end{align}
	
	Together with Eq. \eqref{stationary11}, \eqref{stationary2}, \eqref{stationary3}, \eqref{stationary4} and \eqref{stationary5}, we can conclude that $\bar{\x}$ is a stationary point of \eqref{CGL_A-Rlx}. Next, we only need to prove that each limit of the sequence $\x^k$ satisfies KKT conditions of \eqref{CGL_A-Rlx}. The proof is very similar to that for Theorem \ref{thm:conv} and thus we omit it here.

\end{proof}

\bibliography{refs}

\end{document}